\documentclass{article}

\usepackage{microtype}
\usepackage{graphicx}
\usepackage{booktabs} % for professional tables

\usepackage{amsthm}
\usepackage{amsmath}
\usepackage{amssymb}
\usepackage[utf8]{inputenc}
\usepackage[english]{babel}
 \usepackage{subcaption}
\usepackage[noend]{algorithmic}

\usepackage{makecell}

\usepackage[toc,page]{appendix}

\usepackage{commath}

\usepackage{hyperref}
\usepackage{cleveref}
\crefformat{footnote}{#2\footnotemark[#1]#3}

\usepackage[accepted]{icml2020}

\newtheorem{theorem}{Theorem}

\newtheorem{lemma}{Lemma}
\icmltitlerunning{R2-B2: Recursive Reasoning-Based Bayesian Optimization for No-Regret Learning in Games}

\begin{document}

\twocolumn[
\icmltitle{R2-B2: Recursive Reasoning-Based Bayesian Optimization for\\No-Regret Learning in Games}



\icmlsetsymbol{equal}{*}

\begin{icmlauthorlist}
\icmlauthor{Zhongxiang Dai}{nus}
\icmlauthor{Yizhou Chen}{nus}
\icmlauthor{Bryan Kian Hsiang Low}{nus}
\icmlauthor{Patrick Jaillet}{mit}
\icmlauthor{Teck-Hua Ho}{nus_business}
\end{icmlauthorlist}

\icmlaffiliation{nus}{Department of Computer Science, National University of Singapore, Republic of Singapore}
\icmlaffiliation{mit}{Department of Electrical Engineering and Computer Science, Massachusetts Institute of Technology, USA}
\icmlaffiliation{nus_business}{NUS Business School, National University of Singapore, Republic of Singapore}

\icmlcorrespondingauthor{Bryan Kian Hsiang Low}{lowkh@comp.nus.edu.sg}

\icmlkeywords{Bayesian Optimization, Recursive Reasoning, Repeated Games, Adversarial Machine Learning}

\vskip 0.3in
]



\printAffiliationsAndNotice{}  % leave blank if no need to mention equal contribution

\begin{abstract}
This paper presents a recursive reasoning formalism of \emph{Bayesian optimization} (BO) to model the reasoning process in the interactions between boundedly rational, self-interested agents with unknown, complex, and costly-to-evaluate payoff functions in repeated games, which we call \emph{\underline{R}ecursive \underline{R}easoning-\underline{B}ased \underline{B}O} (R2-B2). 
Our R2-B2 algorithm is general in that it does not constrain the relationship among the payoff functions of different agents and can thus be applied to various types of games such as constant-sum, general-sum, and common-payoff games. We prove that by reasoning at level $2$ or more and at one level higher than the other agents, our R2-B2 agent can achieve faster asymptotic convergence to no regret than that without utilizing recursive reasoning. We also propose a computationally cheaper variant of R2-B2 called R2-B2-Lite at the expense of a weaker convergence guarantee. The performance and generality of our R2-B2 algorithm are empirically demonstrated using synthetic games, adversarial machine learning, and multi-agent reinforcement learning.
\end{abstract}

\section{Introduction}
\label{section:intro}
Several fundamental machine learning tasks in the real world involve intricate interactions between boundedly rational\footnote{\label{brat}Boundedly rational agents are subject to limited cognition and time in making decisions~\cite{Reinhard2002}.}, self-interested agents that can be modeled as a form of repeated games with unknown, complex, and costly-to-evaluate payoff functions for the agents. For example, in adversarial \emph{machine learning} (ML),
the interactions between the \emph{defender} $\mathcal{D}$ and the \emph{attacker} $\mathcal{A}$ of an ML model can be modeled as a repeated game in which the payoffs to $\mathcal{D}$ and $\mathcal{A}$ are the performance of the ML model (e.g., validation accuracy) and its negation, respectively. Specifically, given a fully trained image classification model (say, provided as an online service), $\mathcal{A}$ attempts to fool the ML model into misclassification through repeated queries of the model using perturbed input images. On the other hand, for each queried image that is perturbed by $\mathcal{A}$, $\mathcal{D}$ tries to ensure the correctness of its classification by transforming the perturbed image before feeding it into the ML model. As another example, \emph{multi-agent reinforcement learning} (MARL) in an episodic environment can also be modeled as a repeated game in which the payoff to each agent is its return from the execution of all the agents' selected policies.

Solving such a form of repeated games in a cost-efficient manner is challenging since the payoff functions of the agents are unknown, complex (e.g., possibly noisy, non-convex, and/or with no closed-form expression/derivative), and costly to evaluate. Fortunately, the payoffs corresponding to different actions of each agent tend to be correlated. For example, in adversarial ML, the correlated perturbations performed by the attacker $\mathcal{A}$ (and correlated transformations executed by the defender $\mathcal{D}$) are likely to induce similar effects on the performance of the ML model. Such a correlation can be leveraged to \emph{predict} the payoff associated with any action of an agent using a \emph{surrogate} model such as the rich class of Bayesian nonparametric \emph{Gaussian process} (GP) models~\cite{rasmussen2004gaussian} which is expressive enough to represent a predictive belief of the unknown, complex payoff function over the action space of the agent. Then, in each iteration, the agent can select an action for evaluating its unknown payoff function that trades off between sampling at or near to a likely maximum payoff based on the current GP belief (exploitation) vs. improving the GP belief (exploration) until its cost/sampling budget is expended. To do this, the agent can use a sequential black-box optimizer such as the celebrated \emph{Bayesian optimization} (BO) algorithm~\cite{shahriari2016taking} based on the \emph{GP-upper confidence bound} (GP-UCB) acquisition function~\cite{srinivas2009gaussian}, which guarantees asymptotic no-regret performance and is sample-efficient in practice. How then can we design a BO algorithm to account for its interactions with boundedly rational\cref{brat}, self-interested agents and still guarantee the trademark asymptotic no-regret performance?

Inspired by the cognitive hierarchy model of games~\cite{camerer2004cognitive}, we adopt a recursive reasoning formalism (i.e., typical among humans) to model the reasoning process in the interactions between boundedly rational\cref{brat}, self-interested agents. It comprises $k$ levels of reasoning which represents the cognitive limit of the agent. At level $k=0$ of reasoning, the agent randomizes its choice of actions. At a higher level $k \geq 1$ of reasoning, the agent selects its best response to the actions of the other agents who are reasoning at lower levels $0, 1, \ldots, k - 1$.

This paper presents the first recursive reasoning formalism of BO to model the reasoning process in the interactions between boundedly rational\cref{brat}, self-interested agents with unknown, complex, and costly-to-evaluate payoff functions in repeated games, which we call \emph{\underline{R}ecursive \underline{R}easoning-\underline{B}ased \underline{B}O} (R2-B2) (Section~\ref{r2b2}). 
R2-B2 provides these agents with principled strategies for performing effectively in this type of game.
In this paper, we consider repeated games with simultaneous moves and perfect monitoring\footnote{\label{smpm}In each iteration of a repeated game with (a) simultaneous moves and (b) perfect monitoring, every agent, respectively, (a) chooses its action simultaneously without knowing the other agents' selected actions, and (b) has access to the entire history of game plays, which includes all actions selected and payoffs observed by every agent in the previous iterations.}.
Our R2-B2 algorithm is general in that it does not constrain the relationship among the payoff functions of different agents and can thus be applied to various types of games such as constant-sum games (e.g., adversarial ML in which the attacker $\mathcal{A}$ and defender $\mathcal{D}$ have opposing objectives), general-sum games (e.g., MARL where all agents have possibly different yet not necessarily conflicting goals), and common-payoff games (i.e., all agents have identical payoff functions). We prove that by reasoning at level $k\geq 2$ and one level higher than the other agents, our R2-B2 agent can achieve faster asymptotic convergence to no regret than that without utilizing recursive reasoning (Section~\ref{subsec:level_k_policy}). 
We also propose a computationally cheaper variant of R2-B2 called R2-B2-Lite at the expense of a weaker convergence guarantee (Section~\ref{lite}). The performance and generality of R2-B2 are demonstrated through extensive experiments using synthetic games, adversarial ML, and MARL (Section~\ref{expt}). Interestingly, we empirically show that by reasoning at a higher level, our R2-B2 defender is able to effectively defend against the attacks from the state-of-the-art black-box adversarial attackers (Section~\ref{exp:comparison_parsimonious}), which can be of independent interest to the adversarial ML community.

\section{Background and Problem Formulation}
\label{background}
For simplicity, we will mostly focus on repeated games between two agents, but have extended our R2-B2 algorithm to games involving \emph{more than two} agents, as detailed in Appendix~\ref{subsec:more_than_two_players}. To ease exposition, throughout this paper, we will use adversarial ML as the running example and thus refer to the two agents as the \emph{attacker} $\mathcal{A}$ and the \emph{defender} $\mathcal{D}$. 
For example, the input action space $\mathcal{X}_1\subset \mathbb{R}^{d_1}$ of $\mathcal{A}$ can be a set of allowed perturbations of a test image  
while the input action space $\mathcal{X}_2\subset \mathbb{R}^{d_2}$ of $\mathcal{D}$ can represent a set of feasible transformations of the perturbed test image.
We consider both input domains $\mathcal{X}_1$ and $\mathcal{X}_2$ to be discrete for simplicity; 
generalization of our theoretical results in Section~\ref{r2b2} to continuous, compact domains can be easily achieved through a suitable discretization of the domains~\cite{srinivas2009gaussian}.
When the ML model is an image classification model, the payoff function $f_1:\mathcal{X}_1 \times \mathcal{X}_2 \rightarrow \mathbb{R}$ of $\mathcal{A}$, 
which takes in its perturbation $\mathbf{x}_1 \in \mathcal{X}_1$ and $\mathcal{D}$'s transformation $\mathbf{x}_2 \in \mathcal{X}_2$ as inputs,
can be the maximum predictive probability among all incorrect classes for a test image since $\mathcal{A}$ intends to cause misclassification.
Since $\mathcal{A}$ and $\mathcal{D}$ have opposing objectives  (i.e., $\mathcal{D}$ intends to prevent misclassification), the payoff function $f_2: \mathcal{X}_1 \times \mathcal{X}_2 \rightarrow \mathbb{R}$ of $\mathcal{D}$ can be the negation of that of $\mathcal{A}$, thus resulting in a constant-sum game between $\mathcal{A}$ and $\mathcal{D}$.

In each iteration $t=1,\ldots,T$ of the repeated game with simultaneous moves and perfect monitoring\cref{smpm}\footnote{Note that in some tasks such as adversarial ML, the requirement of perfect monitoring can be relaxed considerably. Refer to Section~\ref{exp:comparison_parsimonious} for more details.}, $\mathcal{A}$ and $\mathcal{D}$ select their respective input actions $\mathbf{x}_{1,t}$ and $\mathbf{x}_{2,t}$ simultaneously using our R2-B2 algorithm (Section~\ref{r2b2}) for evaluating their payoff functions $f_1$ and $f_2$.
Then, $\mathcal{A}$ and $\mathcal{D}$ receive the respective noisy observed payoffs $y_{1,t}\triangleq f_1(\mathbf{x}_{1,t}, \mathbf{x}_{2,t})+\epsilon_1$ and $y_{2,t}\triangleq f_2(\mathbf{x}_{1,t}, \mathbf{x}_{2,t})+\epsilon_2$ with i.i.d. Gaussian noises 
$\epsilon_i \sim \mathcal{N}(0,\sigma_i^2)$ 
and noise variances $\sigma_i^2$ for $i=1,2$.

A common practice in game theory is to measure the performance of $\mathcal{A}$ via its \emph{(external) regret}~\cite{nisan2007algorithmic}:
\begin{equation}
\begin{array}{c}
    R_{1,T} \triangleq \sum^T_{t=1} [f_1(\mathbf{x}^*_1,\mathbf{x}_{2,t})-f_1(\mathbf{x}_{1,t},\mathbf{x}_{2,t})]
\end{array}
\label{extregret}
\end{equation} 
where $\mathbf{x}^*_1\triangleq\mathop{\arg\max}_{\mathbf{x}_1\in \mathcal{X}_1} \sum^T_{t=1} f_1(\mathbf{x}_1, \mathbf{x}_{2,t})$.
The external regret $R_{2,T}$ of $\mathcal{D}$ is defined in a similar manner. 
An algorithm is said to achieve asymptotic \emph{no regret} if $R_{1,T}$ grows sub-linearly in $T$, i.e., $\lim_{T\rightarrow \infty} R_{1,T}/T = 0$.
Intuitively, by following a no-regret algorithm, $\mathcal{A}$ is guaranteed to eventually find its optimal input action $\mathbf{x}^*_1$ in hindsight,
regardless of $\mathcal{D}$'s sequence of input actions.

To guarantee no regret (Section~\ref{r2b2}), $\mathcal{A}$ represents a predictive belief of its unknown, complex payoff function $f_1$ 
using the rich class of \emph{Gaussian process} (GP) models
by modeling $f_1$ as a sample of a GP~\cite{rasmussen2004gaussian}.
$\mathcal{D}$ does likewise with its unknown $f_2$.
Interested readers are referred to Appendix~\ref{app:gp} for a detailed background on GP.
In particular, $\mathcal{A}$ uses the GP predictive/posterior belief of $f_1$ to compute a probabilistic upper bound of $f_1$ called the \emph{GP-upper confidence bound} (GP-UCB)~\cite{srinivas2009gaussian} at any joint input actions $(\mathbf{x}_1, \mathbf{x}_2)$, which will be exploited by our R2-B2 algorithm (Section~\ref{r2b2}):
\begin{equation}
\alpha_{1,t}(\mathbf{x}_1, \mathbf{x}_2)\triangleq\mu_{t-1}(\mathbf{x}_1, \mathbf{x}_2) + \beta^{1/2}_t \sigma_{t-1}(\mathbf{x}_1, \mathbf{x}_2)
\label{eqgpucb}
\end{equation} 
for iteration $t$ where $\mu_{t-1}(\mathbf{x}_1, \mathbf{x}_2)$ and $\sigma^2_{t-1}(\mathbf{x}_1, \mathbf{x}_2)$ 
denote, respectively, the GP posterior mean and variance at $(\mathbf{x}_1, \mathbf{x}_2)$ (Appendix~\ref{app:gp}) conditioned on the history of game plays up till iteration $t-1$ that includes $\mathcal{A}$'s observed payoffs and the actions selected by both agents in iterations $1,\ldots,t-1$.
The GP-UCB acquisition function $\alpha_{2,t}$ for $\mathcal{D}$ is defined likewise.
Supposing $\mathcal{A}$ knows the input action $\mathbf{x}_{2,t}$ selected by $\mathcal{D}$ and chooses an input action $\mathbf{x}_1$ to maximize the GP-UCB acquisition function $\alpha_{1,t}$~\eqref{eqgpucb}, its choice involves trading off between sampling close to an expected maximum payoff (i.e., with large GP posterior mean) given the current GP belief of $f_1$ (exploitation) vs.~that of high predictive uncertainty (i.e., with large GP posterior variance) to improve the GP belief of $f_1$ (exploration) where the parameter $\beta_t$ is set to trade off between exploitation vs.~exploration for bounding its external regret~\eqref{extregret}, as specified later in Theorem~\ref{regret_gp}.

\section{\underline{R}ecursive \underline{R}easoning-\underline{B}ased \underline{B}ayesian Optimization (R2-B2)}
\label{r2b2}
Algorithm~\ref{BORR_attacker} describes the R2-B2 algorithm from the perspective of \emph{attacker} $\mathcal{A}$ which we will adopt in this section. Our R2-B2  algorithm for \emph{defender} $\mathcal{D}$ can be derived analogously. 
We will now discuss the recursive reasoning formalism of BO for $\mathcal{A}$'s action selection in step $2$ of Algorithm~\ref{BORR_attacker}.

\subsection{Recursive Reasoning Formalism of BO}
\label{subsec:recursive_reasoning_model}
\begin{figure}
    \centering
    \begin{subfigure}[t]{0.313\linewidth}
        \includegraphics[width=\linewidth]{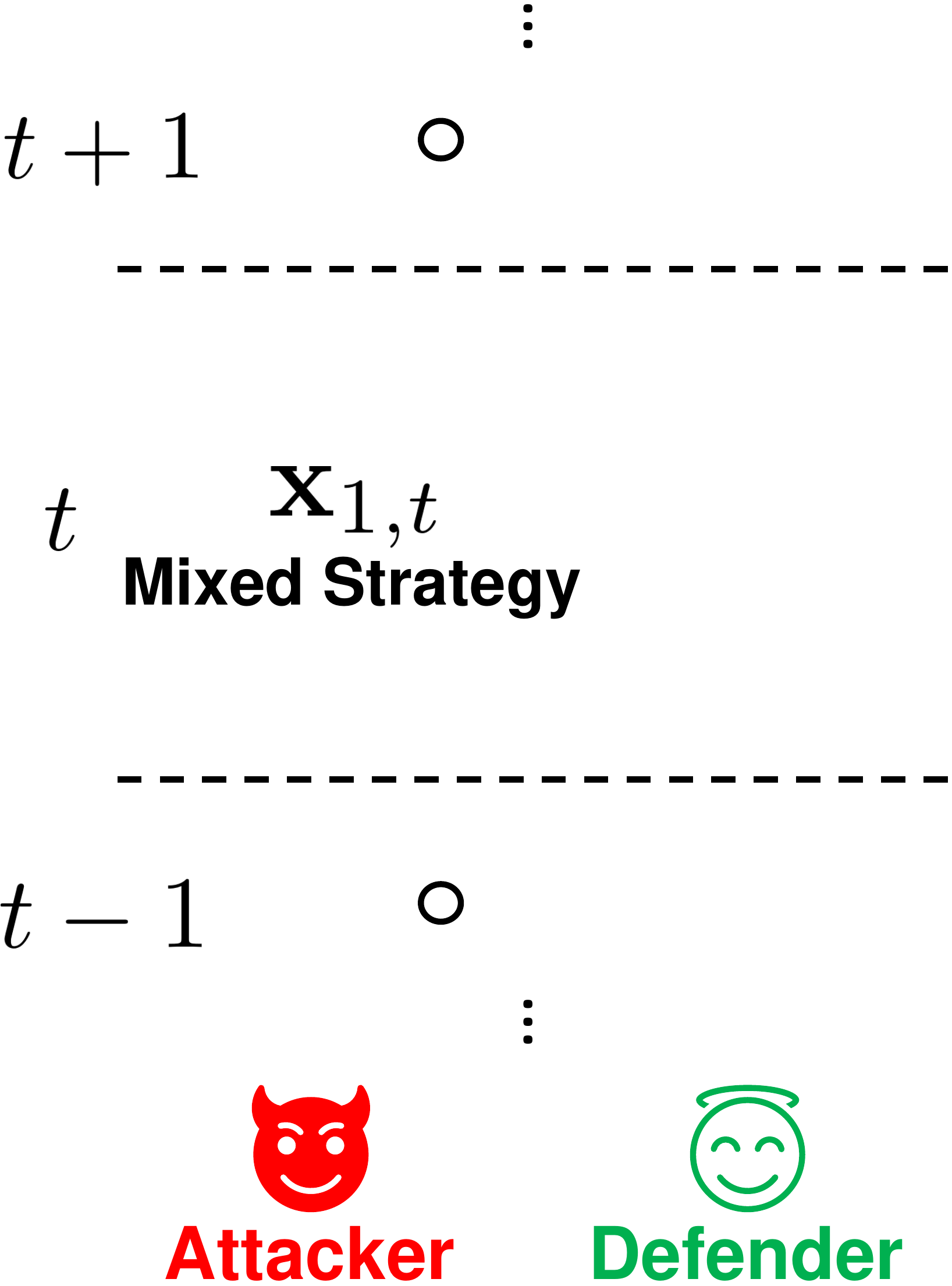}
        \caption{Level $0$}
    \end{subfigure}
    \begin{subfigure}[t]{0.313\linewidth}
        \includegraphics[width=\linewidth]{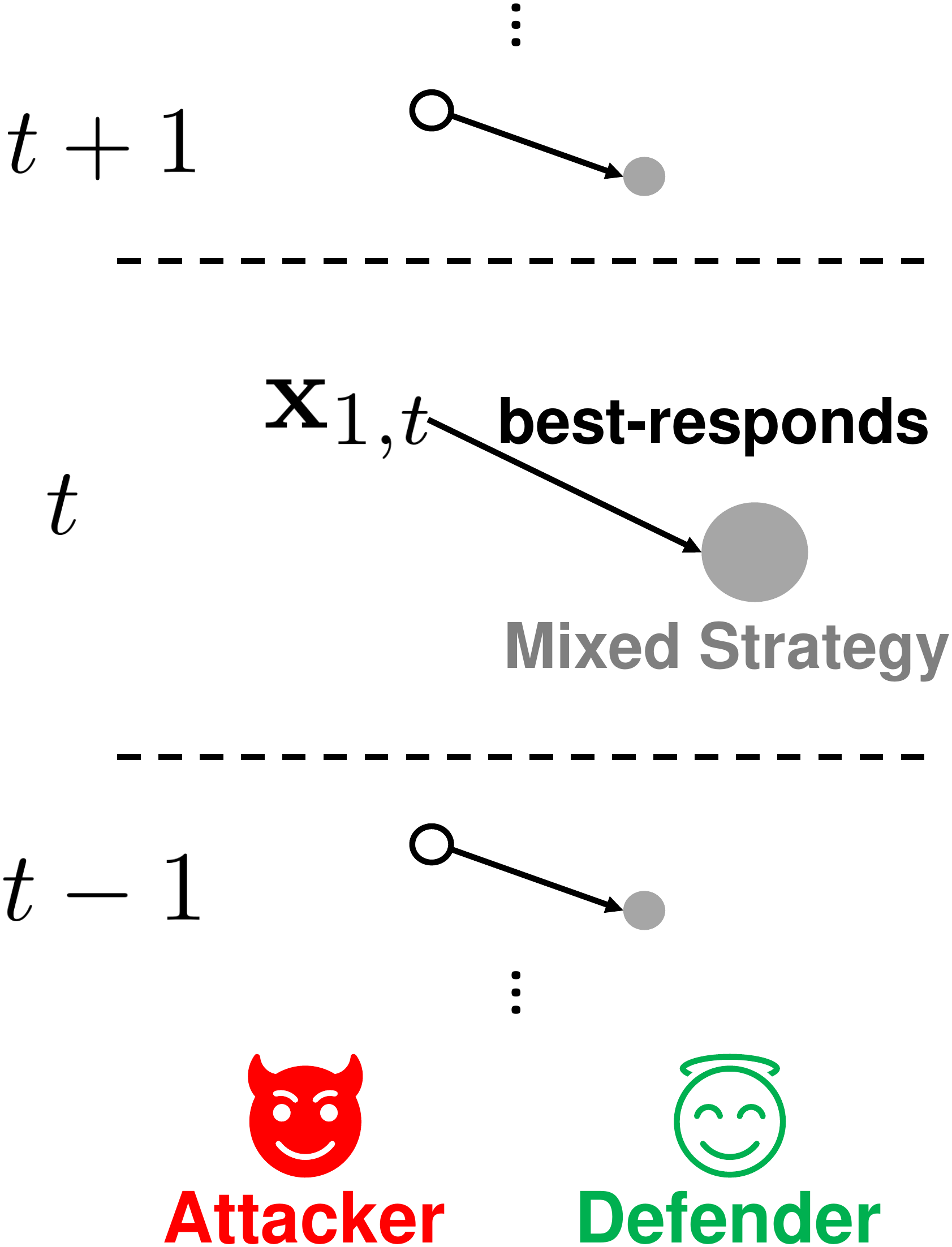}
        \caption{Level $1$}
    \end{subfigure}
    \begin{subfigure}[t]{0.313\linewidth}
        \includegraphics[width=\linewidth]{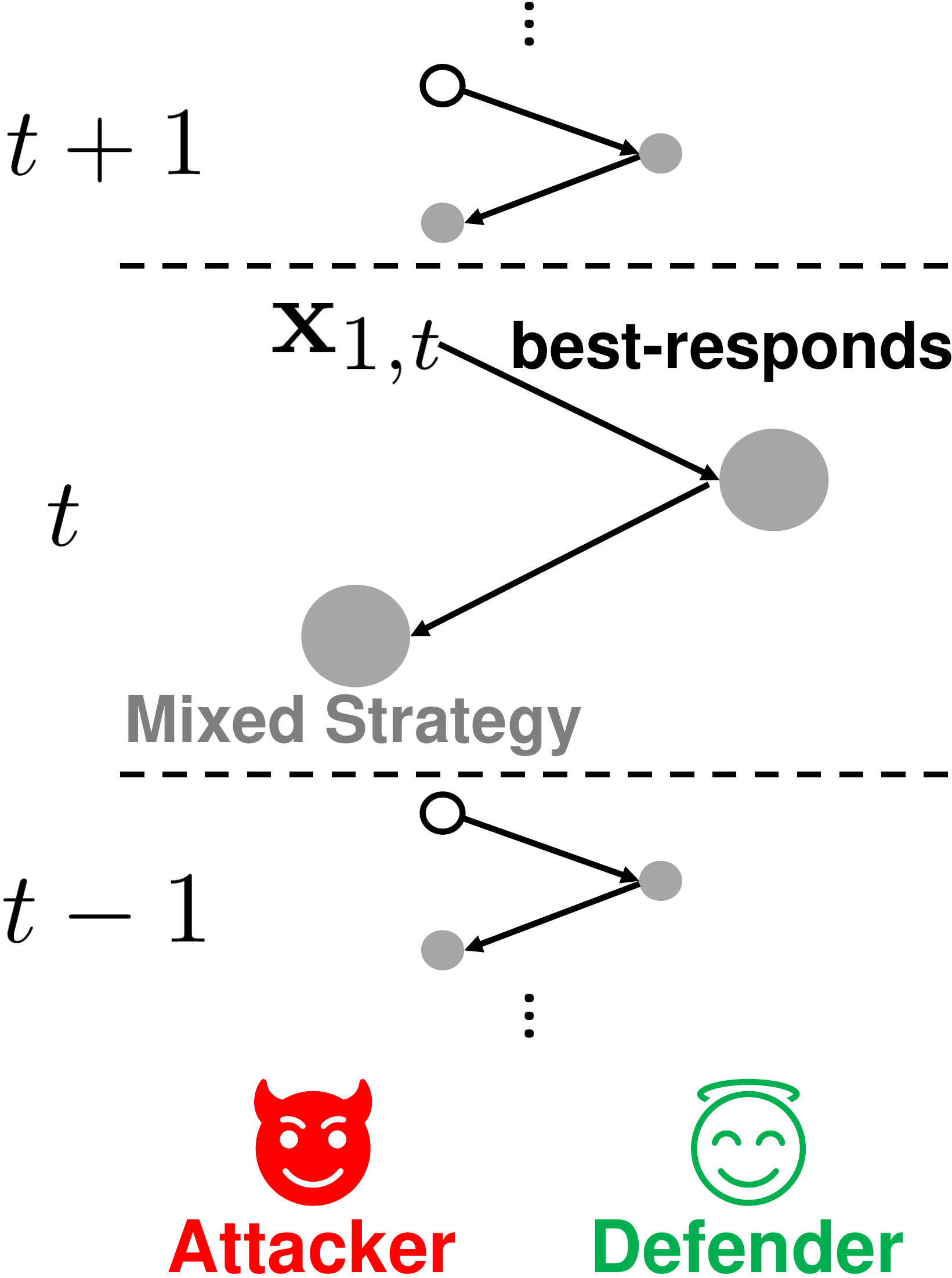}
        \caption{Level $2$}
    \end{subfigure}
    \caption{Illustration of attacker $\mathcal{A}$'s strategies of selecting its input action  
    from reasoning at levels $k=0$, $1$, and $2$.}
    \label{fig:illustration}
\end{figure}

Our recursive reasoning formalism of BO follows a similar principle as the cognitive hierarchy model~\cite{camerer2004cognitive}: 
At level $k=0$ of reasoning, $\mathcal{A}$ adopts some randomized/mixed strategy of selecting its action. At level $k\geq 1$ of reasoning, $\mathcal{A}$ best-responds to the strategy of $\mathcal{D}$ who is reasoning at a lower level.
Let $\mathbf{x}^{k}_{1,t}$ denote the input action $\mathbf{x}_{1,t}$ selected by $\mathcal{A}$'s strategy from reasoning at level $k$ in iteration $t$.
Depending on the (a) degree of knowledge about $\mathcal{D}$ and (b) available computational resource,
$\mathcal{A}$ can choose one of the following three types of strategies of selecting its action with varying levels of reasoning,
as shown in Fig.~\ref{fig:illustration}:

{\bf Level-$k=0$ Strategy.}
Without knowledge of $\mathcal{D}$'s level of reasoning nor its level-$0$ strategy, $\mathcal{A}$ by default can reason at level $0$ and play a mixed strategy $\mathcal{P}^{0}_{1,t}$ of selecting its action by sampling $\mathbf{x}^{0}_{1,t}$ from the probability distribution $\mathcal{P}^{0}_{1,t}$ over its input action space $\mathcal{X}_1$, as discussed in Section~\ref{subsec:level-0}.

{\bf Level-$k=1$ Strategy.}
If $\mathcal{A}$ thinks that $\mathcal{D}$ reasons at level $0$ and has knowledge of $\mathcal{D}$'s level-$0$ mixed strategy $\mathcal{P}^{0}_{2,t}$, 
then $\mathcal{A}$ can reason at level $1$ and play a pure strategy that best-responds to the level-$0$ strategy
of $\mathcal{D}$, as explained in Section~\ref{subsec:level_1_reasoning}.
Such a level-$1$ reasoning of $\mathcal{A}$ is general since it caters to \emph{any} level-$0$ strategy of $\mathcal{D}$ and hence does not require $\mathcal{D}$ to perform recursive reasoning.

{\bf Level-$k\geq 2$ Strategy.}
If $\mathcal{A}$ thinks that $\mathcal{D}$ reasons at level $k - 1$, 
then $\mathcal{A}$ can reason at level $k$ and play a pure strategy that best-responds to $\mathcal{D}$'s level-$(k-1)$ action, as detailed in Section~\ref{subsec:level_k_policy}. 
Different from the level-$1$ reasoning of $\mathcal{A}$, its level-$k$ reasoning assumes that $\mathcal{D}$'s level-$(k-1)$ action is derived using the same recursive reasoning process.
\begin{algorithm}[t]
\begin{algorithmic}[1]
	\FOR{$t=1,2,\ldots, T$}
	    \STATE Select input action $\mathbf{x}_{1,t}$ using its level-$k$ strategy (while defender $\mathcal{D}$ selects input action $\mathbf{x}_{2,t}$)
	    \STATE Observe noisy payoff $y_{1,t}=f_1(\mathbf{x}_{1,t},\mathbf{x}_{2,t}) + \epsilon_1$
        \STATE Update GP posterior belief using $\langle(\mathbf{x}_{1,t}, \mathbf{x}_{2,t}) , y_{1,t}\rangle$
	\ENDFOR
\end{algorithmic}
\caption{R2-B2 for attacker $\mathcal{A}$'s level-$k$ reasoning}
\label{BORR_attacker}
\end{algorithm}

\subsubsection{Level-$k=0$ Strategy}
\label{subsec:level-0}
Level $0$ is a conservative, default choice for $\mathcal{A}$ since it does not require \emph{any} knowledge about $\mathcal{D}$'s strategy of selecting its input action and is computationally lightweight.
At level $0$, $\mathcal{A}$ plays a mixed strategy $\mathcal{P}^{0}_{1,t}$ by sampling $\mathbf{x}^{0}_{1,t}$ from the probability distribution $\mathcal{P}^{0}_{1,t}$ over its input action space 
$\mathcal{X}_1$: $\mathbf{x}^{0}_{1,t} \sim \mathcal{P}^{0}_{1,t}$. 
A mixed/randomized strategy (instead of a pure/deterministic strategy) is considered because 
without knowledge of $\mathcal{D}$'s strategy,  
$\mathcal{A}$ has to treat $\mathcal{D}$ as a black-box adversary.
This setting corresponds to that of an \emph{adversarial bandit} problem in which any deterministic strategy suffers from linear worst-case regret~\cite{lattimore2018bandit} and 
\emph{randomization} alleviates this issue.
Such a randomized design of our level-$0$ strategies is consistent with that of the cognitive hierarchy model in which a level-$0$ thinker does not make any assumption about the other agent and
selects its action via a probability distribution without using strategic thinking~\cite{camerer2004cognitive}.
We will now present a few reasonable choices of level-$0$ mixed strategies.
However, in both theory (Theorems~\ref{theorem_level_1},~\ref{theorem_level_k} and~\ref{theorem_borr_lite}) and practice, \emph{any} strategy of action selection (including existing methods (Section~\ref{exp:comparison_parsimonious})) can be considered as a level-$0$ strategy.

In the simplest setting where $\mathcal{A}$ has no knowledge of $\mathcal{D}$'s strategy, 
a natural choice for its level-$0$ mixed strategy is \emph{random search}. 
That is, $\mathcal{A}$ samples its action from a uniform distribution over $\mathcal{X}_1$. 
An alternative choice is to use the \emph{EXP3 algorithm} for the adversarial linear bandit problem, which requires the GP to be transformed  via a random features approximation~\cite{rahimi2008random} into linear regression 
with random features as inputs. 
Since the regret of EXP3 algorithm is bounded from above by $\mathcal{O}(\sqrt{d'_1 T \log |\mathcal{X}_1|})$~\cite{lattimore2018bandit} where $d'_1$ denotes the number of random features,
it incurs sub-linear regret and can thus achieve asymptotic no regret.

In a more relaxed setting where $\mathcal{A}$ has access to the history of actions selected by $\mathcal{D}$, $\mathcal{A}$ can use the \emph{GP-MW algorithm}~\cite{sessa2019no} to derive its level-$0$ mixed strategy;
for completeness, GP-MW is briefly described in Appendix~\ref{app:gp_mw}.
The result below bounds the regret of $\mathcal{A}$ when using GP-MW for level-$0$ reasoning and   
its proof is slightly modified from that of~\citet{sessa2019no} to account for its payoff function $f_1$
being sampled from a GP (Section~\ref{background}): 
\begin{theorem}
\label{regret_gp}
Let $\delta\in (0,1)$, $\beta_t\triangleq 2\log (|\mathcal{X}_1|t^2\pi^2/(3\delta))$, and $\gamma_T$ denotes the maximum information gain about payoff function $f_1$ from any history of  actions selected by both agents and corresponding noisy payoffs observed by $\mathcal{A}$ up till iteration $T$.
Suppose that $\mathcal{A}$ uses GP-MW to derive its level-$0$ strategy. 
Then, with probability of at least $1 - \delta$,
$$
R_{1,T} = \mathcal{O}(\sqrt{T\log |\mathcal{X}_1|} + \sqrt{T\log (2/\delta)} + \sqrt{T \beta_T \gamma_T} )\ .
$$
\end{theorem}
From Theorem~\ref{regret_gp}, $R_{1,T}$ 
is sub-linear in $T$.\footnote{\label{gammat}The asymptotic growth of $\gamma_T$ has been analyzed for some commonly used kernels: $\gamma_T=\mathcal{O}((\log T)^{d_1+1})$ for squared exponential kernel and $\gamma_T=\mathcal{O}(T^{d_1(d_1+1)/(2\nu + d_1(d_1+1))}\log T)$ for Mat\'ern kernel with parameter $\nu>1$. For both kernels, the last term in the regret bound in Theorem~\ref{regret_gp} grows sub-linearly in $T$.} So, $\mathcal{A}$ using GP-MW for  level-$0$ reasoning achieves asymptotic no regret.

\subsubsection{Level-$k=1$ Strategy}
\label{subsec:level_1_reasoning}
If $\mathcal{A}$ thinks that $\mathcal{D}$ reasons at level $0$ and has knowledge of $\mathcal{D}$'s level-$0$ strategy $\mathcal{P}^{0}_{2,t}$, then $\mathcal{A}$ can reason at level $1$. 
Specifically, $\mathcal{A}$ selects its level-$1$ action $\mathbf{x}^{1}_{1,t}$ that maximizes the expected value of GP-UCB~\eqref{eqgpucb} w.r.t.~$\mathcal{D}$'s level-$0$ strategy:
\begin{equation}
\begin{array}{c}
    \mathbf{x}^{1}_{1,t}\triangleq\mathop{\arg\max}_{\mathbf{x}_1\in \mathcal{X}_1} \mathbb{E}_{\mathbf{x}^{0}_{2,t} \sim \mathcal{P}^{0}_{2,t}}[\alpha_{1,t}(\mathbf{x}_1, \mathbf{x}^{0}_{2,t})]\ .
\end{array}    
\label{eq:level_1_defender}
\end{equation}
If input action space $\mathcal{X}_2$ of $\mathcal{D}$ is discrete and 
not too large, then~\eqref{eq:level_1_defender} can be solved exactly. 
Otherwise,~\eqref{eq:level_1_defender} can be solved approximately via sampling from $\mathcal{P}^{0}_{2,t}$.
Such a level-$1$ reasoning of $\mathcal{A}$ to solve~\eqref{eq:level_1_defender} only requires access to the history of actions selected by $\mathcal{D}$ but not its observed payoffs, which is the same as that needed by GP-MW.
Our first main result (see its proof in Appendix~\ref{app:proof})  bounds the expected regret of $\mathcal{A}$ when using R2-B2 for level-$1$ reasoning:
\begin{theorem}
\label{theorem_level_1}
Let $\delta\in (0, 1)$ and $C_1\triangleq 8/\log(1+\sigma^{-2}_1)$.
Suppose that $\mathcal{A}$ uses R2-B2 (Algorithm~\ref{BORR_attacker}) for level-$1$ reasoning and $\mathcal{D}$ uses mixed strategy $\mathcal{P}^{0}_{2,t}$ for level-$0$ reasoning. 
Then, with probability of at least $1 - \delta$,
$\mathbb{E}[R_{1,T}] \leq \sqrt{C_1 T \beta_T \gamma_T}$ where the expectation is with respect to the history of actions selected and payoffs observed by $\mathcal{D}$.
\end{theorem}
It follows from Theorem~\ref{theorem_level_1} that $\mathbb{E}[R_{1,T}]$ is sublinear in $T$.\cref{gammat}
So, $\mathcal{A}$ using R2-B2 for level-$1$ reasoning achieves asymptotic no expected regret, which holds for \emph{any} level-$0$ strategy of $\mathcal{D}$ regardless of whether $\mathcal{D}$ performs recursive reasoning.\vspace{-1mm}

\subsubsection{Level-$k\geq2$ Strategy}\vspace{-0.3mm}
\label{subsec:level_k_policy}
If $\mathcal{A}$ thinks that $\mathcal{D}$ reasons at level $1$, then $\mathcal{A}$ can reason at level $2$ and select its level-$2$ action $\mathbf{x}^{2}_{1,t}$~\eqref{eq:level_2_defender} to best-respond to  level-$1$ action $\mathbf{x}^{1}_{2,t}$~\eqref{eq:level_1_attacker} selected by $\mathcal{D}$, the latter of which can be computed/simulated by $\mathcal{A}$ in a similar manner as~\eqref{eq:level_1_defender}:
\begin{equation}
\begin{array}{c}
    \mathbf{x}^{2}_{1,t}\triangleq\mathop{\arg\max}_{\mathbf{x}_1\in \mathcal{X}_1} \alpha_{1,t}(\mathbf{x}_1, \mathbf{x}^{1}_{2,t})\ ,
\end{array} 
\label{eq:level_2_defender}
\end{equation}
\begin{equation}
\begin{array}{c}
    \mathbf{x}^{1}_{2,t}\triangleq\mathop{\arg\max}_{\mathbf{x}_2\in \mathcal{X}_2} \mathbb{E}_{\mathbf{x}^{0}_{1,t} \sim \mathcal{P}^{0}_{1,t}}[\alpha_{2,t}(\mathbf{x}^{0}_{1,t}, \mathbf{x}_2)]\ .
\end{array} 
\label{eq:level_1_attacker}
\end{equation}
In the general case, if $\mathcal{A}$ thinks that $\mathcal{D}$ reasons at level $k-1\geq 2$,
then $\mathcal{A}$ can reason at level $k\geq 3$ and 
select its level-$k$ action $\mathbf{x}^{k}_{1,t}$~\eqref{eq:determ_best_response} that best-responds to level-$(k-1)$ action $\mathbf{x}^{k-1}_{2,t}$~\eqref{eq:determ_best_response_2} selected by $\mathcal{D}$:\vspace{-0.5mm}
\begin{equation}
\begin{array}{c}
    \mathbf{x}^{k}_{1,t}\triangleq\mathop{\arg\max}_{\mathbf{x}_1\in \mathcal{X}_1} \alpha_{1,t}(\mathbf{x}_1, \mathbf{x}^{k-1}_{2,t})\ ,
\end{array} 
\label{eq:determ_best_response}
\end{equation}
\begin{equation}
\begin{array}{c}
    \mathbf{x}^{k-1}_{2,t}\triangleq\mathop{\arg\max}_{\mathbf{x}_2\in \mathcal{X}_2} \alpha_{2,t}(\mathbf{x}^{k-2}_{1,t}, \mathbf{x}_2)\ .
\end{array} 
\label{eq:determ_best_response_2}
\end{equation}
Since 
$\mathcal{A}$ thinks that 
$\mathcal{D}$'s level-$(k-1)$ action $\mathbf{x}^{k-1}_{2,t}$~\eqref{eq:determ_best_response_2}
is derived using the same recursive reasoning process, $\mathbf{x}^{k-1}_{2,t}$ best-responds to level-$(k-2)$ action $\mathbf{x}^{k-2}_{1,t}$ selected by $\mathcal{A}$,
the latter of which in turn best-responds to level-$(k-3)$ action $\mathbf{x}^{k-3}_{2,t}$ selected by $\mathcal{D}$ and can be computed in the same way as~\eqref{eq:determ_best_response}.
This recursive reasoning process continues until it reaches the base case of the level-$1$ action selected by either 
(a) $\mathcal{A}$~\eqref{eq:level_1_defender} 
if $k$ is odd (in this case, recall from Section~\ref{subsec:level_1_reasoning} that $\mathcal{A}$ requires knowledge of $\mathcal{D}$'s level-$0$ strategy $\mathcal{P}^{0}_{2,t}$ to compute~\eqref{eq:level_1_defender}),
or (b) $\mathcal{D}$~\eqref{eq:level_1_attacker} if $k$ is even.
Note that $\mathcal{A}$ has to perform the computations made by $\mathcal{D}$ to derive $\mathbf{x}^{k-1}_{2,t}$~\eqref{eq:determ_best_response_2} as well as the computations to best-respond to $\mathbf{x}^{k-1}_{2,t}$ via~\eqref{eq:determ_best_response}.
Our next main result (see its proof in Appendix~\ref{app:proof}) bounds the regret of $\mathcal{A}$ when using R2-B2 for level-$k\geq 2$ reasoning:
\begin{theorem}
\label{theorem_level_k}
Let $\delta\in (0, 1)$.
Suppose that 
$\mathcal{A}$ and $\mathcal{D}$ use R2-B2 (Algorithm~\ref{BORR_attacker}) 
for level-$k\geq 2$ and level-$(k-1)$ reasoning, respectively. 
Then, with probability of at least $1 - \delta$,
$R_{1,T} \leq \sqrt{C_1 T \beta_T \gamma_T}$.\vspace{-1mm}
\end{theorem}
Theorem~\ref{theorem_level_k} reveals that $R_{1,T}$ grows sublinearly in $T$.\cref{gammat}
So, $\mathcal{A}$ using R2-B2 for level-$k\geq 2$ reasoning achieves asymptotic no regret   
regardless of $\mathcal{D}$'s level-$0$ strategy $\mathcal{P}^{0}_{2,t}$.
By comparing Theorems~\ref{regret_gp} and~\ref{theorem_level_k}, we can observe that if $\mathcal{A}$ uses GP-MW as its
%to derive $\mathcal{D}$'s 
level-$0$ strategy, then  
it can achieve faster asymptotic convergence to no regret by  using R2-B2 to reason at level $k\geq 2$ and one level higher than $\mathcal{D}$.
However, when $\mathcal{A}$ reasons at a higher level $k$, 
its computational cost grows due to an additional optimization of the GP-UCB acquisition function per increase in level of reasoning.
So, $\mathcal{A}$ is expected to favor reasoning at a lower level, which agrees with the observation in the work of~\citet{camerer2004cognitive} on the cognitive hierarchy model that humans usually reason at a level no higher than $2$.

\subsection{R2-B2-Lite}
\label{lite}
We also propose a computationally cheaper variant of R2-B2 for  level-$1$ reasoning called R2-B2-Lite at the expense of a weaker convergence guarantee. 
When using R2-B2-Lite for level-$1$ reasoning, instead of following~\eqref{eq:level_1_defender}, 
$\mathcal{A}$ selects its level-$1$ action $\mathbf{x}^{1}_{1,t}$ by sampling $\widetilde{\mathbf{x}}^{0}_{2,t}$ from level-$0$ strategy $\mathcal{P}^{0}_{2,t}$ of $\mathcal{D}$ and best-responding to this sampled action:
\begin{equation}
\begin{array}{c}
    \mathbf{x}^{1}_{1,t}\triangleq\mathop{\arg\max}_{\mathbf{x}_1 \in \mathcal{X}_1}\alpha_{1,t}(\mathbf{x}_1,\widetilde{\mathbf{x}}^{0}_{2,t})\ .
\end{array}    
\label{eq:borr_light_level_1}
\end{equation}
Our final main result (its proof is in Appendix~\ref{app:proof_borr_lite}) bounds the expected regret of $\mathcal{A}$ using R2-B2-Lite for level-$1$ reasoning:
\begin{theorem}
\label{theorem_borr_lite}
Let $\delta\in (0, 1)$.
Suppose that $\mathcal{A}$ uses R2-B2-Lite for level-$1$ reasoning and $\mathcal{D}$ uses mixed strategy $\mathcal{P}^{0}_{2,t}$ for level-$0$ reasoning. 
If the trace of the covariance matrix of $\mathbf{x}^{0}_{2,t} \sim\mathcal{P}^{0}_{2,t}$
is not more than $\omega_t$ for $t= 1,\ldots,T$, then with probability of at least $1-\delta$,
$
\mathbb{E}[R_{1,T}] = \mathcal{O}(\sum^T_{t=1}\sqrt{\omega_t} + \sqrt{T \beta_T \gamma_T})
$
where the expectation is with respect to the history of actions selected and payoffs observed by $\mathcal{D}$ as well as $\widetilde{\mathbf{x}}^{0}_{2,t}$ for $t=1,\ldots,T$.
\end{theorem}
From Theorem~\ref{theorem_borr_lite}, the expected regret bound tightens 
if $\mathcal{D}$'s level-$0$ mixed strategy $\mathcal{P}^{0}_{2,t}$ has a smaller variance for each dimension of input action $\mathbf{x}^{0}_{2,t}$.
As a result, the level-$0$ action $\widetilde{\mathbf{x}}^{0}_{2,t}$ of $\mathcal{D}$ that is sampled by  $\mathcal{A}$
tends to be closer to the true level-$0$ action $\mathbf{x}^{0}_{2,t}$ selected by $\mathcal{D}$. 
Then, $\mathcal{A}$ can select level-$1$ action $\mathbf{x}^{1}_{1,t}$ that best-responds to a more precise estimate $\widetilde{\mathbf{x}}^{0}_{2,t}$ of the level-$0$ action $\mathbf{x}^{0}_{2,t}$ selected by $\mathcal{D}$, hence improving its expected payoff.
Theorem~\ref{theorem_borr_lite} also reveals that $\mathcal{A}$ using R2-B2-Lite for level-$1$ reasoning achieves asymptotic no expected regret 
if the sequence $(\omega_t)_{t\in\mathbb{Z}^+}$ uniformly decreases to $0$ 
(i.e., $\omega_{t+1} < \omega_{t}$ for $t\in\mathbb{Z}^+$ and $\lim_{T \rightarrow \infty} \omega_{T} = 0$). 
Interestingly, such a sufficient condition for achieving asymptotic no expected regret has a natural and elegant interpretation in terms of the exploration-exploitation trade-off: 
This condition is satisfied if $\mathcal{D}$ uses a level-$0$ mixed strategy $\mathcal{P}^{0}_{2,t}$ with a decreasing variance for each dimension of input action $\mathbf{x}^{0}_{2,t}$, which 
corresponds to transitioning from exploration (i.e., 
a large variance results in a diffused $\mathcal{P}^{0}_{2,t}$
and hence many actions being sampled)
to exploitation (i.e., a small variance results in a peaked $\mathcal{P}^{0}_{2,t}$ and hence fewer actions being sampled).
\begin{figure}[t]
	\centering
	\begin{tabular}{cc}
		\hspace{-3mm} 
		\includegraphics[width=0.49\linewidth]{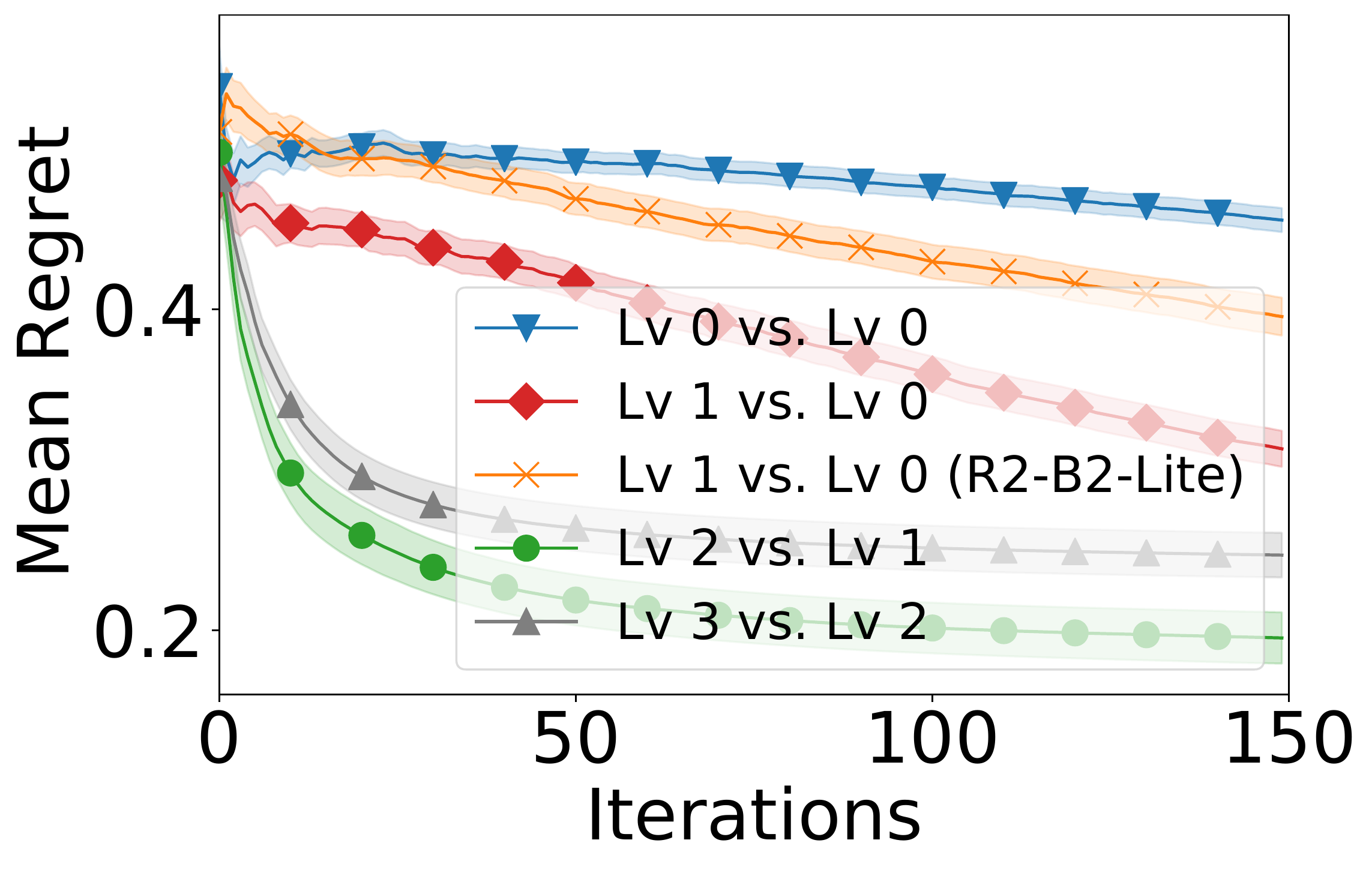} &  \hspace{-5mm} \includegraphics[width=0.49\linewidth]{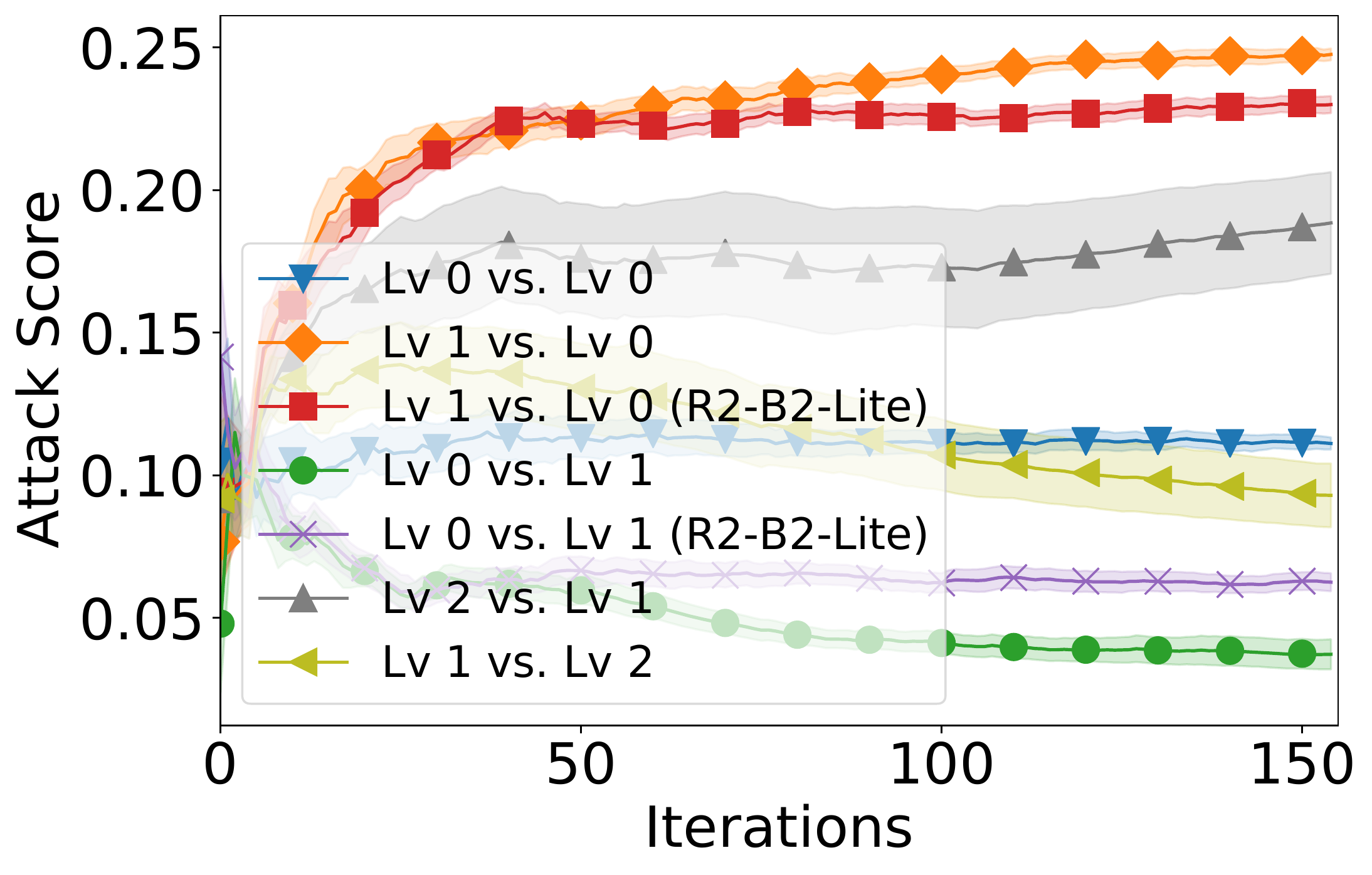} \\
		\hspace{-3mm} (a) common-payoff games & \hspace{-5mm} (d) random search \\
		\hspace{-3mm}
		\includegraphics[width=0.49\linewidth]{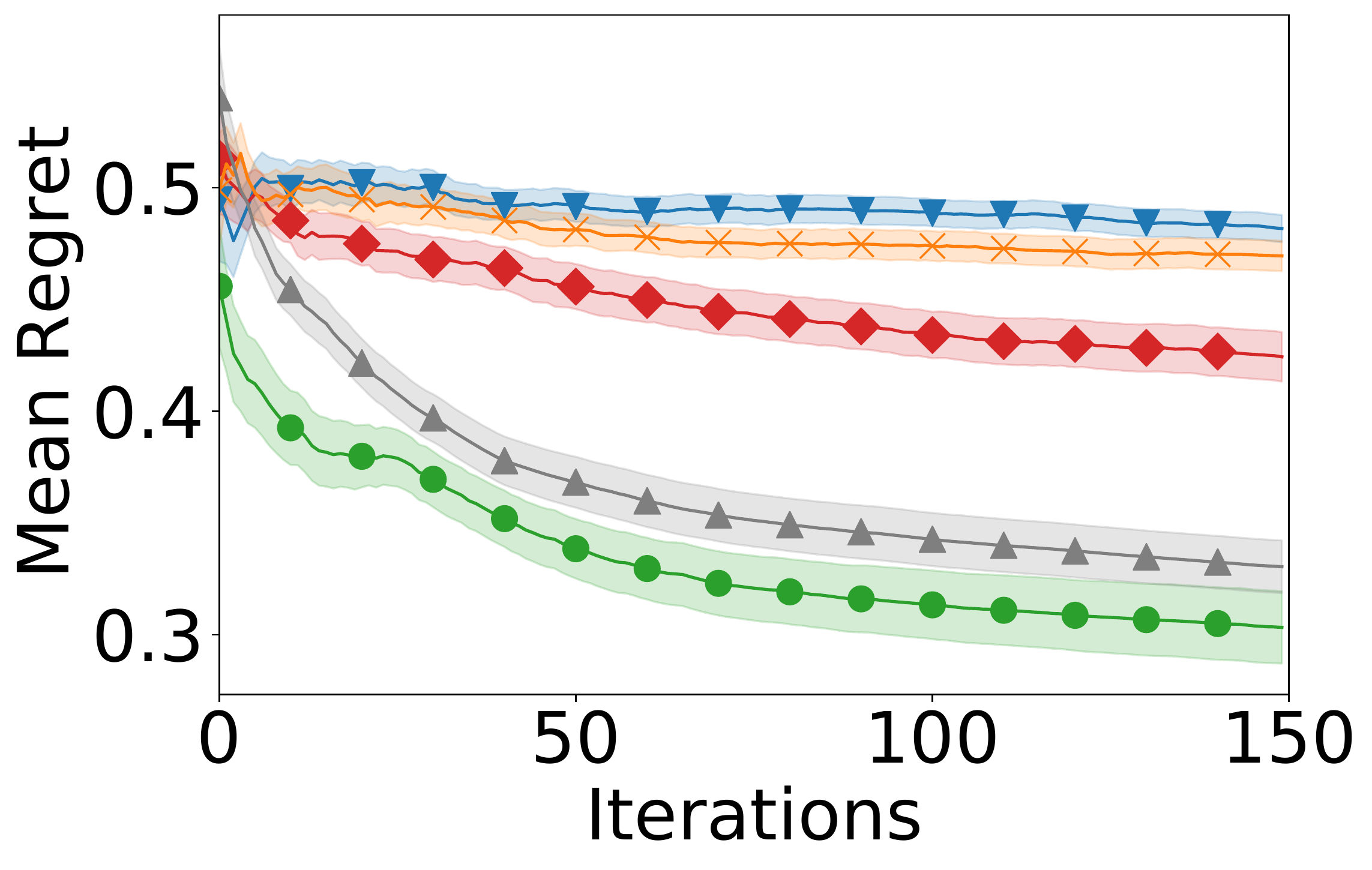} &    \hspace{-5mm} \includegraphics[width=0.49\linewidth]{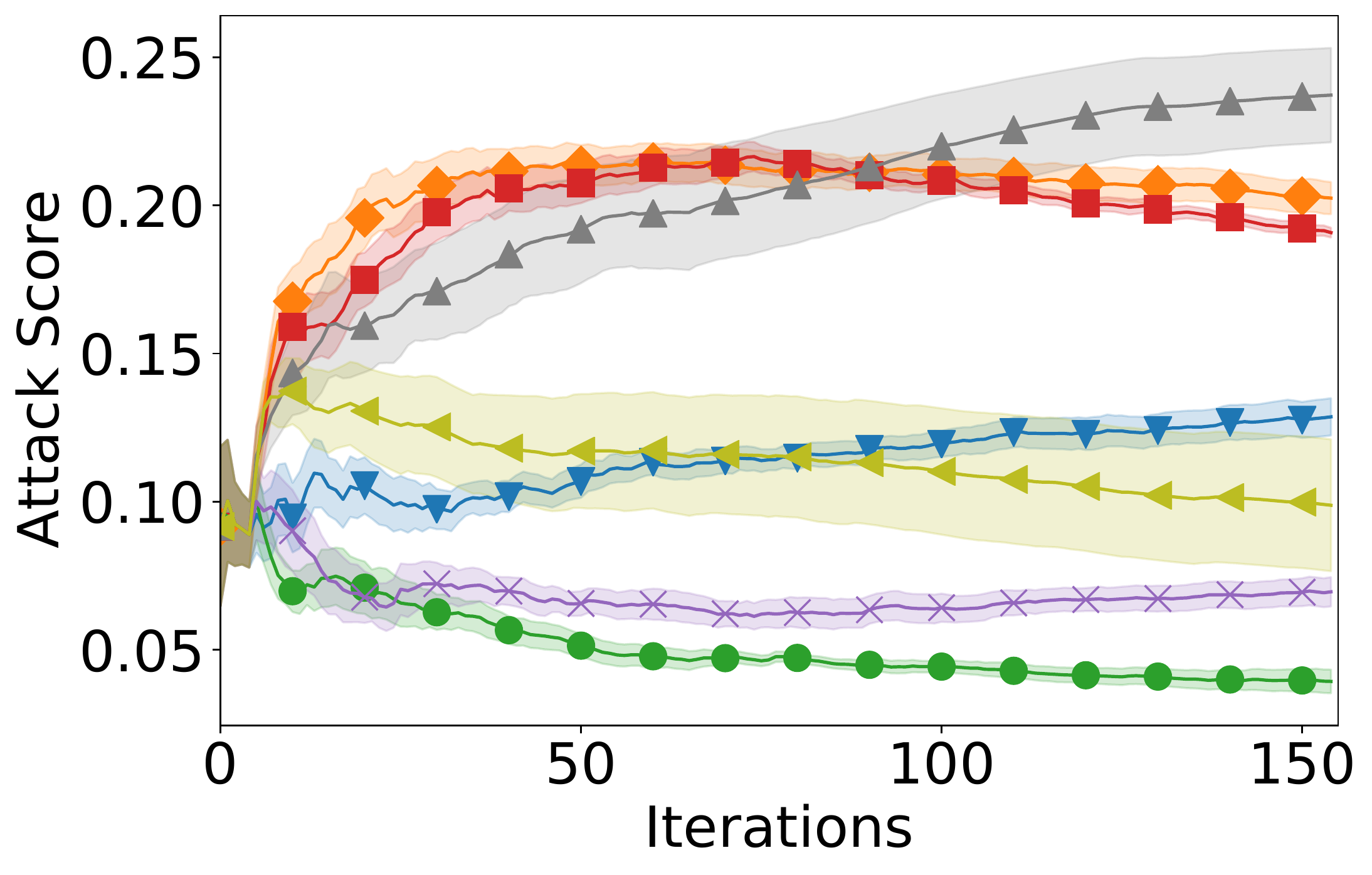}\\
		\hspace{-3mm} (b) general-sum games & \hspace{-5mm} (e) GP-MW\\
		\hspace{-3mm}
		\includegraphics[width=0.49\linewidth]{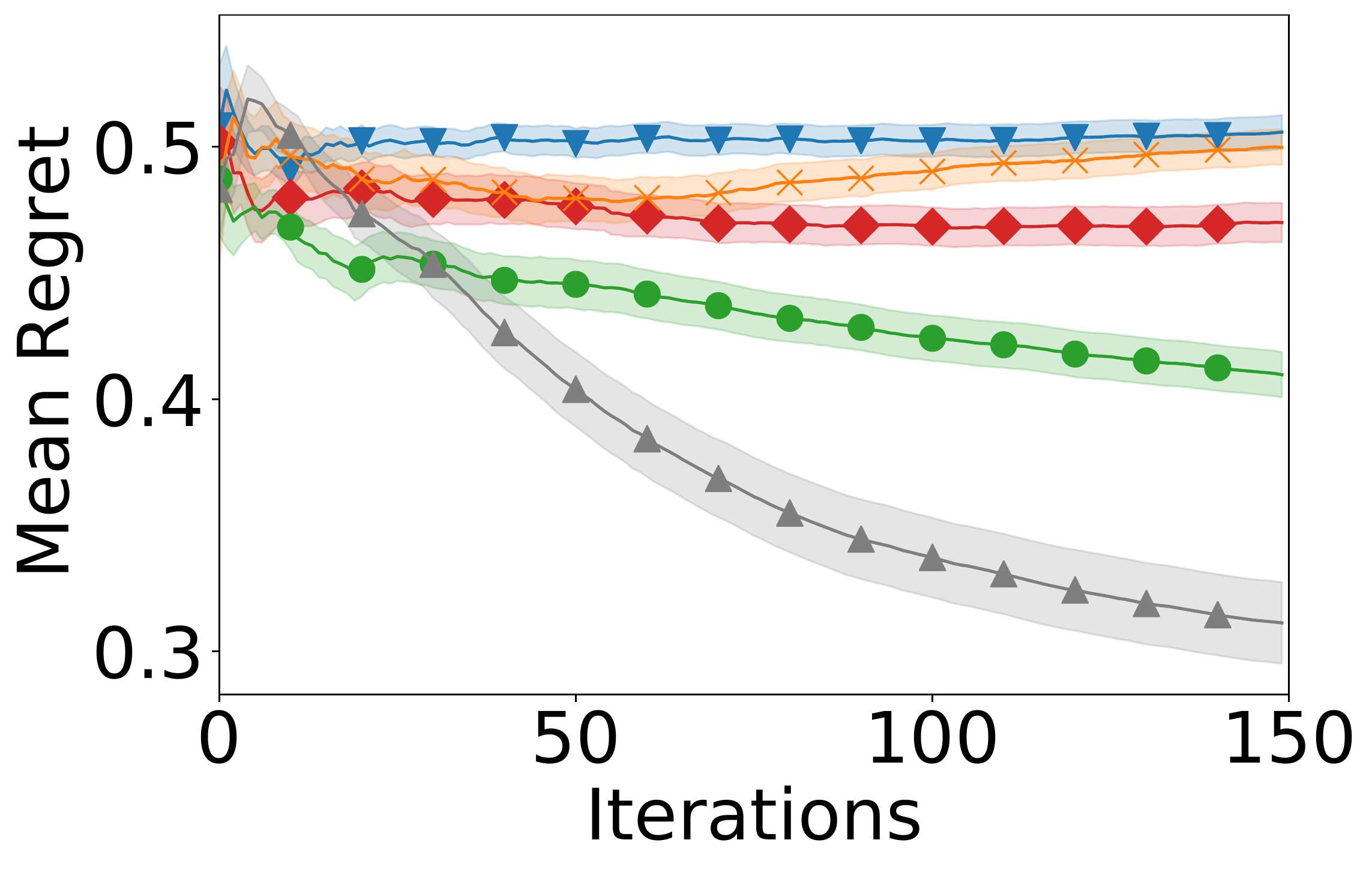} &    \hspace{-3mm} \includegraphics[width=0.49\linewidth]{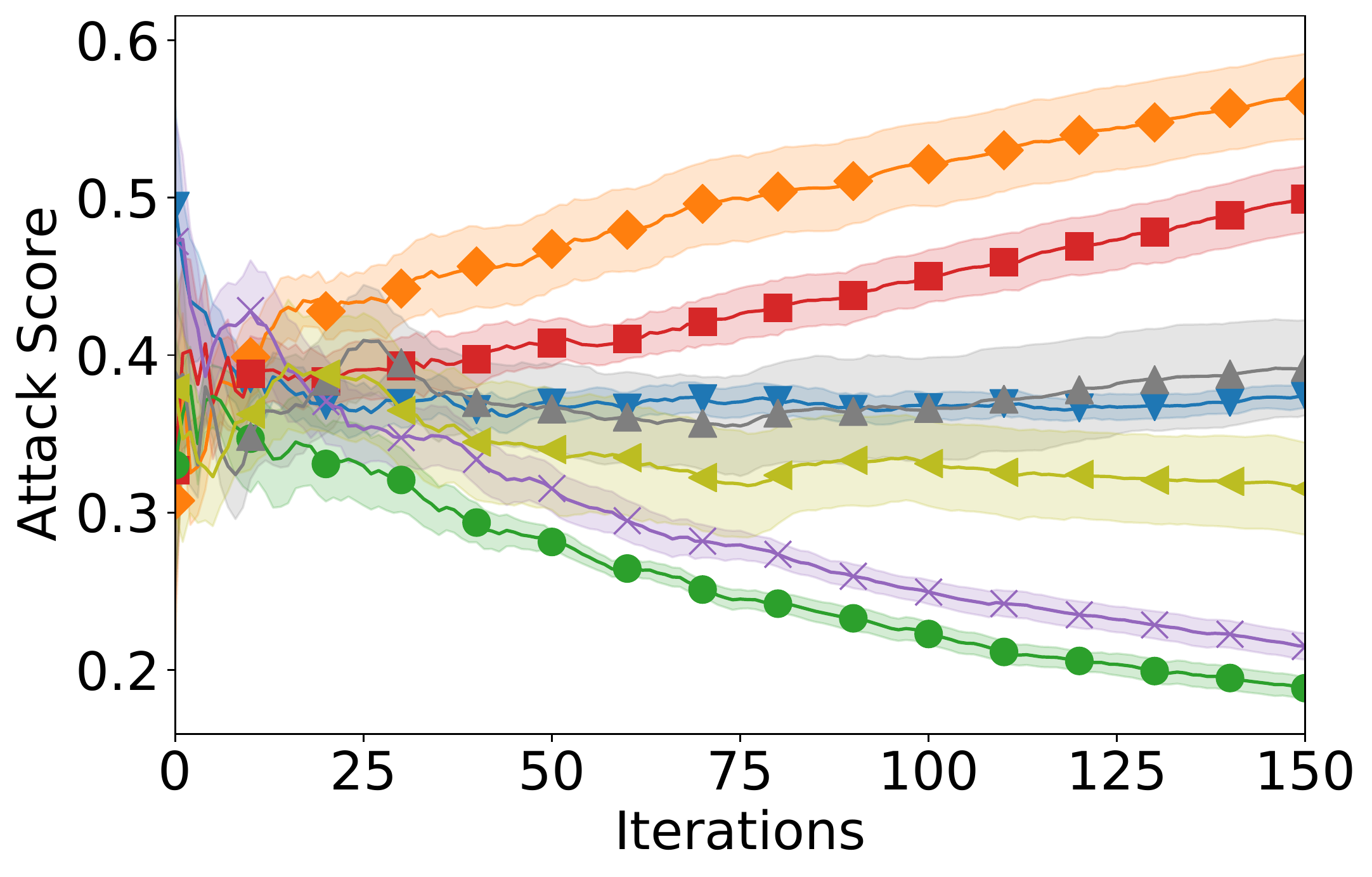}\\
		\hspace{-3mm} (c) constant-sum games & \hspace{-3mm} (f)
	\end{tabular}\vspace{-1mm}
	\caption{(a-c) Mean regret of agent $1$ in synthetic games where the legend in (a) represents the levels of reasoning of agents $1$ vs.~$2$. Attack score of $\mathcal{A}$ in adversarial ML for (d-e) MNIST and (f) CIFAR-10 datasets where the legend in (d) represents the levels of reasoning of $\mathcal{A}$ vs. $\mathcal{D}$.}
	\label{fig:player_1}\vspace{-1.9mm}
\end{figure}

\section{Experiments and Discussion}
\label{expt}
This section empirically evaluates the performance of our R2-B2 algorithm and demonstrates its generality using synthetic games, adversarial ML, and MARL.
Some of our experimental comparisons can be interpreted as comparisons with existing baselines used as level-$0$ strategies (Section~\ref{subsec:level-0}).
Specifically, we can compare the performance of our level-$1$ agent with that of a baseline method when they are against the same level-$0$ agent.
Moreover, in constant-sum games, we can perform a more direct comparison by playing our level-$1$ agent against an opponent using a baseline method as a level-$0$ strategy (Section~\ref{exp:comparison_parsimonious}). 
Additional experimental details and results are reported in Appendix~\ref{app:experiment} due to lack of space.
All error bars represent standard error.

\subsection{Synthetic Games}
\label{subsec:synth_func}
Firstly, we empirically evaluate the performance of R2-B2 using synthetic games with two agents whose payoff functions are sampled from GP over a discrete input domain.
Both agents use GP-MW and R2-B2/R2-B2-Lite for level-$0$ and level-$k\geq1$ reasoning, respectively.
We consider $3$ types of games: common-payoff, general-sum, and constant-sum games.
Figs.~\ref{fig:player_1}a to~\ref{fig:player_1}c show results of the mean regret\footnote{The mean regret $T^{-1}\sum^T_{t=1}(\max_{\mathbf{x}_{1}\in\mathcal{X}_1, \mathbf{x}_{2}\in\mathcal{X}_2} f_1(\mathbf{x}_{1}, \mathbf{x}_{2}) -f_1(\mathbf{x}_{1,t}, \mathbf{x}_{2,t}))$ of agent $1$ pessimistically estimates (i.e., upper bounds) $R_{1,T}/T$~\eqref{extregret} and is thus not expected to converge to $0$. Nevertheless, it serves as an appropriate performance metric here. 
}
of agent $1$ averaged over $10$ random samples of GP and $5$ initializations of $1$ randomly selected action with observed payoff per sample:
In all types of games, when agent $1$ reasons at one level higher than agent $2$, it incurs a smaller mean regret than when 
reasoning at level $0$ (blue curve), which demonstrates the performance advantage of recursive reasoning and corroborates our theoretical results (Theorems~\ref{theorem_level_1} and~\ref{theorem_level_k}).
The same can be observed for agent $1$ using R2-B2-Lite for level-$1$ reasoning (orange curve) but it does not perform as well as that using R2-B2 (red curve), 
which again agrees with our theoretical result (Theorem~\ref{theorem_borr_lite}).
Moreover, comparing the red (orange) and blue curves shows that when against the same level-$0$ agent, our R2-B2 (R2-B2-Lite) level-$1$ agent outperforms the baseline method of GP-MW (as a level-$0$ strategy).

Figs.~\ref{fig:player_1}a and~\ref{fig:player_1}c also reveal the effect of incorrect thinking of the level of reasoning of the other agent on its performance:
Since agent $2$ uses recursive reasoning at level $1$ or more, agent $2$ thinks that it is reasoning at one level higher than agent $1$. However, it is in fact reasoning at one level lower in these two figures.
In common-payoff games, since agents $1$ and $2$ have identical payoff functions, the mean regret of agent $2$ is the same as that of agent $1$ in Fig.~\ref{fig:player_1}a.
So, from agent $2$'s perspective, it benefits from such an incorrect thinking in common-payoff games.
In constant-sum games, since the payoff function of agent $2$ is negated from that of agent $1$,
the mean regret of agent $2$ increases with 
a decreasing mean regret of agent $1$ in Fig.~\ref{fig:player_1}c. 
So, from agent $2$'s viewpoint, it hurts from such an incorrect thinking in constant-sum games.
Further experimental results on such incorrect thinking are reported
in Appendix~\ref{app:exp_synth_two_player}b.

An intriguing observation from Figs.~\ref{fig:player_1}a to~\ref{fig:player_1}c is that 
when agent $1$ reasons at level $k\geq 2$,
it incurs a smaller mean regret than when reasoning at level $1$.
A possible explanation is 
that when agent $1$ reasons at level $k\geq 2$, 
its selected level-$k$ action~\eqref{eq:determ_best_response} best-responds to the actual level-$(k-1)$ action~\eqref{eq:determ_best_response_2} selected by agent $2$.
In contrast, when agent $1$ reasons at level $1$, its selected level-$1$ action~\eqref{eq:level_1_defender} maximizes the \emph{expected} value of GP-UCB w.r.t.~agent $2$'s level-$0$ \emph{mixed} strategy rather than the actual level-$0$ action selected by agent $2$.
However, as we shall see in the experiments on adversarial ML in Section~\ref{subsec:adv_ml}, 
when the expectation in level-$1$ reasoning~\eqref{eq:level_1_defender} needs to be approximated via sampling but insufficient samples are used,
the performance of level-$k\geq 2$ reasoning can be potentially diminished due to propagation of the approximation error from level $1$.

Moreover, Fig.~\ref{fig:player_1}c shows another interesting observation that is unique for constant-sum games: Agent $1$ achieves a significantly better performance when reasoning at level $3$ (i.e., agent $2$ reasons at level $2$) than at level $2$ (i.e., agent $2$ reasons at level $1$).
This can be explained by the fact that when agent $2$ reasons at level $2$, it best-responds to the level-$1$ action of agent $1$, which is most likely different from the actual action selected by agent $1$ since agent $1$ is in fact reasoning at level $3$. In contrast, when agent $2$ reasons at level $1$, instead of best-responding to a single (most likely wrong) action of agent $1$, it best-responds to the expected behavior of agent $1$ by attributing a distribution over all actions of agent $1$.
As a result, 
%given that agent $2$ reasons at one level \emph{lower} than agent $1$, 
agent $2$ suffers from a smaller performance deficit when reasoning at level $1$ (i.e., agent $1$ reasons at level $2$) compared with reasoning at level $2$ (i.e., agent $1$ reasons at level $3$) or higher. Therefore, agent $1$ obtains a more dramatic performance advantage when reasoning at level $3$ (gray curve) due to the constant-sum nature of the game.
A deeper implication of this insight is that although level-$1$ reasoning may not yield a better performance than level-$k\geq2$ reasoning as analyzed in the previous paragraph, it is more robust against incorrect estimates of the opponent's level of reasoning in constant-sum games.

Experimental results on the use of random search and EXP3 (Section~\ref{subsec:level-0}) for level-$0$ reasoning (instead of GP-MW) are reported in Appendix~\ref{app:exp_synth_two_player}c; the resulting 
observations and insights are consistent with those presented here.
This demonstrates the robustness of R2-B2 and corroborates the generality of our theoretical results (Theorems~\ref{theorem_level_1} and~\ref{theorem_level_k}) which hold for any level-$0$ strategy of the other agent.
We have also performed experiments using synthetic games involving \emph{more than two} agents (Appendix~\ref{app:exp_multi_player_games}),
which yield some interesting observations that are consistent with our theoretical analysis.

\subsection{Adversarial Machine Learning (ML)}
\subsubsection{R2-B2 for Adversarial ML}
\label{subsec:adv_ml}
We apply our R2-B2 algorithm to black-box adversarial ML for image classification problems with \emph{deep neural networks} (DNNs) using the MNIST and CIFAR-$10$ image datasets.
We consider \emph{evasion attacks}: The attacker $\mathcal{A}$ perturbs a test image to fool a fully trained DNN (referred to as the \emph{target ML model} hereafter) into misclassifying the image,
while the defender $\mathcal{D}$ transforms the perturbed image with the goal of ensuring the correct prediction by the classifier.
To improve query efficiency, dimensionality reduction techniques such as autoencoders have been commonly used for black-box adversarial attacks~\cite{tu2019autozoom}.
In our experiments, \emph{variational autoencoders} (VAE)~\cite{kingma2013auto} are used by both $\mathcal{A}$ and $\mathcal{D}$ to project the images to a lower-dimensional space (i.e., $2$D for MNIST and $8$D for CIFAR-$10$).\footnote{We have detailed in Appendix~\ref{app:adv_ml}a how VAE can be realistically incorporated into our algorithm.}
Following a common practice in adversarial ML, we focus on perturbations with bounded infinity norm as actions of $\mathcal{A}$ and $\mathcal{D}$:   
The maximum allowed perturbation to each pixel added by either $\mathcal{A}$ or $\mathcal{D}$ is no more than a pre-defined value $\epsilon$ where $\epsilon=0.2$  for MNIST and $\epsilon=0.05$ for CIFAR-$10$.
We consider \emph{untargeted attacks} whereby the goal of $\mathcal{A}$ ($\mathcal{D}$) is to cause (prevent) misclassification by the target ML model.
So, the payoff function of $\mathcal{A}$ is the maximum predictive probability among all incorrect classes (referred to as \emph{attack score} hereafter) and its negation is the payoff function of $\mathcal{D}$.
As a result, the application of R2-B2 to black-box adversarial ML represents a \emph{constant-sum game}.
An attack is considered \emph{successful} if the attack score is larger than the predictive probability of the correct class, hence resulting in misclassification of the test image. 
Both $\mathcal{A}$ and $\mathcal{D}$ use GP-MW/random search\footnote{For CIFAR-$10$ dataset, $\mathcal{A}$ uses only random search for level-$0$ reasoning due to high dimensions, as explained in Appendix~\ref{app:adv_ml}a.} and R2-B2/R2-B2-Lite for level-$0$ and level-$k\geq1$ reasoning, respectively.

Figs.~\ref{fig:player_1}d to~\ref{fig:player_1}f show results of
the attack score of $\mathcal{A}$ in adversarial ML 
for both image datasets while Table~\ref{table:mnist_cifar} shows results of the number of successful attacks by $\mathcal{A}$ over $150$ iterations of the game; the results are averaged over $10$ initializations of $5$ randomly selected actions with observed payoffs.\footnote{The results here use a test image from each dataset that can clearly illustrate the effects of both attack and defense. 
Refer to Appendix~\ref{app:adv_ml}b for more details and results using more test images; the observations are consistent with those presented here.}
It can be observed from Figs.~\ref{fig:player_1}d to~\ref{fig:player_1}f that 
when $\mathcal{A}$ reasons at one level higher than $\mathcal{D}$ (orange, red, and gray curves), its attack score is higher than when
reasoning at level $0$ (blue, green, and purple curves).
Similarly, when $\mathcal{D}$ reasons at one level higher (green, purple, and yellow curves), the attack score of $\mathcal{A}$ is reduced.
These observations demonstrate the performance advantage of using recursive reasoning in adversarial ML.
Such an advantage of recursive reasoning can also be seen from Table~\ref{table:mnist_cifar}:
For MNIST, when random search is used for level-$0$ reasoning and $\mathcal{A}$ reasons at one level higher than $\mathcal{D}$, it achieves a larger number of successful attacks ($12.8$, $10.2$, and $3.0$) than when reasoning at level $0$ ($2.6$, $0.8$, and $1.8$).
Similarly, when $\mathcal{D}$ reasons at one level higher, it reduces the number of successful attacks by $\mathcal{A}$ ($0.8$, $1.8$, and $0.9$)
than when reasoning at level $0$ ($2.6$, $12.8$, and $10.2$).
The observations are similar for MNIST with GP-MW for level-$0$ reasoning as well as for CIFAR-$10$ (Table~\ref{table:mnist_cifar}).

The performance advantage of $\mathcal{A}$ reasoning at level $2$ is observed to be smaller than that at level $1$; this may be explained by the propagation of error of approximating the expectation in level-$1$ reasoning~\eqref{eq:level_1_defender}, as explained previously in Section~\ref{subsec:synth_func}.
We investigate and report the effect of the number of samples for such an approximation in Appendix~\ref{app:adv_ml}c, which reveals that the performance improves with more samples, albeit with higher computational cost.
Moreover, some insights can also be drawn regarding the consequence of an incorrect thinking about the opponent's level of reasoning in constant-sum games.
For example, for the gray curves in Figs.~\ref{fig:player_1}d to~\ref{fig:player_1}f, $\mathcal{D}$ reasons at level $1$ because it thinks that $\mathcal{A}$ reasons at level $0$.
However, $\mathcal{A}$ is in fact reasoning at level $2$. As a result, in this constant-sum game, $\mathcal{D}$'s incorrect thinking about the opponent's level of reasoning negatively impacts
$\mathcal{D}$'s performance since the attack scores are increased.
This is consistent with the corresponding analysis in synthetic games regarding the effect of incorrect thinking about the level of reasoning of the other agent (Section~\ref{subsec:synth_func}).
\begin{table}
\vspace{-2mm}
	\scriptsize
	\caption{Average number of successful attacks by $\mathcal{A}$ over $150$ iterations in adversarial ML for MNIST and CIFAR-$10$ datasets where
	 the levels of reasoning are in the form of $\mathcal{A}$ vs. $\mathcal{D}$.}
	\centering
	\begin{tabular}{c|ccc}
	\hline
	Levels of reasoning & MNIST (random)\hspace{-1mm} & MNIST (GP-MW)\hspace{-1mm} & CIFAR-$10$\\
	\hline
	$0$ vs. $0$ & $2.6$ & $4.3$ & $70.1$ \\
	$1$ vs. $0$ & $12.8$ & $6.0$ & $113.1$ \\
	$1$ vs. $0$  (R2-B2-Lite) & $10.2$ & $6.8$ & $99.7$ \\
	 $0$ vs. $1$   & $0.8$  & $0.4$ & $25.2$ \\
	  $0$ vs. $1$  (R2-B2-Lite)  & $1.8$ & $1.0$ & $29.7$ \\
	$2$ vs. $1$   & $3.0$ & $5.2$ & $70.9$ \\
	$1$ vs. $2$  & $0.9$ & $0.4$ & $54.0$\\
	\hline
	\end{tabular}
	\label{table:mnist_cifar}\vspace{-4mm}	
\end{table}
\subsubsection{Comparison with State-of-the-art Adversarial Attack Methods}
\label{exp:comparison_parsimonious}
It was mentioned in Section~\ref{subsec:recursive_reasoning_model} that 
our theoretical results hold for
\emph{any}
level-$0$ strategy of the other agent.
So, any existing adversarial attack (defense) method can be used the level-$0$ strategy of $\mathcal{A}$ ($\mathcal{D}$).
In this experiment, we perform a direct comparison of R2-B2 with the state-of-the-art black-box adversarial attack method called \emph{Parsimonious}~\cite{moon2019parsimonious}: We use Parsimonious as the level-$0$ strategy of $\mathcal{A}$ and let $\mathcal{D}$ use R2-B2 for level-$1$ reasoning.
We consider a realistic setting where in each iteration, 
$\mathcal{D}$ only needs to receive the image perturbed by $\mathcal{A}$ and choose its action that best-responds to this perturbed image.
In this manner, $\mathcal{D}$ naturally has access to the history of actions selected by $\mathcal{A}$ (as required by \emph{perfect monitoring} in our repeated game) since it receives all images perturbed by $\mathcal{A}$.
Additional details of the experimental setting are reported in Appendix~\ref{app:adv_ml_parsimonious}a.

\begin{figure}
    \centering
    \includegraphics[width=1\linewidth]{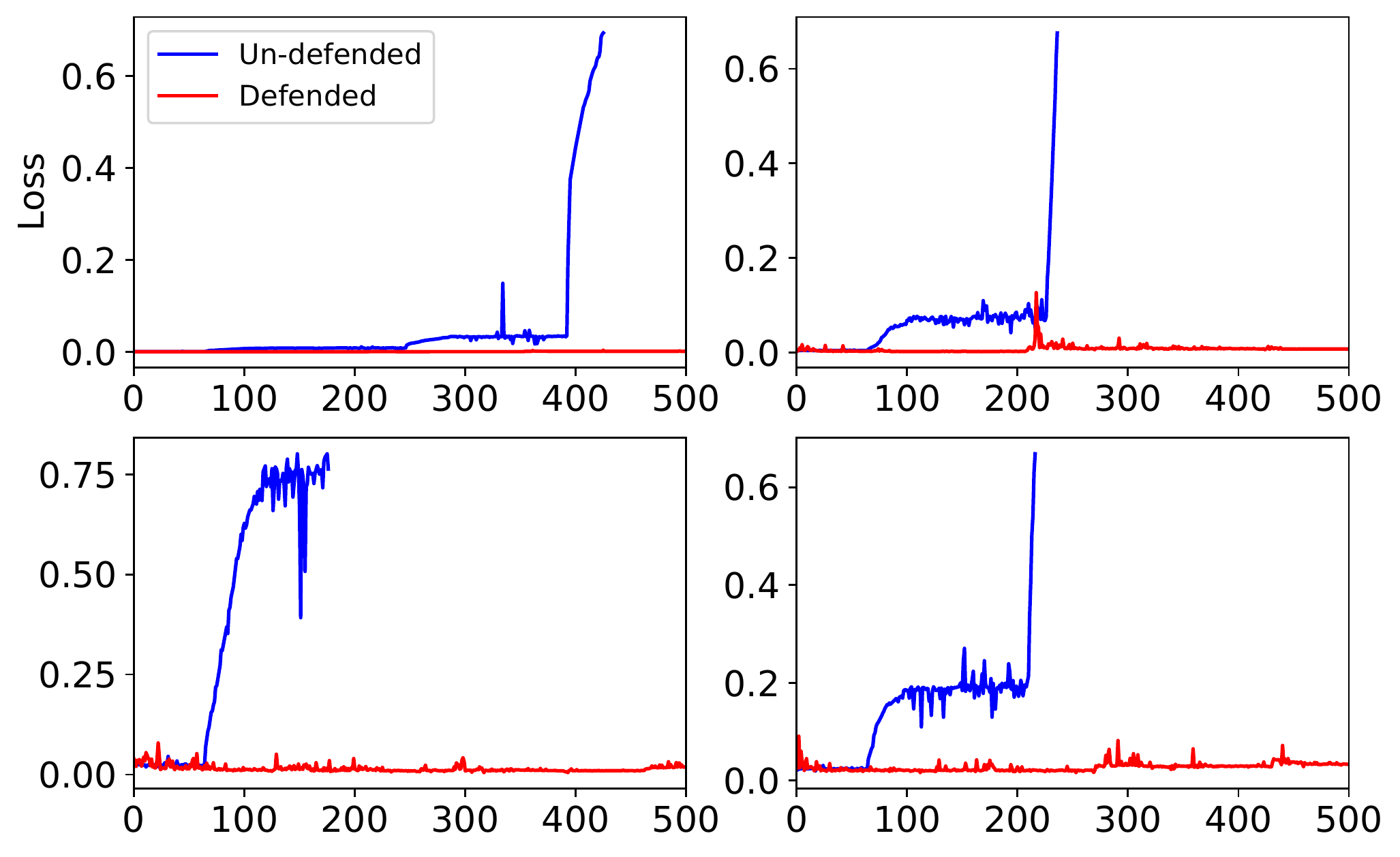}\vspace{-2.6mm}
    \caption{Loss incurred by Parsimonious with and without our level-$1$ R2-B2 defender on $4$ randomly selected images that are successfully attacked by Parsimonious.}
    \label{fig:parsimonious_some}\vspace{-4mm}
\end{figure}

We randomly select $70$ images from the CIFAR-$10$ dataset that are successfully attacked by Parsimonious using $\epsilon=0.05$ over $500$ iterations without the defender $\mathcal{D}$.\footnote{Compared to the work of~\citet{moon2019parsimonious}, we use fewer iterations and a larger $\epsilon$,
which we think is more realistic as attacks with an excessively large no.~of queries may be easily detected.}
Our level-$1$ R2-B2 defender manages to \emph{completely prevent any successful attacks} for $53$ of these images and requires Parsimonious to use \emph{more than $3.5$ times} more queries on average to succeed for $10$ other images.\footnote{The remaining $7$ images are so easy to attack such that the attacks are already successful during the initial exploration phase of our level-$1$ R2-B2 defender.}
Fig.~\ref{fig:parsimonious_some} shows results of the loss incurred by Parsimonious  (i.e., its original attack objective) with and without our level-$1$ R2-B2 defender for $4$ of the successfully defended images; results for other images are shown in Appendix~\ref{app:adv_ml_parsimonious}a.
This experiment not only demonstrates the generality of our R2-B2 algorithm, but can also be of significant independent interest to the adversarial ML community
as a defense method 
against black-box adversarial attacks.

In addition, as another comparison, we use the same experimental setting with the CIFAR-10 dataset in Section~\ref{subsec:adv_ml} and play Parsimonious against a level-$0$ defender using random search.
The results show that when against the same level-$0$ defender, Parsimonious achieves a significantly smaller average number of successful attacks (27.6) compared with our level-$1$ attacker (113.1, as shown in Table~\ref{table:mnist_cifar}). In other words, our level-$1$ defender can defend effectively against Parsimonious, while our level-$1$ attacker can attack better than Parsimonious.
Note that the unsatisfactory performances of Parsimonious in our experiments might be largely explained the fact that it does not consider the presence of a defender.
Moreover, our level-$1$ R2-B2 defender can also defend against black-box adversarial attacks from standard BO algorithms (Appendix~\ref{app:adv_ml_parsimonious}b)\footnote{The BO attacker here only takes its perturbations as inputs and thus does not consider the defender.}, which have become popular recently~\cite{ru2020bayesopt}.

\subsection{Multi-Agent Reinforcement Learning (MARL)}
We apply R2-B2 to policy search for MARL with \emph{more than two} agents.
Each action of an agent represents a particular set of policy parameters controlling the behavior of the agent in an environment.
The payoff to each agent corresponding to a selected set of its policy parameters (i.e., action) is its mean return (i.e., cumulative reward) from the execution of all the agents’ selected policies across $5$ independent episodes.
Since the agents interact in the environment, the payoff function of each agent  depends on the policies (actions) selected by all agents.
We use the predator-prey game from the widely used multi-agent particle environment in~\cite{lowe2017multi}. 
This $3$-agent game (see Fig.~\ref{fig:simple_tag_illu} in Appendix~\ref{app:marl}) contains two predators who are trying to catch a prey.
The prey is rewarded for being far from the predators and penalized for stepping outside the boundary.
The two predators have identical payoff functions and are rewarded for being close to the prey (if the prey stays within the boundary).
So, the predator-prey game represents a \emph{general-sum game}.
All agents use random search\footnote{All agents use only random search for level-$0$ reasoning due to high dimensions, as explained in Appendix~\ref{app:marl}.} and R2-B2 for level-$0$ and level-$k\geq 1$ reasoning, respectively.

\begin{figure}
	\centering
	\begin{tabular}{cc}
        \hspace{-3mm}\includegraphics[width=0.5\linewidth]{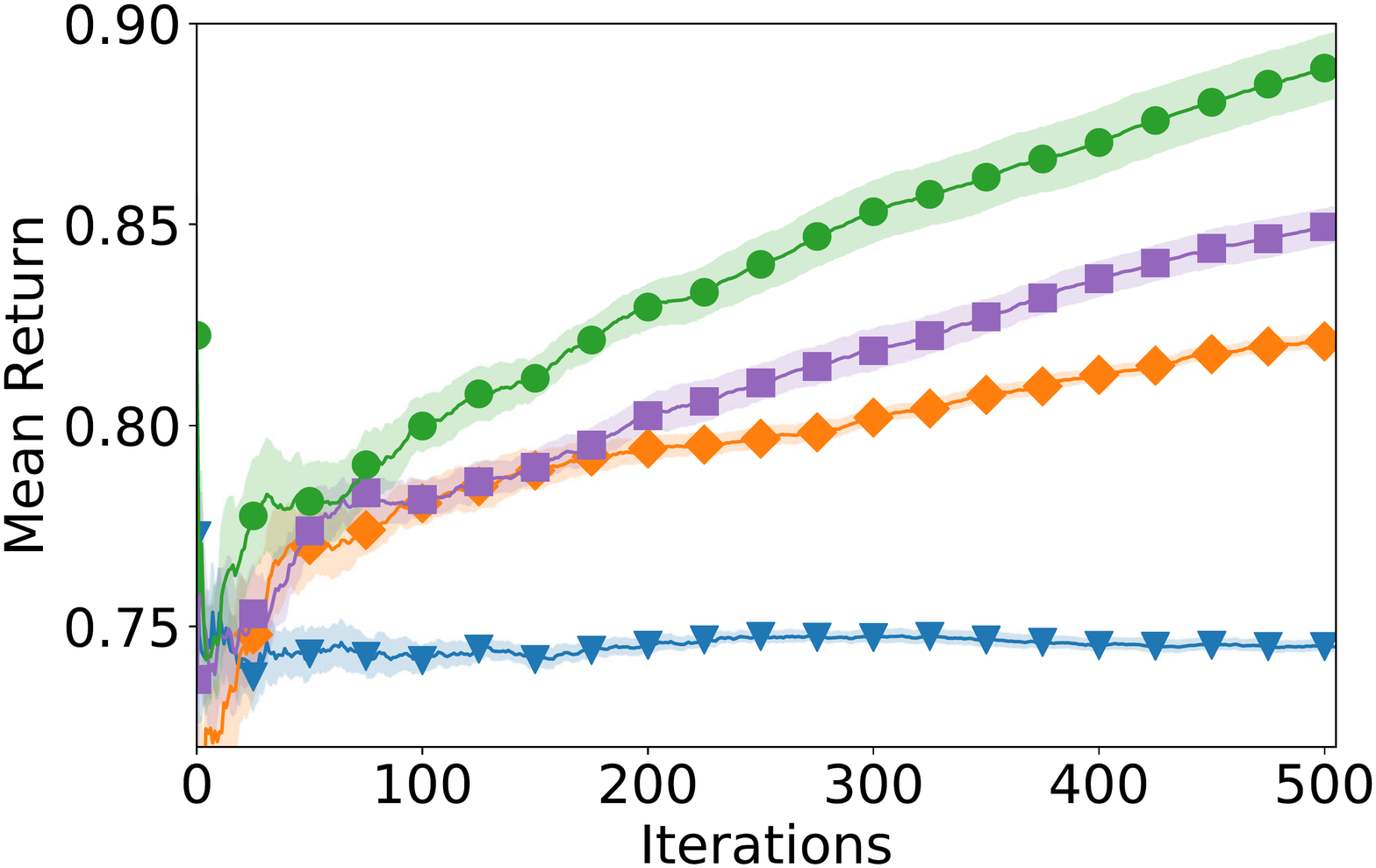} & \hspace{-3mm}\includegraphics[width=0.5\linewidth]{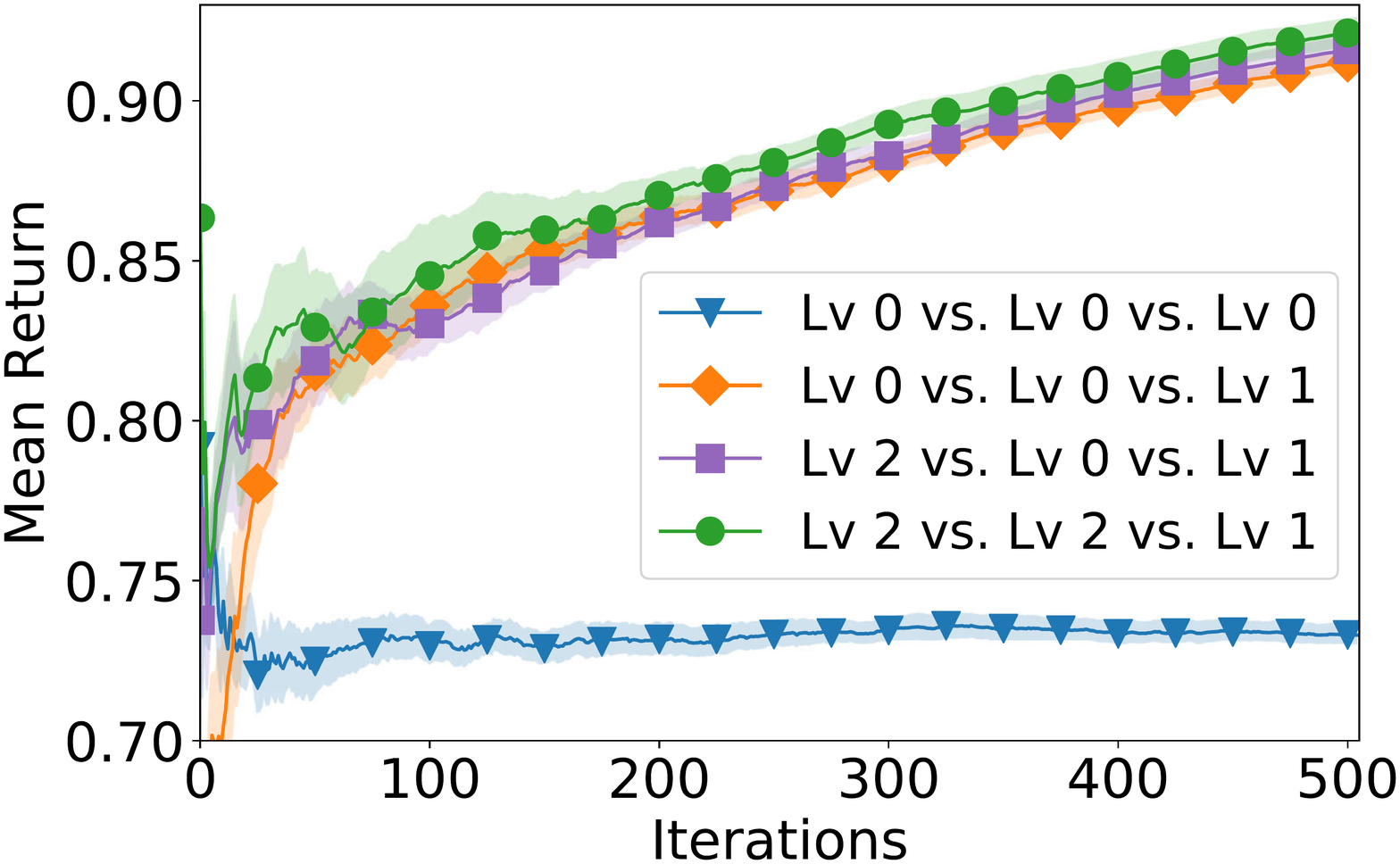}\\
       \hspace{-3mm} (a) {predators} & \hspace{-3mm}(b) {prey}
\end{tabular}
    \caption{Mean return of predators and prey in predator-prey game where the legend in (b) represents the levels of reasoning of predator $1$ vs. predator $2$ vs. prey.}
    \label{fig:simple_tag}\vspace{-3mm}
\end{figure}

Fig.~\ref{fig:simple_tag} shows results of the (scaled) mean return of the agents averaged over $10$ initializations of $5$ randomly selected actions with observed payoffs.
It can be observed from Fig.~\ref{fig:simple_tag}b that 
when the prey reasons at level $1$ and both predators reason at level $0$ (orange curve), its mean return is much higher than when reasoning at level $0$ (blue curve);
this results from the prey's ability to learn to stay within the boundary.
Specifically, there exist some ``dominated actions'' in this game, namely, those causing the prey to step beyond the boundary.
Regardless of the predators' policies, such dominated actions never give large returns to the prey and are thus likely to yield small values of GP-UCB for any actions (policies) selected by the predators.
So, by reasoning at level $1$ (i.e., by maximizing the expected value of GP-UCB), the prey is able to eliminate those dominated actions and thus learn to stay within the boundary.
From Fig.~\ref{fig:simple_tag}a, the mean return of the predators is also improved (orange curve) %when the prey reasons at level $1$, 
because the prey's ability to stay within the boundary allows the predators to improve their rewards by being close to the prey despite using random search for level-$0$  reasoning.
In contrast, when the prey reasons at level $0$, the predators rarely get rewarded (blue curve) since the prey repeatedly steps beyond the boundary.
On the other hand, when predator $1$ reasons at level $2$ (purple curve), the mean return of the predators is further increased since predator $1$ is now able to 
learn to actively move close to the prey instead of moving around using random search for level-$0$ reasoning (orange curve).
When both predators reason at level $2$ (green curve), their mean return is improved even further.
In both of these scenarios, the mean return of the prey stays close to that associated with the orange curve: Although the predators are able to actively approach the prey, 
this also further helps to prevent the prey from moving beyond the boundary, which compensates for the loss in its mean return due to the more strategic predators.

\section{Related Work}
The recent work of~\citet{sessa2019no} combines online learning and GP-UCB to derive a no-regret learning algorithm called \emph{GP-multiplicative weight} (GP-MW) for repeated games.
As explained in Section~\ref{subsec:level-0}, GP-MW can be used as a level-$0$ mixed strategy (i.e., no recursive reasoning) in our R2-B2 algorithm.
Moreover, BO has also been recently applied in game theory to find the Nash equilibria~\cite{picheny2019bayesian}.

Humans possess the ability to reason about the mental states of others~\cite{goldman2012theory}.
In particular, a person tends to reason recursively by analyzing the others' thinking about himself, which gives rise to recursive reasoning~\cite{pynadath2005psychsim}.
The recursive reasoning model of humans has inspired the development of the cognitive hierarchy model in behavioral game theory, 
which uses recursive reasoning to explain the behavior of players in games~\cite{camerer2004cognitive}.
Moreover, the improved decision-making capability offered by recursive reasoning has motivated its application in ML and sequential decision-making problems such as 
interactive partially observable Markov decisionn processes~\cite{gmytrasiewicz2005framework,hoang2013interactive}, MARL~\cite{wen2019probabilistic}, among others.

\emph{Deep neural networks} (DNNs) have recently been found to be vulnerable to carefully crafted adversarial examples~\cite{szegedy2013intriguing}.
Since then, a variety of adversarial attack methods have been developed to exploit this vulnerability of DNNs~\cite{goodfellow2014explaining}.
However, most of the existing attack methods are \emph{white-box} attacks since they require access to the gradient of the ML model.
In contrast, the more realistic \emph{black-box attacks}~\cite{tu2019autozoom,moon2019parsimonious}, which we have adopted in our experiments,
only require query access to the target ML model and have been attracting significant attention recently.
Of note, BO has recently been used for black-box adversarial attacks (without considering defenses) and demonstrated promising query efficiency~\cite{ru2020bayesopt}.
On the other hand, many attempts have been made to design adversarial defense methods~\cite{madry2017towards,tramer2017ensemble} to make ML models robust against adversarial attacks. 
In our experiments, we have adopted the input reconstruction/transformation technique~\cite{meng2017magnet,samangouei2018defense} as the defense mechanism, 
in which the defender attempts to transform the perturbed input to ensure the correct prediction by the ML model.
Refer to the detailed survey of adversarial ML in~\cite{yuan2019adversarial}.

\section{Conclusion and Future Work}
This paper describes the first BO algorithm called R2-B2 that is endowed with the capability of recursive reasoning to model the reasoning process in the interactions between boundedly rational\cref{brat}, self-interested agents with unknown, complex, and expensive-to-evaluate payoff functions in repeated games.
We prove that by reasoning at level $k\geq 2$ and one level higher than the other agents, our R2-B2 agent can achieve faster asymptotic convergence to no regret than that without utilizing recursive reasoning.
We empirically demonstrate the competitive performance and generality of R2-B2 through extensive experiments using synthetic games, adversarial ML, and MARL.
For our future work, we plan to investigate the connection of R2-B2 to other game-theoretic solution concepts such as Nash equilibrium.
We will also explore the extension of R2-B2 to a more general setting where
a level-$k$ agent selects its best response to the action of the other agent who reasons according to a distribution (e.g., Poisson) over lower levels instead of only at level $k-1$, which is also captured by the cognitive hierarchy model~\cite{camerer2004cognitive}.
We will consider generalizing R2-B2 to nonmyopic BO~\citep{dmitrii20a,ling16},
%batch active learning~\citep{LowAAMAS13,LowECML14b,NghiaICML14,LowAAMAS08,LowICAPS09,LowAAMAS11,LowAAMAS12,LowDyDESS14,LowAAMAS14,YehongAAAI16}, 
batch BO~\citep{daxberger17}, high-dimensional BO~\citep{NghiaAAAI18}, differentially private BO~\citep{dmitrii20b}, and multi-fidelity BO~\citep{yehong17,ZhangUAI19} settings and incorporating early stopping~\citep{dai2019}.
For applications with a huge budget of function evaluations, we like to couple R2-B2 with the use of distributed/decentralized~\citep{LowUAI12,Chen13,LowRSS13,LowTASE15,HoangICML16,NghiaAAAI19,HoangICML19,low15,Ruofei18} or online/stochastic~\citep{NghiaICML16,MinhAAAI17,LowECML14a,LowAAAI14,teng20,Haibin19,HaibinAPP} sparse GP models to represent the belief of the unknown objective function efficiently.
\section*{Acknowledgements}
This research/project is supported in part by the Singapore National Research Foundation through the Singapore-MIT Alliance for Research and Technology (SMART) Centre for Future Urban Mobility (FM) and in part by A$^*$STAR under its RIE$2020$ Advanced Manufacturing and Engineering (AME) Industry Alignment Fund -- Pre Positioning (IAF-PP) (Award A$19$E$4$a$0101$).
Teck-Hua Ho acknowledges funding from the Singapore National Research Foundation's Returning Singaporean Scientists Scheme, grant NRFRSS$2014$-$001$.

\bibliography{R2_B2}

\begin{thebibliography}{52}
\providecommand{\natexlab}[1]{#1}
\providecommand{\url}[1]{\texttt{#1}}
\expandafter\ifx\csname urlstyle\endcsname\relax
  \providecommand{\doi}[1]{doi: #1}\else
  \providecommand{\doi}{doi: \begingroup \urlstyle{rm}\Url}\fi

\bibitem[Camerer et~al.(2004)Camerer, Ho, and Chong]{camerer2004cognitive}
Camerer, C.~F., Ho, T.-H., and Chong, J.-K.
\newblock A cognitive hierarchy model of games.
\newblock \emph{Quarterly J. Economics}, 119\penalty0 (3):\penalty0 861--898,
  2004.

\bibitem[Chen et~al.(2012)Chen, Low, Tan, Oran, Jaillet, Dolan, and
  Sukhatme]{LowUAI12}
Chen, J., Low, K.~H., Tan, C. K.-Y., Oran, A., Jaillet, P., Dolan, J.~M., and
  Sukhatme, G.~S.
\newblock Decentralized data fusion and active sensing with mobile sensors for
  modeling and predicting spatiotemporal traffic phenomena.
\newblock In \emph{Proc. UAI}, pp.\  163--173, 2012.

\bibitem[Chen et~al.(2013{\natexlab{a}})Chen, Cao, Low, Ouyang, Tan, and
  Jaillet]{Chen13}
Chen, J., Cao, N., Low, K.~H., Ouyang, R., Tan, C. K.-Y., and Jaillet, P.
\newblock Parallel {Gaussian} process regression with low-rank covariance
  matrix approximations.
\newblock In \emph{Proc. UAI}, pp.\  152--161, 2013{\natexlab{a}}.

\bibitem[Chen et~al.(2013{\natexlab{b}})Chen, Low, and Tan]{LowRSS13}
Chen, J., Low, K.~H., and Tan, C. K.-Y.
\newblock {Gaussian} process-based decentralized data fusion and active sensing
  for mobility-on-demand system.
\newblock In \emph{Proc. RSS}, 2013{\natexlab{b}}.

\bibitem[Chen et~al.(2015)Chen, Low, Jaillet, and Yao]{LowTASE15}
Chen, J., Low, K.~H., Jaillet, P., and Yao, Y.
\newblock Gaussian process decentralized data fusion and active sensing for
  spatiotemporal traffic modeling and prediction in mobility-on-demand systems.
\newblock \emph{{IEEE} Trans. Autom. Sci. Eng.}, 12:\penalty0 901--921, 2015.

\bibitem[Dai et~al.(2019)Dai, Yu, Low, and Jaillet]{dai2019}
Dai, Z., Yu, H., Low, K.~H., and Jaillet, P.
\newblock Bayesian optimization meets {B}ayesian optimal stopping.
\newblock In \emph{Proc. {ICML}}, pp.\  1496--1506, 2019.

\bibitem[Daxberger \& Low(2017)Daxberger and Low]{daxberger17}
Daxberger, E.~A. and Low, K.~H.
\newblock Distributed batch {Gaussian} process optimization.
\newblock In \emph{Proc. {ICML}}, pp.\  951--960, 2017.

\bibitem[Gigerenzer \& Selten(2002)Gigerenzer and Selten]{Reinhard2002}
Gigerenzer, G. and Selten, R.
\newblock \emph{Bounded Rationality}.
\newblock MIT Press, 2002.

\bibitem[Gmytrasiewicz \& Doshi(2005)Gmytrasiewicz and
  Doshi]{gmytrasiewicz2005framework}
Gmytrasiewicz, P.~J. and Doshi, P.
\newblock A framework for sequential planning in multi-agent settings.
\newblock \emph{J. Artif. Intell. Res.}, 24:\penalty0 49--79, 2005.

\bibitem[Goldman(2012)]{goldman2012theory}
Goldman, A.~I.
\newblock Theory of mind.
\newblock In Margolis, E., Samuels, R., and Stich, S.~P. (eds.), \emph{The
  Oxford Handbook of Philosophy of Cognitive Science}. Oxford Univ. Press,
  2012.

\bibitem[Goodfellow et~al.(2015)Goodfellow, Shlens, and
  Szegedy]{goodfellow2014explaining}
Goodfellow, I.~J., Shlens, J., and Szegedy, C.
\newblock Explaining and harnessing adversarial examples.
\newblock In \emph{Proc. {ICLR}}, 2015.

\bibitem[Hoang et~al.(2017)Hoang, Hoang, and Low]{MinhAAAI17}
Hoang, Q.~M., Hoang, T.~N., and Low, K.~H.
\newblock A generalized stochastic variational {Bayesian} hyperparameter
  learning framework for sparse spectrum {Gaussian} process regression.
\newblock In \emph{Proc. {AAAI}}, pp.\  2007--2014, 2017.

\bibitem[Hoang et~al.(2019{\natexlab{a}})Hoang, Hoang, Low, and
  Kingsford]{HoangICML19}
Hoang, Q.~M., Hoang, T.~N., Low, K.~H., and Kingsford, C.
\newblock Collective model fusion for multiple black-box experts.
\newblock In \emph{Proc. ICML}, pp.\  2742--2750, 2019{\natexlab{a}}.

\bibitem[Hoang \& Low(2013)Hoang and Low]{hoang2013interactive}
Hoang, T.~N. and Low, K.~H.
\newblock Interactive {POMDP Lite}: Towards practical planning to predict and
  exploit intentions for interacting with self-interested agents.
\newblock In \emph{Proc. {IJCAI}}, 2013.

\bibitem[Hoang et~al.(2015)Hoang, Hoang, and Low]{NghiaICML16}
Hoang, T.~N., Hoang, Q.~M., and Low, K.~H.
\newblock A unifying framework of anytime sparse {Gaussian} process regression
  models with stochastic variational inference for big data.
\newblock In \emph{Proc. {ICML}}, pp.\  569--578, 2015.

\bibitem[Hoang et~al.(2016)Hoang, Hoang, and Low]{HoangICML16}
Hoang, T.~N., Hoang, Q.~M., and Low, K.~H.
\newblock A distributed variational inference framework for unifying parallel
  sparse {Gaussian} process regression models.
\newblock In \emph{Proc. ICML}, pp.\  382--391, 2016.

\bibitem[Hoang et~al.(2018)Hoang, Hoang, and Low]{NghiaAAAI18}
Hoang, T.~N., Hoang, Q.~M., and Low, K.~H.
\newblock Decentralized high-dimensional {Bayesian} optimization with factor
  graphs.
\newblock In \emph{Proc. {AAAI}}, pp.\  3231--3238, 2018.

\bibitem[Hoang et~al.(2019{\natexlab{b}})Hoang, Hoang, Low, and
  How]{NghiaAAAI19}
Hoang, T.~N., Hoang, Q.~M., Low, K.~H., and How, J.~P.
\newblock Collective online learning of {Gaussian} processes in massive
  multi-agent systems.
\newblock In \emph{Proc. {AAAI}}, pp.\  7850--7857, 2019{\natexlab{b}}.

\bibitem[Kharkovskii et~al.(2020{\natexlab{a}})Kharkovskii, Dai, and
  Low]{dmitrii20b}
Kharkovskii, D., Dai, Z., and Low, K.~H.
\newblock Private outsourced {Bayesian} optimization.
\newblock In \emph{Proc. ICML}, 2020{\natexlab{a}}.

\bibitem[Kharkovskii et~al.(2020{\natexlab{b}})Kharkovskii, Ling, and
  Low]{dmitrii20a}
Kharkovskii, D., Ling, C.~K., and Low, K.~H.
\newblock Nonmyopic {Gaussian} process optimization with macro-actions.
\newblock In \emph{Proc. AISTATS}, pp.\  4593--4604, 2020{\natexlab{b}}.

\bibitem[Kim \& Choi(2019)Kim and Choi]{kim2019local}
Kim, J. and Choi, S.
\newblock On local optimizers of acquisition functions in {Bayesian}
  optimization.
\newblock {arXiv}:1901.08350, 2019.

\bibitem[Kingma \& Welling(2014)Kingma and Welling]{kingma2013auto}
Kingma, D.~P. and Welling, M.
\newblock Auto-encoding variational {Bayes}.
\newblock In \emph{Proc. {ICLR}}, 2014.

\bibitem[Lattimore \& Szepesv{\'a}ri(2020)Lattimore and
  Szepesv{\'a}ri]{lattimore2018bandit}
Lattimore, T. and Szepesv{\'a}ri, C.
\newblock \emph{{Bandit Algorithms}}.
\newblock 2020.

\bibitem[Ling et~al.(2016)Ling, Low, and Jaillet]{ling16}
Ling, C.~K., Low, K.~H., and Jaillet, P.
\newblock {Gaussian} process planning with {Lipschitz} continuous reward
  functions: Towards unifying {Bayesian} optimization, active learning, and
  beyond.
\newblock In \emph{Proc. {AAAI}}, pp.\  1860--1866, 2016.

\bibitem[Low et~al.(2014)Low, Xu, Chen, Lim, and {\"{O}zg\"{u}l}]{LowECML14a}
Low, K.~H., Xu, N., Chen, J., Lim, K.~K., and {\"{O}zg\"{u}l}, E.~B.
\newblock Generalized online sparse {Gaussian} processes with application to
  persistent mobile robot localization.
\newblock In \emph{Proc. {ECML/PKDD Nectar Track}}, pp.\  499--503, 2014.

\bibitem[Low et~al.(2015)Low, Yu, Chen, and Jaillet]{low15}
Low, K.~H., Yu, J., Chen, J., and Jaillet, P.
\newblock Parallel {Gaussian} process regression for big data: Low-rank
  representation meets {M}arkov approximation.
\newblock In \emph{Proc. {AAAI}}, pp.\  2821--2827, 2015.

\bibitem[Lowe et~al.(2017)Lowe, Wu, Tamar, Harb, Abbeel, and
  Mordatch]{lowe2017multi}
Lowe, R., Wu, Y., Tamar, A., Harb, J., Abbeel, P., and Mordatch, I.
\newblock Multi-agent actor-critic for mixed cooperative-competitive
  environments.
\newblock In \emph{Proc. {NeurIPS}}, pp.\  6379--6390, 2017.

\bibitem[Madry et~al.(2017)Madry, Makelov, Schmidt, Tsipras, and
  Vladu]{madry2017towards}
Madry, A., Makelov, A., Schmidt, L., Tsipras, D., and Vladu, A.
\newblock Towards deep learning models resistant to adversarial attacks.
\newblock In \emph{Proc. {ICLR}}, 2017.

\bibitem[Meng \& Chen(2017)Meng and Chen]{meng2017magnet}
Meng, D. and Chen, H.
\newblock {MagNet}: A two-pronged defense against adversarial examples.
\newblock In \emph{Proc. {CCS}}, pp.\  135--147, 2017.

\bibitem[Moon et~al.(2019)Moon, An, and Song]{moon2019parsimonious}
Moon, S., An, G., and Song, H.~O.
\newblock Parsimonious black-box adversarial attacks via efficient
  combinatorial optimization.
\newblock In \emph{Proc. {ICML}}, 2019.

\bibitem[Nisan et~al.(2007)Nisan, Roughgarden, Tardos, and
  Vazirani]{nisan2007algorithmic}
Nisan, N., Roughgarden, T., Tardos, E., and Vazirani, V.~V.
\newblock \emph{Algorithmic {Game Theory}}.
\newblock Cambridge Univ. Press, 2007.

\bibitem[Ouyang \& Low(2018)Ouyang and Low]{Ruofei18}
Ouyang, R. and Low, K.~H.
\newblock Gaussian process decentralized data fusion meets transfer learning in
  large-scale distributed cooperative perception.
\newblock In \emph{Proc. AAAI}, pp.\  3876--3883, 2018.

\bibitem[Picheny et~al.(2019)Picheny, Binois, and Habbal]{picheny2019bayesian}
Picheny, V., Binois, M., and Habbal, A.
\newblock A {Bayesian} optimization approach to find {Nash} equilibria.
\newblock \emph{Journal of Global Optimization}, 73\penalty0 (1):\penalty0
  171--192, 2019.

\bibitem[Pynadath \& Marsella(2005)Pynadath and Marsella]{pynadath2005psychsim}
Pynadath, D.~V. and Marsella, S.~C.
\newblock {PsychSim}: Modeling theory of mind with decision-theoretic agents.
\newblock In \emph{Proc. {IJCAI}}, pp.\  1181--1186, 2005.

\bibitem[Rahimi \& Recht(2007)Rahimi and Recht]{rahimi2008random}
Rahimi, A. and Recht, B.
\newblock Random features for large-scale kernel machines.
\newblock In \emph{Proc. {NeurIPS}}, pp.\  1177--1184, 2007.

\bibitem[Rasmussen \& Williams(2006)Rasmussen and
  Williams]{rasmussen2004gaussian}
Rasmussen, C.~E. and Williams, C. K.~I.
\newblock \emph{{Gaussian Processes} for {Machine Learning}}.
\newblock MIT Press, 2006.

\bibitem[Ru et~al.(2020)Ru, Cobb, Blaas, and Gal]{ru2020bayesopt}
Ru, B., Cobb, A., Blaas, A., and Gal, Y.
\newblock {BayesOpt} adversarial attack.
\newblock In \emph{Proc. {ICLR}}, 2020.

\bibitem[Samangouei et~al.(2018)Samangouei, Kabkab, and
  Chellappa]{samangouei2018defense}
Samangouei, P., Kabkab, M., and Chellappa, R.
\newblock Defense-{GAN}: Protecting classifiers against adversarial attacks
  using generative models.
\newblock In \emph{Proc. {ICLR}}, 2018.

\bibitem[Sessa et~al.(2019)Sessa, Bogunovic, Kamgarpour, and
  Krause]{sessa2019no}
Sessa, P.~G., Bogunovic, I., Kamgarpour, M., and Krause, A.
\newblock No-regret learning in unknown games with correlated payoffs.
\newblock In \emph{Proc. {NeurIPS}}, 2019.

\bibitem[Shahriari et~al.(2016)Shahriari, Swersky, Wang, Adams, and {de
  Freitas}]{shahriari2016taking}
Shahriari, B., Swersky, K., Wang, Z., Adams, R.~P., and {de Freitas}, N.
\newblock Taking the human out of the loop: A review of {Bayesian}
  optimization.
\newblock \emph{Proc. of the IEEE}, 104\penalty0 (1):\penalty0 148--175, 2016.

\bibitem[Srinivas et~al.(2010)Srinivas, Krause, Kakade, and
  Seeger]{srinivas2009gaussian}
Srinivas, N., Krause, A., Kakade, S.~M., and Seeger, M.
\newblock {Gaussian} process optimization in the bandit setting: No regret and
  experimental design.
\newblock In \emph{Proc. {ICML}}, pp.\  1015--1022, 2010.

\bibitem[Szegedy et~al.(2014)Szegedy, Zaremba, Sutskever, Bruna, Erhan,
  Goodfellow, and Fergus]{szegedy2013intriguing}
Szegedy, C., Zaremba, W., Sutskever, I., Bruna, J., Erhan, D., Goodfellow, I.,
  and Fergus, R.
\newblock Intriguing properties of neural networks.
\newblock In \emph{Proc. {ICLR}}, 2014.

\bibitem[Teng et~al.(2020)Teng, Chen, Zhang, and Low]{teng20}
Teng, T., Chen, J., Zhang, Y., and Low, K.~H.
\newblock Scalable variational bayesian kernel selection for sparse {Gaussian}
  process regression.
\newblock In \emph{Proc. {AAAI}}, pp.\  5997--6004, 2020.

\bibitem[Tram{\`e}r et~al.(2018)Tram{\`e}r, Kurakin, Papernot, Goodfellow,
  Boneh, and McDaniel]{tramer2017ensemble}
Tram{\`e}r, F., Kurakin, A., Papernot, N., Goodfellow, I., Boneh, D., and
  McDaniel, P.
\newblock Ensemble adversarial training: Attacks and defenses.
\newblock In \emph{Proc. {ICLR}}, 2018.

\bibitem[Tu et~al.(2019)Tu, Ting, Chen, Liu, Zhang, Yi, Hsieh, and
  Cheng]{tu2019autozoom}
Tu, C.-C., Ting, P., Chen, P.-Y., Liu, S., Zhang, H., Yi, J., Hsieh, C.-J., and
  Cheng, S.-M.
\newblock {AutoZOOM}: Autoencoder-based zeroth order optimization method for
  attacking black-box neural networks.
\newblock In \emph{Proc. {AAAI}}, pp.\  742--749, 2019.

\bibitem[Wen et~al.(2019)Wen, Yang, Luo, Wang, and Pan]{wen2019probabilistic}
Wen, Y., Yang, Y., Luo, R., Wang, J., and Pan, W.
\newblock Probabilistic recursive reasoning for multi-agent reinforcement
  learning.
\newblock In \emph{Proc. {ICLR}}, 2019.

\bibitem[Xu et~al.(2014)Xu, Low, Chen, Lim, and {\"{O}zg\"{u}l}]{LowAAAI14}
Xu, N., Low, K.~H., Chen, J., Lim, K.~K., and {\"{O}zg\"{u}l}, E.~B.
\newblock {GP-Localize}: Persistent mobile robot localization using online
  sparse {Gaussian} process observation model.
\newblock In \emph{Proc. {AAAI}}, pp.\  2585--2592, 2014.

\bibitem[Yu et~al.(2019{\natexlab{a}})Yu, Chen, Dai, Low, and
  Jaillet]{Haibin19}
Yu, H., Chen, Y., Dai, Z., Low, K.~H., and Jaillet, P.
\newblock Implicit posterior variational inference for deep {Gaussian}
  processes.
\newblock In \emph{Proc. {NeurIPS}}, pp.\  14475--14486, 2019{\natexlab{a}}.

\bibitem[Yu et~al.(2019{\natexlab{b}})Yu, Hoang, Low, and Jaillet]{HaibinAPP}
Yu, H., Hoang, T.~N., Low, K.~H., and Jaillet, P.
\newblock Stochastic variational inference for {Bayesian} sparse {Gaussian}
  process regression.
\newblock In \emph{Proc. {IJCNN}}, 2019{\natexlab{b}}.

\bibitem[Yuan et~al.(2019)Yuan, He, Zhu, and Li]{yuan2019adversarial}
Yuan, X., He, P., Zhu, Q., and Li, X.
\newblock Adversarial examples: Attacks and defenses for deep learning.
\newblock \emph{IEEE Trans. Neural Netw. Learning Syst.}, 30\penalty0
  (9):\penalty0 2805--2824, 2019.

\bibitem[Zhang et~al.(2017)Zhang, Hoang, Low, and Kankanhalli]{yehong17}
Zhang, Y., Hoang, T.~N., Low, K.~H., and Kankanhalli, M.
\newblock Information-based multi-fidelity {Bayesian} optimization.
\newblock In \emph{Proc. {NIPS} Workshop on {Bayesian} Optimization}, 2017.

\bibitem[Zhang et~al.(2019)Zhang, Dai, and Low]{ZhangUAI19}
Zhang, Y., Dai, Z., and Low, K.~H.
\newblock Bayesian optimization with binary auxiliary information.
\newblock In \emph{Proc. UAI}, 2019.

\end{thebibliography}
\bibliographystyle{icml2020}

%\newpage
\onecolumn
\appendix

\section{More Background}
\subsection{Background on Gaussian Processes}
\label{app:gp}
In the repeated game, the attacker ($\mathcal{A}$) models its belief about its payoff function $f_1$ using a \emph{Gaussian process} (GP) 
$\{f_1(\mathbf{x}_1, \mathbf{x}_2)\}_{\mathbf{x}_1\in {\mathcal{X}_1}, \mathbf{x}_2\in {\mathcal{X}_2}}$.
In particular, any finite subset of $\{f_1(\mathbf{x}_1, \mathbf{x}_2)\}_{\mathbf{x}_1\in {\mathcal{X}_1}, \mathbf{x}_2\in {\mathcal{X}_2}}$ 
follows a multivariate Gaussian distribution \cite{rasmussen2004gaussian}.
A GP is fully specified by the prior mean $\mu(\mathbf{x}_1, \mathbf{x}_2)$ and kernel function $k([\mathbf{x}_1, \mathbf{x}_2], [\mathbf{x}'_1, \mathbf{x}'_2])$, 
and we assume w.l.o.g. that $\mu(\mathbf{x}_1, \mathbf{x}_2)=0$ and $k([\mathbf{x}_1, \mathbf{x}_2], [\mathbf{x}'_1, \mathbf{x}'_2]) \leq 1$ 
for all $\mathbf{x}_1, \mathbf{x}'_1 \in \mathcal{X}_1$ and $\mathbf{x}_2, \mathbf{x}'_2 \in \mathcal{X}_2$.
Given a set of $T$ noisy observations $\mathbf{y}_{T}\triangleq [y_t]^{\top}_{t=1,\ldots,T}$ at inputs $[\mathbf{x}_{1,1}, \mathbf{x}_{2,1}],\ldots,[\mathbf{x}_{1,T}, \mathbf{x}_{2,T}]$, 
the posterior GP belief of $f_1$ at any input $[\mathbf{x}_1, \mathbf{x}_2]$ is a Gaussian distribution with the following posterior mean and variance:
\begin{equation}
\begin{array}{rcl}
    \mu_T(\mathbf{x}_1, \mathbf{x}_2)&\hspace{-2.4mm}\triangleq&\hspace{-2.4mm}\displaystyle\mathbf{k}_T(\mathbf{x}_1, \mathbf{x}_2)^\top(\mathbf{K}_T+\sigma^2I)^{-1}\mathbf{y}_{T}\ , \vspace{0.5mm}\\
    \sigma_T^2(\mathbf{x}_1, \mathbf{x}_2)&\hspace{-2.4mm}\triangleq&\hspace{-2.4mm}\displaystyle k([\mathbf{x}_1, \mathbf{x}_2],[\mathbf{x}_1, \mathbf{x}_2])-\mathbf{k}_T(\mathbf{x}_1, \mathbf{x}_2)^\top(\mathbf{K}_T+\sigma^2I)^{-1}\mathbf{k}_T(\mathbf{x}_1, \mathbf{x}_2)
\end{array}
\label{gp_posterior}
\end{equation}
where $\mathbf{K}_T\triangleq \left[k([\mathbf{x}_{1,t}, \mathbf{x}_{2,t}], [\mathbf{x}_{1,t'}, \mathbf{x}_{2,t'}])\right]_{t,t'=1,\ldots,T}$ and 
$\mathbf{k}_T(\mathbf{x}_1, \mathbf{x}_2)\triangleq \left[k([\mathbf{x}_{1,t}, \mathbf{x}_{2,t}],[\mathbf{x}_1, \mathbf{x}_2])\right]^\top_{t=1,\ldots,T}$.

\subsection{The GP-MW Algorithm}
\label{app:gp_mw}
When $\mathcal{A}$ (the attacker) adopts the GP-MW algorithm as the level-$0$ strategy,
after iteration $t$ of the repeated game, $\mathcal{A}$ calculates the updated value of the GP-UCB acquisition function at every input in its entire domain $\mathcal{X}_1$ 
(while fixing the defender's input $\mathbf{x}_2$ at the value selected in iteration $t$: $\mathbf{x}_{2,t}$), 
plugs in the (negative) GP-UCB values as the loss vector (with the length of the vector being equal to the size of its domain: $|\mathcal{X}_1|$) 
in the widely used multiplicative-weight online learning algorithm to update the randomized/mixed strategy $\mathcal{P}^{0}_{1, t+1}$. 
Subsequently, the resulting updated distribution will be used to sample $\mathcal{A}$'s action in the next iteration $t+1$, i.e., $\mathbf{x}_{1,t+1}\sim \mathcal{P}^{0}_{1, t+1}$. 
Note that the proof of Theorem~\ref{regret_gp} results from a slight modification to the proof of GP-MW~\cite{sessa2019no}, i.e., 
the work of~\citet{sessa2019no} has assumed that the payoff function has bounded norm in a reproducing kernel Hilbert space,
whereas we assume that the payoff function is sampled from a GP.
Both assumptions are commonly used in the analysis of BO algorithms.
Refer to the work of~\citet{sessa2019no} for more details about the GP-MW algorithm.

\section{Extension to Games Involving More than Two Agents}
\label{subsec:more_than_two_players}
The R2-B2, as well as R2-B2-Lite, algorithm can be extended to repeated games involving more than two ($M > 2$) agents. 
A motivating scenario for this type of games with $M>2$ agents is MARL, in which every individual agent attempts to maximize its own return (payoff).
Here, we use $\mathcal{A}_{1},\ldots,\mathcal{A}_{M}$ to represent the $M$ agents.

{\bf Level-$k=0$ Strategy.}
The extension of level-$0$ reasoning is trivial since level-$0$ strategies are agnostic with respect to the other agent's action selection strategies, 
and can thus treat all other agents as a single collective agent.
As a result, if GP-MW is adopted as the level-$0$ strategy, the theoretical guarantee of Theorem~\ref{regret_gp} still holds.

{\bf Level-$k=1$ Strategy.}
If the agent $\mathcal{A}_1$ thinks that all other agents ($\mathcal{A}_2,\ldots,\mathcal{A}_M$) reason at level $0$ and knows the level-$0$ strategies of all other agents,
$\mathcal{A}_1$ can reason at level $1$ by:
\begin{equation}
\begin{split}
    \mathbf{x}^{1}_{1,t}=&\mathop{\arg\max}_{\mathbf{x}_1\in \mathcal{X}_1} \mathbb{E}_{\mathbf{x}^{0}_{2,t},\ldots,\mathbf{x}^{0}_{M,t}}
    \left[\alpha_{1,t}(\mathbf{x}_1, \mathbf{x}^{0}_{2,t},\ldots,\mathbf{x}^{0}_{M,t})\right],
\end{split}
\label{eq:level_1_defender_more_than_2}
\end{equation}
in which the expectation is taken over the level-$0$ strategies of all other agents $\mathcal{A}_2,\ldots,\mathcal{A}_M$.
R2-B2-Lite can also be applied:
\begin{equation}
    \mathbf{x}^{1}_{1,t}=\mathop{\arg\max}_{\mathbf{x}_1 \in \mathcal{X}_1}\alpha_{1,t}(\mathbf{x}_1,\widetilde{\mathbf{x}}^{0}_{2,t},\ldots,\widetilde{\mathbf{x}}^{0}_{M,t}),
\label{eq:borr_light_level_1_more_than_2}
\end{equation}
in which $\widetilde{\mathbf{x}}^{0}_{2,t},\ldots,\widetilde{\mathbf{x}}^{0}_{M,t}$ are sampled from the corresponding level-$0$ strategies of agents $\mathcal{A}_2,\ldots,\mathcal{A}_M$.

For level-$1$ reasoning, the actions of all other agents can be viewed as the joint action of a single collective agent, 
whose level-$0$ strategy (action distribution) factorizes across different agents.
As a result, the theoretical guarantees of Theorems~\ref{theorem_level_1} and~\ref{theorem_borr_lite}
are still valid. 

{\bf Level-$k\geq 2$ Strategy.}
Level-$k\geq 2$ reasoning with $M > 2$ agents is significantly more complicated than the two-agent setting, 
mainly due to the fact that the other agents may not reason at the same level. 
For simplicity, we consider the scenario in which the agent $\mathcal{A}_1$ reasons at level $2$, and thus all other agents reason at either level $1$ or $0$. 
This is a common scenario since as discussed in Section~\ref{subsec:level_k_policy} and will be explained at the end of this section, the agents have a strong tendency to reason at lower levels in the setting with $M>2$ agents.
Without loss of generality, we assume that agents $2$ to $M_0$ reason at level $0$, and agents $M_0+1$ to $M$ reason at level 1 
(by following the strategy of~\eqref{eq:level_1_defender_more_than_2}). 
In this case, the level-$2$ action of agent $\mathcal{A}_1$ is selected by best-responding to the corresponding strategy of each of the other agents:
\begin{equation}
\begin{split}
    \mathbf{x}^{2}_{1,t}&=\mathop{\arg\max}_{\mathbf{x}_1\in \mathcal{X}_1} \mathbb{E}_{\mathbf{x}^{0}_{2,t},\ldots,\mathbf{x}^{0}_{M_0,t}}
    \left[\alpha_{1,t}(\mathbf{x}_1, \mathbf{x}^{0}_{2,t},\ldots,\mathbf{x}^{0}_{M_0,t}, \mathbf{x}^{1}_{M_0+1,t}, \ldots, \mathbf{x}^{1}_{M,t})\right].
\label{eq:level_2_defender_more_than_2}
\end{split}
\end{equation}
Specifically, the level-$1$ actions of those agents reasoning at level $1$ ($\mathbf{x}^{1}_{M_0+1,t}, \ldots, \mathbf{x}^{1}_{M,t}$) can be calculated using~\eqref{eq:level_1_defender_more_than_2}, 
and the expectation in~\eqref{eq:level_2_defender_more_than_2} is taken with respect to the level-$0$ strategies of those agents reasoning at level $0$ ($\mathbf{x}^{0}_{2,t},\ldots,\mathbf{x}^{0}_{M_0,t}$). 
Interestingly, the level-$2$ reasoning strategy of~\eqref{eq:level_2_defender_more_than_2} enjoys the same regret upper bound as shown in Theorem~\ref{theorem_level_1} or
Theorem~\ref{theorem_level_k}, depending on whether there exists level-$0$ agents (see the detailed explanation and the proof in Appendix~\ref{app:proof_multiplayer}). 
Unfortunately, the complexity of reasoning at levels $k\geq 3$ grows excessively. 
Firstly, every other agent reasoning at a lower level $k\geq 2$ may best-respond to the other agents in multiple ways. 
For example, if there are $M=3$ agents in the environment and agent $\mathcal{A}_1$ reasons at level $2$, 
$\mathcal{A}_1$ might choose its level-$2$ action in three different ways, with the corresponding reasoning levels of the $3$ agents being 
$[2, 1, 1]$, $[2, 1, 0]$ or $[2, 0, 1]$. As a result, if Agent $\mathcal{A}_2$ chooses to reason at level $3$, 
in addition to obtaining the information that agent $\mathcal{A}_1$ reasons at level $2$, 
$\mathcal{A}_2$ also needs to additionally know in which of the three ways will the level-$2$ reasoning of $\mathcal{A}_1$ be performed. 
Therefore, when $M>2$ agents are present, as the reasoning level increases, the reasoning complexity, as well as computational cost, grows significantly.
As a consequence, compared with the agents in $2$-agent games, the agents in games with $M>2$ agents are expected to display a stronger preference to 
reasoning at low levels.

\section{Proof of Theorems~\ref{theorem_level_1} and~\ref{theorem_level_k}}
\label{app:proof}
Before proving the main theorems, we need the following lemma showing a high-probability uniform upper bound on the value of the payoff function.
\begin{lemma}
\label{lemma:high_prob}
Let $\delta\in (0, 1)$ and $\beta_t=2\log (|\mathcal{X}_1|t^2\pi^2/3\delta)$, then with probability $\geq 1 - \delta$,
\[
|f_1(\mathbf{x}_1,\mathbf{x}_2) - \mu_{t-1}(\mathbf{x}_1,\mathbf{x}_2)| \leq \beta^{1/2}_t \sigma_{t-1}(\mathbf{x}_1,\mathbf{x}_2)
\]
for all $\mathbf{x}_1\in \mathcal{X}_1$, $\mathbf{x}_2\in \mathcal{X}_2$, and $t\geq 1$.
\end{lemma}
The proof of Lemma~\ref{lemma:high_prob} makes use of the Gaussian concentration inequality and the union bound, and the proof
can be found in Lemma $5.1$ of~\citet{srinivas2009gaussian}.
Note that a tighter confidence bound (i.e., a smaller value of $\beta_t=2\log (|\mathcal{X}_1|t^2\pi^2/6\delta)$) is possible, however, the value of 
$\beta_t$ in Lemma~\ref{lemma:high_prob} is selected for convenience to match the requirement of GP-MW (Theorem~\ref{regret_gp}).

\subsection{Theorem~\ref{theorem_level_1}}
\label{app:proof_theorem_2}
Denote the history of game plays for $\mathcal{D}$ (the defender) up to iteration $t-1$ as $\mathcal{H}_{t-1}$, which includes $\mathcal{D}$'s selected actions (inputs) and observed payoffs (outputs) in every iteration from $1$ to $t-1$:
$\mathcal{H}_{t-1}=[\mathbf{x}_{2,1}, y_{2,1}, \mathbf{x}_{2,2}, y_{2,2}, \ldots, \mathbf{x}_{2,t-1}, y_{2,t-1}]$.
Again, we use superscripts to denote the reasoning level such that if $\mathcal{D}$ reasons at level $0$, $\mathcal{H}_{t-1}=[\mathbf{x}^{0}_{2,1}, y^{0}_{2,1}, \mathbf{x}^{0}_{2,2}, y^{0}_{2,2}, \ldots, \mathbf{x}^{0}_{2,t-1}, y^{0}_{2,t-1}]$.

Here, we analyze the regret of the level-$1$ strategy, i.e., when $\mathcal{A}$ (the attacker) reasons at level $k=1$ and $\mathcal{D}$ (the defender) reasons at level $k'=0$.
Note that in iteration $t$, the level-$0$ strategy of $\mathcal{D}$ (i.e., the distribution of $\mathbf{x}_{2,t}$) may depend on the history of input-output pairs of $\mathcal{D}$, i.e., $\mathcal{H}_{t-1}$,
which is true for both the GP-MW and EXP3 strategies.
Therefore, when analyzing $\mathcal{A}$'s expected regret in iteration $t$ (with the expectation taken over the level-$0$ strategy of $\mathcal{D}$ in iteration $t$), we need to condition on $\mathcal{H}_{t-1}$.
We denote the regret of $\mathcal{A}$ in iteration $t$ as $r_{1,t}$, i.e., $R_{1,T}=\sum^T_{t=1}r_{1,t}$ in which $R_{1,T}$ represents external regret defined in~\eqref{extregret}.
As a result, with probability of at least $1-\delta$, the expected regret of $\mathcal{A}$ (the attacker) in iteration $t$, given $\mathcal{H}_{t-1}$, can be analyzed as
\begin{equation}
\begin{split}
    \mathbb{E}_{\mathbf{x}^{0}_{2,t}}[&r_{1,t}|\mathcal{H}_{t-1}]
    =\mathbb{E}_{\mathbf{x}^{0}_{2,t}}\left[f_1\left(\mathbf{x}_1^*,\mathbf{x}^{0}_{2,t}\right) - f_1\left(\mathbf{x}^{1}_{1,t}, \mathbf{x}^{0}_{2,t}]\right)|\mathcal{H}_{t-1}\right] \\
    &\stackrel{\text{(a)}}{\leq} \mathbb{E}_{\mathbf{x}^{0}_{2,t}}\left[\alpha_{1,t}\left(\mathbf{x}_1^*,\mathbf{x}^{0}_{2,t}\right) - f_1\left(\mathbf{x}^{1}_{1,t}, \mathbf{x}^{0}_{2,t}\right)|\mathcal{H}_{t-1}\right] \\
    &\stackrel{\text{(b)}}{\leq}\mathbb{E}_{\mathbf{x}^{0}_{2,t}}\left[\alpha_{1,t}\left(\mathbf{x}^{1}_{1,t},\mathbf{x}^{0}_{2,t}\right)
    - f_1\left(\mathbf{x}^{1}_{1,t}, \mathbf{x}^{0}_{2,t}\right) |\mathcal{H}_{t-1}\right]\\
    &\stackrel{\text{(c)}}{\leq}\mathbb{E}_{\mathbf{x}^{0}_{2,t}}\left[\mu_{t-1}(\mathbf{x}^{1}_{1,t}, \mathbf{x}^{0}_{2,t}) + \beta^{1/2}_t \sigma_{t-1}(\mathbf{x}^{1}_{1,t}, \mathbf{x}^{0}_{2,t})
    - f_1\left(\mathbf{x}^{1}_{1,t}, \mathbf{x}^{0}_{2,t}\right) |\mathcal{H}_{t-1}\right]\\
    &\stackrel{\text{(d)}}{\leq} \mathbb{E}_{\mathbf{x}^{0}_{2,t}}\left[2\beta^{1/2}_t\sigma_{t-1}(\mathbf{x}^{1}_{1,t}, \mathbf{x}^{0}_{2,t})|\mathcal{H}_{t-1}\right] 
\end{split}
\label{proof_determ}
\end{equation}
in which (a) results from Lemma~\ref{lemma:high_prob} and the definition of the GP-UCB acquisition function ($\alpha$) in Section~\ref{background}, 
(b) follows from the definition of the level-$1$ strategy~\eqref{eq:level_1_defender} as well as the linearity of the expectation operator,
(c) results from the definition of the GP-UCB acquisition function, and (d) is again a consequence of Lemma~\ref{lemma:high_prob}.

Next, the expected external regret of $\mathcal{A}$ reasoning at level $1$ can be upper-bounded:
\begin{equation}
\begin{split}
    \mathbb{E}[R_{1,T}] &= \mathbb{E}_{\mathbf{x}^{0}_{2,1}, y^{0}_{2,1},\ldots,\mathbf{x}^{0}_{2,T-1}, y^{0}_{2,T-1}, \mathbf{x}^{0}_{2,T}}[R_{1,T}]\\
    &=\mathbb{E}_{\mathbf{x}^{0}_{2,1}, y^{0}_{2,1},\ldots,\mathbf{x}^{0}_{2,T-1}, y^{0}_{2,T-1}, \mathbf{x}^{0}_{2,T}}\left[\sum^T_{t=1}r_{1,t}\right]\\
    &\stackrel{\text{(a)}}{=}\mathbb{E}_{\mathbf{x}^{0}_{2,1}}\left[r_{1,1}\right] + \mathbb{E}_{\mathbf{x}^{0}_{2,1}, y^{0}_{2,1}, \mathbf{x}^{0}_{2,2}}\left[r_{1,2}\right] + \ldots + \mathbb{E}_{\mathbf{x}^{0}_{2,1}, y^{0}_{2,1},\ldots,\mathbf{x}^{0}_{2,T-1}, y^{0}_{2,T-1}, \mathbf{x}^{0}_{2,T}}\left[r_{1,T}\right]\\
    &\stackrel{\text{(b)}}{=}\mathbb{E}_{\mathbf{x}^{0}_{2,1}}\left[r_{1,1}\right] + \mathbb{E}_{\mathbf{x}^{0}_{2,1}, y^{0}_{2,1}} \left[ \mathbb{E}_{\mathbf{x}^{0}_{2,2}}\left[r_{1,2}|\mathbf{x}^{0}_{2,1}, y^{0}_{2,1} \right]\right] + \ldots +\\ 
    &\qquad \mathbb{E}_{\mathbf{x}^{0}_{2,1}, y^{0}_{2,1},\ldots,\mathbf{x}^{0}_{2,T-1}, y^{0}_{2,T-1}} \left[ \mathbb{E}_{\mathbf{x}^{0}_{2,T}}\left[r_{1,T}|\mathbf{x}^{0}_{2,1}, y^{0}_{2,1},\ldots,\mathbf{x}^{0}_{2,T-1}, y^{0}_{2,T-1} \right]\right]\\
    &=\mathbb{E}_{\mathbf{x}^{0}_{2,1}}\left[r_{1,1}\right] + \mathbb{E}_{\mathcal{H}_1} \left[ \mathbb{E}_{\mathbf{x}^{0}_{2,2}}\left[r_{1,2}|\mathcal{H}_1 \right]\right] + \ldots + \mathbb{E}_{\mathcal{H}_{T-1}} \left[ \mathbb{E}_{\mathbf{x}^{0}_{2,T}}\left[r_{1,T}|\mathcal{H}_{T-1} \right]\right]\\
    &\stackrel{\text{(c)}}{\leq} \mathbb{E}_{\mathbf{x}^{0}_{2,1}}\left[2\beta^{1/2}_1\sigma_{0}(\mathbf{x}_{1,1}, \mathbf{x}_{2,1})\right] + \mathbb{E}_{\mathcal{H}_1} \left[ \mathbb{E}_{\mathbf{x}^{0}_{2,2}}\left[2\beta^{1/2}_2\sigma_{1}(\mathbf{x}_{1,2}, \mathbf{x}_{2,2})|\mathcal{H}_1 \right]\right] + \ldots + \\
    &\qquad \mathbb{E}_{\mathcal{H}_{T-1}} \left[ \mathbb{E}_{\mathbf{x}^{0}_{2,T}}\left[2\beta^{1/2}_T\sigma_{T-1}(\mathbf{x}_{1,T}, \mathbf{x}_{2,T})|\mathcal{H}_{T-1} \right]\right]\\
    &\stackrel{\text{(d)}}{=} \mathbb{E}_{\mathbf{x}^{0}_{2,1}}\left[2\beta^{1/2}_1\sigma_{0}(\mathbf{x}_{1,1}, \mathbf{x}_{2,1})\right] + \mathbb{E}_{\mathcal{H}_1, \mathbf{x}^{0}_{2,2}} \left[ 2\beta^{1/2}_2\sigma_{1}(\mathbf{x}_{1,2}, \mathbf{x}_{2,2})\right] + \ldots + \\
    &\qquad \mathbb{E}_{\mathcal{H}_{T-1}, \mathbf{x}^{0}_{2,T}} \left[ 2\beta^{1/2}_T\sigma_{T-1}(\mathbf{x}_{1,T}, \mathbf{x}_{2,T})\right]\\
    &\stackrel{\text{(e)}}{=} \mathbb{E}_{\mathcal{H}_{T-1}, \mathbf{x}^{0}_{2,T}}\left[\sum^T_{t=1}2\beta^{1/2}_t\sigma_{t-1}(\mathbf{x}_{1,t}, \mathbf{x}_{2,t}) \right]\\
    &\stackrel{\text{(f)}}{\leq} \mathbb{E}_{\mathcal{H}_{T-1}, \mathbf{x}^{0}_{2,T}} \left[\sqrt{C_1 T \beta_T \gamma_T}\right]\\
    &\stackrel{\text{(g)}}{=} \sqrt{C_1 T \beta_T \gamma_T}
\end{split}
\label{proof_cum_regret}
\end{equation}
in which $C_1=8/\log (1 + \sigma^{-2}_1)$, $\beta_T$ is defined in Lemma~\ref{lemma:high_prob}, and $\gamma_T$ is the maximum information gain about the function $f_1$ obtained from any set of observations of size $T$.
Steps (a) and (e) both result from the fact that $r_{1,t}$ only depends on the level-$0$ strategy of iteration $t$ and the history up to iteration $t-1$ (through the level-$0$ strategy of iteration $t$),
and is thus independent of those input actions and output observations in future iterations $t+1,\ldots,T$.
(b) and (d) both follow from the law of total expectation, $(c)$ results from~\eqref{proof_determ}, (f) follows from Lemmas $5.3$ and $5.4$ of~\citet{srinivas2009gaussian},
(g) follows since all terms inside the expectation are independent of the history of input-output pairs.
Note that the expectation in~\eqref{proof_cum_regret} is taken over the history of selected actions and observed payoffs of $\mathcal{D}$.
Note that an upper bound on the regret can be easily derived using the upper bound on the expected regret~\eqref{proof_cum_regret} through Markov's inequality,
which suggests that level-$1$ reasoning achieves no regret asymptotically.

Of note, in the scenario in which more than two ($M > 2$) agents are present (Appendix~\ref{subsec:more_than_two_players}), 
with the modified level-$1$ policy given by~\eqref{eq:level_1_defender_more_than_2}, 
the proofs of~\eqref{proof_determ} and~\eqref{proof_cum_regret} still go through by simply replacing $\mathbf{x}^{0}_{2,t}$ with the 
concatenated vector of $[\mathbf{x}^{0}_{2,t},\ldots,\mathbf{x}^{0}_{M,t}]$ (i.e., the concatenation of the level-$0$ actions of all other agents) in every step of the proof.
Similarly, the expectation of the regret would be taken over the history of input-output pairs of all other agents $2,\ldots,M$.

\subsection{Theorem~\ref{theorem_level_k}}
For level-$k\geq 2$ reasoning, i.e., when $\mathcal{A}$ reasons at level $k$ (for $k\geq 2$) and $\mathcal{D}$ 
reasons at level $k'=k-1 \geq 1$,
the regret of $\mathcal{A}$ in iteration $t$ can be analyzed as:
\begin{equation}
\begin{split}
    r_{1,t}&=f_1\left(\mathbf{x}_1^*, \mathbf{x}_{2,t}\right)-f_1\left(\mathbf{x}_{1,t}, \mathbf{x}_{2,t}\right)\\
    &=f_1\left(\mathbf{x}_1^*,\mathbf{x}^{k-1}_{2,t}\right) - f_1\left(\mathbf{x}^{k}_{1,t}, \mathbf{x}^{k-1}_{2,t}\right) \\
    &\stackrel{\text{(a)}}{\leq} \alpha_{1,t}\left(\mathbf{x}_1^*,\mathbf{x}^{k-1}_{2,t}\right) - f_1\left(\mathbf{x}^{k}_{1,t}, \mathbf{x}^{k-1}_{2,t}\right) \\
    &\stackrel{\text{(b)}}{\leq} \alpha_{1,t}\left(\mathbf{x}^{k}_{1,t}, \mathbf{x}^{k-1}_{2,t}\right) - f_1\left(\mathbf{x}^{k}_{1,t}, \mathbf{x}^{k-1}_{2,t}\right)\\
    &\leq 2\beta^{1/2}_t\sigma_{t-1}(\mathbf{x}_{1,t}, \mathbf{x}_{2,t})
\end{split}
\label{proof_determ_k}
\end{equation}
in which (a) follows from Lemma~\ref{lemma:high_prob}, 
(b) results from the fact that $\mathbf{x}^{k}_{1,t}$ is selected by maximizing the GP-UCB acquisition function $\alpha$ 
with respect to $\mathbf{x}^{k-1}_{2,t}$ according to~\eqref{eq:determ_best_response}.
\eqref{proof_determ_k} also holds with probability of at least $1 - \delta$.

Next, the external regret can be upper bounded in a similar way as~\eqref{proof_cum_regret}:
\begin{equation}
\begin{split}
    R_{1,T} &= \sum^T_{t=1}r_{1,t} \stackrel{\text{(a)}}{\leq} \sum^T_{t=1}2\beta^{1/2}_t\sigma_{t-1}(\mathbf{x}_{1,t}, \mathbf{x}_{2,t}) \stackrel{\text{(b)}}{\leq} \sqrt{C_1 T \beta_T \gamma_T}
\end{split}
\label{proof_cum_regret_level_k}
\end{equation}
in which (a) results from~\eqref{proof_determ_k}, and (b) again follows from Lemmas $5.3$ and $5.4$ of~\citet{srinivas2009gaussian}.

\section{Proof of Theorem~\ref{theorem_borr_lite}}
\label{app:proof_borr_lite}
Note that the level-$1$ action selected by $\mathcal{A}$ (the attacker) following R2-B2-Lite~\eqref{eq:borr_light_level_1} is stochastic, instead of being deterministic as in R2-B2~\eqref{eq:level_1_defender}.
In the following, we denote the level-$1$ action of $\mathcal{A}$ following R2-B2-Lite as $\mathbf{x}^{1}_{1,t}(\widetilde{\mathbf{x}}^{0}_{2,t})$ since, 
conditioned on all the game history up to iteration $t-1$, 
the selected level-$1$ action is a deterministic function of $\mathcal{A}$'s simulated action of $\mathcal{D}$ (the defender) at level $0$ ($\widetilde{\mathbf{x}}^{0}_{2,t}$).
Note that, in contrast to the corresponding definition in Appendix~\ref{app:proof_theorem_2}, the history of game plays $\mathcal{H}'_{t-1}$ we define here additionally includes $\mathcal{A}$'s simulated action of $\mathcal{D}$ in every iteration:
$\mathcal{H}'_{t-1}=[\mathbf{x}^{0}_{2,1}, \widetilde{\mathbf{x}}^{0}_{2,1}, y^{0}_{2,1}, \mathbf{x}^{0}_{2,2}, \widetilde{\mathbf{x}}^{0}_{2,2}, y^{0}_{2,2}, \ldots, \mathbf{x}^{0}_{2,t-1}, \widetilde{\mathbf{x}}^{0}_{2,t-1}, y^{0}_{2,t-1}]$.
We use $\Sigma_{2,t}$ to denote the covariance matrix of the level-$0$ mixed strategy of $\mathcal{D}$ in iteration $t$ ($\mathcal{P}_{2,t}$), and use $\textrm{Tr}(\Sigma_{2,t})$ to represent its trace.
As a result, the expected regret of $\mathcal{A}$ in iteration $t$ can be analyzed as:
\begin{align}
    \mathbb{E}_{\mathbf{x}^{0}_{2,t}, \widetilde{\mathbf{x}}^{0}_{2,t}}[&r_{1,t}|\mathcal{H}'_{t-1}]
    =\mathbb{E}_{\mathbf{x}^{0}_{2,t}, \widetilde{\mathbf{x}}^{0}_{2,t}}\left[f_1\left(\mathbf{x}_1^*,\mathbf{x}^{0}_{2,t}\right) - f_1\left(\mathbf{x}^{1}_{1,t}(\widetilde{\mathbf{x}}^{0}_{2,t}), \mathbf{x}^{0}_{2,t}\right)|\mathcal{H}'_{t-1}\right] \nonumber \\
    &\stackrel{\text{(a)}}{\leq} \mathbb{E}_{\mathbf{x}^{0}_{2,t}, \widetilde{\mathbf{x}}^{0}_{2,t}}\left[\alpha_{1,t}\left(\mathbf{x}_1^*,\mathbf{x}^{0}_{2,t}\right) - f_1\left(\mathbf{x}^{1}_{1,t}(\widetilde{\mathbf{x}}^{0}_{2,t}), \mathbf{x}^{0}_{2,t}\right)|\mathcal{H}'_{t-1}\right] \nonumber \\
    &\stackrel{\text{(b)}}{=} \mathbb{E}_{\mathbf{x}^{0}_{2,t}, \widetilde{\mathbf{x}}^{0}_{2,t}}\left[\alpha_{1,t}\left(\mathbf{x}_1^*,\widetilde{\mathbf{x}}^{0}_{2,t}\right) - f_1\left(\mathbf{x}^{1}_{1,t}(\widetilde{\mathbf{x}}^{0}_{2,t}), \mathbf{x}^{0}_{2,t}\right)|\mathcal{H}'_{t-1}\right] \nonumber \\
    &\stackrel{\text{(c)}}{\leq} \mathbb{E}_{\mathbf{x}^{0}_{2,t}, \widetilde{\mathbf{x}}^{0}_{2,t}}\left[\alpha_{1,t}\left(\mathbf{x}^{1}_{1,t}(\widetilde{\mathbf{x}}^{0}_{2,t}),\widetilde{\mathbf{x}}^{0}_{2,t}\right) - f_1\left(\mathbf{x}^{1}_{1,t}(\widetilde{\mathbf{x}}^{0}_{2,t}), \mathbf{x}^{0}_{2,t}\right)|\mathcal{H}'_{t-1}\right]\nonumber \\
    &\stackrel{\text{(d)}}{=} \mathbb{E}_{\mathbf{x}^{0}_{2,t}, \widetilde{\mathbf{x}}^{0}_{2,t}}\bigg[\underline{\alpha_{1,t}\left(\mathbf{x}^{1}_{1,t}(\widetilde{\mathbf{x}}^{(0, 1)}_{2,t}),\widetilde{\mathbf{x}}^{0}_{2,t}\right) 
    - \alpha_{1,t}\left(\mathbf{x}^{1}_{1,t}(\widetilde{\mathbf{x}}^{0}_{2,t}),\mathbf{x}^{0}_{2,t}\right)}\nonumber \\
    &\quad + \underline{\alpha_{1,t}\left(\mathbf{x}^{1}_{1,t}(\widetilde{\mathbf{x}}^{0}_{2,t}),\mathbf{x}^{0}_{2,t}\right)
    - f_1\left(\mathbf{x}^{1}_{1,t}(\widetilde{\mathbf{x}}^{0}_{2,t}), \mathbf{x}^{0}_{2,t}\right)}|\mathcal{H}'_{t-1}\bigg]\nonumber \\
    &\stackrel{\text{(e)}}{\leq} \mathbb{E}_{\mathbf{x}^{0}_{2,t}, \widetilde{\mathbf{x}}^{0}_{2,t}}\left[ L_{\alpha_1} \norm{\widetilde{\mathbf{x}}^{0}_{2,t} - \mathbf{x}^{0}_{2,t}}_2|\mathcal{H}'_{t-1}\right] \nonumber \\
    &\quad + \mathbb{E}_{\mathbf{x}^{0}_{2,t}, \widetilde{\mathbf{x}}^{0}_{2,t}}\left[\alpha_{1,t}\left(\mathbf{x}^{1}_{1,t}(\widetilde{\mathbf{x}}^{0}_{2,t}),\mathbf{x}^{0}_{2,t}\right)
    - f_1\left(\mathbf{x}^{1}_{1,t}(\widetilde{\mathbf{x}}^{0}_{2,t}), \mathbf{x}^{0}_{2,t}\right) |\mathcal{H}'_{t-1}\right]\nonumber \\
    &\stackrel{\text{(f)}}{\leq} \mathbb{E}_{\mathbf{x}^{0}_{2,t}, \widetilde{\mathbf{x}}^{0}_{2,t}}\left[ L_{\alpha_1} \sqrt{\norm{\widetilde{\mathbf{x}}^{0}_{2,t} - \mathbf{x}^{0}_{2,t}}^2_2}|\mathcal{H}'_{t-1}\right] + \mathbb{E}_{\mathbf{x}^{0}_{2,t}, \widetilde{\mathbf{x}}^{0}_{2,t}}\left[2\beta^{1/2}_t\sigma_{t-1}(\mathbf{x}^{1}_{1,t}(\widetilde{\mathbf{x}}^{0}_{2,t}), \mathbf{x}^{0}_{2,t})|\mathcal{H}'_{t-1}
    \right]\nonumber \\
    &\stackrel{\text{(g)}}{\leq} L_{\alpha_1}  \sqrt{\mathbb{E}_{\mathbf{x}^{0}_{2,t}, \widetilde{\mathbf{x}}^{0}_{2,t}}\left[\norm{\widetilde{\mathbf{x}}^{0}_{2,t} - \mathbf{x}^{0}_{2,t}}^2_2|\mathcal{H}'_{t-1}\right]} + \mathbb{E}_{\mathbf{x}^{0}_{2,t}, \widetilde{\mathbf{x}}^{0}_{2,t}}\left[2\beta^{1/2}_t\sigma_{t-1}(\mathbf{x}^{1}_{1,t}(\widetilde{\mathbf{x}}^{0}_{2,t}), \mathbf{x}^{0}_{2,t})|\mathcal{H}'_{t-1}
    \right]\nonumber \\
    &= L_{\alpha_1}  \sqrt{\mathbb{E}_{\mathbf{x}^{0}_{2,t}, \widetilde{\mathbf{x}}^{0}_{2,t}}\left[\left(\widetilde{\mathbf{x}}^{0}_{2,t} - \mathbf{x}^{0}_{2,t}\right)^{\top}\left(\widetilde{\mathbf{x}}^{0}_{2,t} - \mathbf{x}^{0}_{2,t}\right)|\mathcal{H}'_{t-1}\right]} + 
    \mathbb{E}_{\mathbf{x}^{0}_{2,t}, \widetilde{\mathbf{x}}^{0}_{2,t}}\left[2\beta^{1/2}_t\sigma_{t-1}(\mathbf{x}^{1}_{1,t}(\widetilde{\mathbf{x}}^{0}_{2,t}), \mathbf{x}^{0}_{2,t})|\mathcal{H}'_{t-1}
    \right]\nonumber \\
    &= L_{\alpha_1}  \sqrt{\mathbb{E}_{\mathbf{x}^{0}_{2,t}, \widetilde{\mathbf{x}}^{0}_{2,t}}\left[\left(\widetilde{\mathbf{x}}^{0}_{2,t}\right)^{\top}\left(\widetilde{\mathbf{x}}^{0}_{2,t}\right) + \left(\mathbf{x}^{0}_{2,t}\right)^{\top}\left(\mathbf{x}^{0}_{2,t}\right) - 2 \left(\widetilde{\mathbf{x}}^{0}_{2,t}\right)^{\top} \left(\mathbf{x}^{0}_{2,t}\right) |\mathcal{H}'_{t-1}\right]} +\nonumber \\
    &\quad \mathbb{E}_{\mathbf{x}^{0}_{2,t}, \widetilde{\mathbf{x}}^{0}_{2,t}}\left[2\beta^{1/2}_t\sigma_{t-1}(\mathbf{x}^{1}_{1,t}(\widetilde{\mathbf{x}}^{0}_{2,t}), \mathbf{x}^{0}_{2,t})|\mathcal{H}'_{t-1}
    \right]\nonumber \\
    &\stackrel{\text{(h)}}{=} L_{\alpha_1}  \sqrt{\mathbb{E}_{\widetilde{\mathbf{x}}^{0}_{2,t}}\left[\left(\widetilde{\mathbf{x}}^{0}_{2,t}\right)^{\top}\left(\widetilde{\mathbf{x}}^{0}_{2,t}\right)\right] + \mathbb{E}_{\mathbf{x}^{0}_{2,t}}\left[\left(\mathbf{x}^{0}_{2,t}\right)^{\top}\left(\mathbf{x}^{0}_{2,t}\right)\right] - 2 \mathbb{E}_{\widetilde{\mathbf{x}}^{0}_{2,t}}\left[\widetilde{\mathbf{x}}^{0}_{2,t}\right]^{\top} \mathbb{E}_{\mathbf{x}^{0}_{2,t}}\left[\mathbf{x}^{0}_{2,t} \right]} +\nonumber \\
    &\quad \mathbb{E}_{\mathbf{x}^{0}_{2,t}, \widetilde{\mathbf{x}}^{0}_{2,t}}\left[2\beta^{1/2}_t\sigma_{t-1}(\mathbf{x}^{1}_{1,t}(\widetilde{\mathbf{x}}^{0}_{2,t}), \mathbf{x}^{0}_{2,t})|\mathcal{H}'_{t-1}
    \right]\nonumber \\
    &\stackrel{\text{(i)}}{=} L_{\alpha_1}  \sqrt{\mathbb{E}_{\mathbf{x}^{0}_{2,t}}\left[\left(\mathbf{x}^{0}_{2,t}\right)^{\top}\left(\mathbf{x}^{0}_{2,t}\right)\right] + \mathbb{E}_{\mathbf{x}^{0}_{2,t}}\left[\left(\mathbf{x}^{0}_{2,t}\right)^{\top}\left(\mathbf{x}^{0}_{2,t}\right)\right] - 2 \mathbb{E}_{\mathbf{x}^{0}_{2,t}}\left[\mathbf{x}^{0}_{2,t}\right]^{\top} \mathbb{E}_{\mathbf{x}^{0}_{2,t}}\left[\mathbf{x}^{0}_{2,t} \right]} +\nonumber \\
    &\quad \mathbb{E}_{\mathbf{x}^{0}_{2,t}, \widetilde{\mathbf{x}}^{0}_{2,t}}\left[2\beta^{1/2}_t\sigma_{t-1}(\mathbf{x}^{1}_{1,t}(\widetilde{\mathbf{x}}^{0}_{2,t}), \mathbf{x}^{0}_{2,t})|\mathcal{H}'_{t-1}
    \right]\nonumber \\
    &= \sqrt{2} L_{\alpha_1}  \sqrt{\mathbb{E}_{\mathbf{x}^{0}_{2,t}}\left[\left(\mathbf{x}^{0}_{2,t}\right)^{\top}\left(\mathbf{x}^{0}_{2,t}\right)\right] -  \mathbb{E}_{\mathbf{x}^{0}_{2,t}}\left[\mathbf{x}^{0}_{2,t}\right]^{\top} \mathbb{E}_{\mathbf{x}^{0}_{2,t}}\left[\mathbf{x}^{0}_{2,t} \right]} + \mathbb{E}_{\mathbf{x}^{0}_{2,t}, \widetilde{\mathbf{x}}^{0}_{2,t}}\left[2\beta^{1/2}_t\sigma_{t-1}(\mathbf{x}^{1}_{1,t}(\widetilde{\mathbf{x}}^{0}_{2,t}), \mathbf{x}^{0}_{2,t})|\mathcal{H}'_{t-1}
    \right]\nonumber \\
    &\stackrel{\text{(j)}}{=} \sqrt{2} L_{\alpha_1} \sqrt{\textrm{Tr}\left(\Sigma_{2,t}\right)} + \mathbb{E}_{\mathbf{x}^{0}_{2,t}, \widetilde{\mathbf{x}}^{0}_{2,t}}\left[2\beta^{1/2}_t\sigma_{t-1}(\mathbf{x}^{1}_{1,t}(\widetilde{\mathbf{x}}^{0}_{2,t}), \mathbf{x}^{0}_{2,t})|\mathcal{H}'_{t-1}
    \right]\nonumber \\
    &\stackrel{\text{(k)}}{\leq} \sqrt{2} L_{\alpha_1}  \sqrt{\omega_t} + \mathbb{E}_{\mathbf{x}^{0}_{2,t}, \widetilde{\mathbf{x}}^{0}_{2,t}}\left[2\beta^{1/2}_t\sigma_{t-1}(\mathbf{x}^{1}_{1,t}(\widetilde{\mathbf{x}}^{0}_{2,t}), \mathbf{x}^{0}_{2,t})|\mathcal{H}'_{t-1}
    \right]
\label{proof:r2_b2_lite}
\end{align}
in which (a) results from Lemma~\ref{lemma:high_prob}; 
(b) holds because, conditioned on $\mathcal{H}_{t-1}$, $\mathbf{x}^{0}_{2,t}$ and $\widetilde{\mathbf{x}}^{(0, 1)}_{2,t}$ are sampled from the same distribution and thus identically distributed; 
(c) follows from the way in which $\mathbf{x}^{1}_{1,t}$ is selected using the R2-B2-Lite algorithm~\eqref{eq:borr_light_level_1}, i.e., by deterministically best-responding to $\widetilde{\mathbf{x}}^{0}_{2,t}$ in terms of the GP-UCB acquisition function; 
(d) simply subtracts and adds the same GP-UCB term; 
(e) follows from the Lipschitz continuity of the GP-UCB acquisition function, whose Lipschitz constant (denoted as $L_{\alpha_1}$) has been shown to be finite in~\cite{kim2019local}; 
(f) is a result of the definition of the GP-UCB acquisition function (Section~\ref{background}) and Lemma~\ref{lemma:high_prob};
(g) results from the concavity of the square root function;
(h) follows from the linearity of expectation and the fact that $\widetilde{\mathbf{x}}^{0}_{2,t}$ and $\mathbf{x}^{0}_{2,t}$ are independent;
(i) again results from the fact that $\widetilde{\mathbf{x}}^{0}_{2,t}$ and $\mathbf{x}^{0}_{2,t}$ are identically distributed;
(j) follows from the definition of $\Sigma_{2,t}$, i.e., the covariance matrix of the level-$0$ mixed strategy of the defender in iteration $t$;
(k) follows from our assumption in Theorem~\ref{theorem_borr_lite} that the trace of $\Sigma_{2,t}$ is upper-bounded by the sequence $\{\omega_t\}$ for all $t\geq1$.
Note that all expectations in~\eqref{proof:r2_b2_lite} are conditioned on $\mathcal{D}'_{t-1}$, and some of the conditioning are omitted to shorten the expression.

Next, the expected external regret can be upper-bounded in a similar way as~\eqref{proof_cum_regret}:
\begin{equation}
\begin{split}
    \mathbb{E}_{\mathbf{x}^{0}_{2,1}, \widetilde{\mathbf{x}}^{0}_{2,1}, y^{0}_{2,1},\ldots,\mathbf{x}^{0}_{2,T-1}, \widetilde{\mathbf{x}}^{0}_{2,T-1}, y^{0}_{2,T-1}, \mathbf{x}^{0}_{2,T}, \widetilde{\mathbf{x}}^{0}_{2,T}}[R_{1,T}] 
    &=\mathbb{E}_{\mathbf{x}^{0}_{2,1}, \widetilde{\mathbf{x}}^{0}_{2,1}, y^{0}_{2,1},\ldots,\mathbf{x}^{0}_{2,T-1}, \widetilde{\mathbf{x}}^{0}_{2,T-1}, y^{0}_{2,T-1}, \mathbf{x}^{0}_{2,T}, \widetilde{\mathbf{x}}^{0}_{2,T}}\left[\sum^T_{t=1}r_{1,t}\right] \\
    &\leq \sqrt{2} L_{\alpha_1}\sum^T_{t=1}\sqrt{\omega_t}+ \sqrt{C_1 T \beta_T \gamma_T}
\end{split}
\end{equation}

Note that compared with Theorem~\ref{theorem_level_1}, the expectation in Theorem~\ref{theorem_borr_lite} is additionally taken over $\mathcal{A}$'s simulated action of $\mathcal{D}$ in all iterations, i.e., 
$\widetilde{\mathbf{x}}^{0}_{2,1}, \ldots, \widetilde{\mathbf{x}}^{0}_{2,T}$.
Finally, Theorem~\ref{theorem_borr_lite} follows:
\begin{equation}
    \mathbb{E}[R_{1,T}] \leq \mathcal{O}\left(\sum^T_{t=1}\sqrt{\omega_t} + \sqrt{T \beta_T \gamma_T}\right)
\label{eq:cum_regret_final_borr_lite}
\end{equation}

Similar to the analysis of R2-B2, in the scenario where more than two ($M > 2$) agents are involved, with the modified level-$1$ R2-B2-Lite algorithm given by~\eqref{eq:borr_light_level_1_more_than_2}, 
the proofs given above still go through by simply replacing $\mathbf{x}^{0}_{2,t}$ with the concatenated vector of $[\mathbf{x}^{0}_{2,t},\ldots,\mathbf{x}^{0}_{M,t}]$ 
(and replacing $\widetilde{\mathbf{x}}^{0}_{2,t}$ with the concatenated vector of $[\widetilde{\mathbf{x}}^{0}_{2,t},\ldots,\widetilde{\mathbf{x}}^{0}_{M,t}]$)  
in every step of the proof.
Again, the expectation of the regret of agent $\mathcal{A}_1$ is taken over the history of input-output pairs of all other agents, 
as well as $\mathcal{A}_1$'s simulated level-$0$ actions of all other agents in every iteration.

\section{Proof of Theorems~\ref{theorem_level_1} and~\ref{theorem_level_k} for $M>2$ Agents}
\label{app:proof_multiplayer}
We prove here that the regret upper bound in Theorems~\ref{theorem_level_1} and~\ref{theorem_level_k} also hold in games with $M>2$ agents.
We only give the proof for level-$k\geq2$ strategy since the proofs for level-$0$ and level-$1$ strategies are straightforward as explained in Appendices~\ref{subsec:more_than_two_players} and~\ref{app:proof}.
For simplicity, we only focus on the scenario in which agent $\mathcal{A}_1$ reasons at level $2$, whereas all other agents reason at either level $0$ or level $1$.
However, the proof can be generalized to the settings in which agent $\mathcal{A}_1$ reasons at a higher level $k>2$.
Following the notations of Appendix~\ref{subsec:more_than_two_players}, the expected regret of $\mathcal{A}_1$ in iteration $t$ can be upper bounded as:
\begin{equation}
\begin{split}
    \mathbb{E}_{\mathbf{x}^{0}_{2,t},\ldots,\mathbf{x}^{0}_{M_0,t}}[r_{1,t}|\mathcal{H}_{t-1}]=&\mathbb{E}_{\mathbf{x}^{0}_{2,t},\ldots,\mathbf{x}^{0}_{M_0,t}}\left[f_1\big(\mathbf{x}_1^*, \mathbf{x}^{0}_{2,t},\ldots,\mathbf{x}^{0}_{M_0,t}, \mathbf{x}^{1}_{M_0+1,t}, \ldots, \mathbf{x}^{1}_{M,t}\right)-\\
    &f_1\left(\mathbf{x}_{1,t}, \mathbf{x}^{0}_{2,t},\ldots,\mathbf{x}^{0}_{M_0,t}, \mathbf{x}^{1}_{M_0+1,t}, \ldots, \mathbf{x}^{1}_{M,t}\right)
    |\mathcal{H}_{t-1}\big]\\
    &\leq \mathbb{E}_{\mathbf{x}^{0}_{2,t},\ldots,\mathbf{x}^{0}_{M_0,t}}\big[\alpha_{1,t}\left(\mathbf{x}_1^*, \mathbf{x}^{0}_{2,t},\ldots,\mathbf{x}^{0}_{M_0,t}, \mathbf{x}^{1}_{M_0+1,t}, \ldots, \mathbf{x}^{1}_{M,t}\right) - \\
    &\qquad f_1\left(\mathbf{x}^{1}_{2,t}, \mathbf{x}^{0}_{2,t},\ldots,\mathbf{x}^{0}_{M_0,t}, \mathbf{x}^{1}_{M_0+1,t}, \ldots, \mathbf{x}^{1}_{M,t}\right)|\mathcal{H}_{t-1}\big] \\
    &\leq\mathbb{E}_{\mathbf{x}^{0}_{2,t},\ldots,\mathbf{x}^{0}_{M_0,t}}\big[\alpha_{1,t}\left(\mathbf{x}^{2}_{1,t}, \mathbf{x}^{0}_{2,t},\ldots,\mathbf{x}^{0}_{M_0,t}, \mathbf{x}^{1}_{M_0+1,t}, \ldots, \mathbf{x}^{1}_{M,t}\right) - \\
    &\qquad f_1\left(\mathbf{x}^{1}_{2,t}, \mathbf{x}^{0}_{2,t},\ldots,\mathbf{x}^{0}_{M_0,t}, \mathbf{x}^{1}_{M_0+1,t}, \ldots, \mathbf{x}^{1}_{M,t}\right) |\mathcal{H}_{t-1}\big]\\
    &\leq \mathbb{E}_{\mathbf{x}^{0}_{2,t},\ldots,\mathbf{x}^{0}_{M_0,t}}\left[2\beta^{1/2}_t\sigma_{t-1}(\mathbf{x}^{2}_{1,t}, \mathbf{x}^{0}_{2,t},\ldots,\mathbf{x}^{0}_{M_0,t}, \mathbf{x}^{1}_{M_0+1,t}, \ldots, \mathbf{x}^{1}_{M,t}|\mathcal{H}_{t-1}\right]
\end{split}
\label{eq:proof_multiplayer}
\end{equation}
The proof given in~\eqref{eq:proof_multiplayer} is analogous to~\eqref{proof_determ}.
The key difference from~\eqref{proof_determ} is that in this case, the expectation here is taken over the level-$0$ strategies of those agents reasoning at level $0$,
i.e., $\mathcal{A}_2,\ldots,\mathcal{A}_{M_0}$.
In contrast, in~\eqref{proof_determ}, the expectation is only taken over the level-$0$ strategy of the single opponent reasoning at level $0$.

Note that if none of the other agents reason at level $0$, the expectation operator in~\eqref{eq:proof_multiplayer} can be dropped.
As a result,~\eqref{proof_cum_regret_level_k} can be directly used to show that the resulting upper bound on the regret is the same as that given in Theorem~\ref{theorem_level_k}.
On the other hand, if there exists at least $1$ level-$0$ agents, the expectation operator remains.
Therefore, the subsequent proof follows from~\eqref{proof_cum_regret} and the resulting regret upper bound becomes the same as that shown in Theorem~\ref{theorem_level_1},
except that the expectation of the regret is taken over the history of input-output pairs of all level-$0$ agents.

\section{More Experimental Details and Results}
\label{app:experiment}
All experiments are run on computers with 16 cores of Intel Xeon processor, 5 NVIDIA GTX1080 Ti GPUs, and a RAM of 256G.
\subsection{Synthetic Games}
\label{appendix_subsec:synth}
\subsubsection{$2$-Agent Synthetic Games}
\label{app:exp_synth_two_player}
\textbf{(a) Detailed Experimental Setting}\\
The payoff functions used in the synthetic games are sampled from GPs with the Squared Exponential kernel with length scale $0.1$. 
All payoff functions are defined on a $2$-dimensional grid of equally spaced points in $[0,1]^2$ with size $|\mathcal{X}_1| \times |\mathcal{X}_2|=100\times 100$. 
Therefore, the action spaces of agent $1$ and agent $2$ both consist of $|\mathcal{X}_1|=|\mathcal{X}_2|=100$ points.
For common-payoff games, we randomly sample a function $f_1$ from a GP on the domain $\mathcal{X}_1\times\mathcal{X}_2$ and 
set $f_2(\mathbf{x}_1, \mathbf{x}_2)=f_1(\mathbf{x}_1, \mathbf{x}_2)$ for all $\mathbf{x}_1\in \mathcal{X}_1$ and $\mathbf{x}_2\in \mathcal{X}_2$;
regarding general-sum games, we randomly and independently sample two functions, $f_1$ and $f_2$, from the same GP; 
as for constant-sum games, we draw a function $f_1$ from the GP, and set $f_2(\mathbf{x}_1, \mathbf{x}_2)=1-f_1(\mathbf{x}_1, \mathbf{x}_2)$ for all $\mathbf{x}_1\in \mathcal{X}_1$ and $\mathbf{x}_2\in \mathcal{X}_2$.
All payoff functions are scaled into the range $[0, 1]$.
Note that since the domain size is not excessively large, the level-$1$ action can be selected by solving~\eqref{eq:level_1_defender} exactly instead of approximately. 
The true GP hyperparameters, with which the synthetic payoff functions are sampled, are used as the GP hyperparameters.

\begin{figure}
    \centering
    \begin{subfigure}[b]{0.45\linewidth}
        \includegraphics[width=\linewidth]{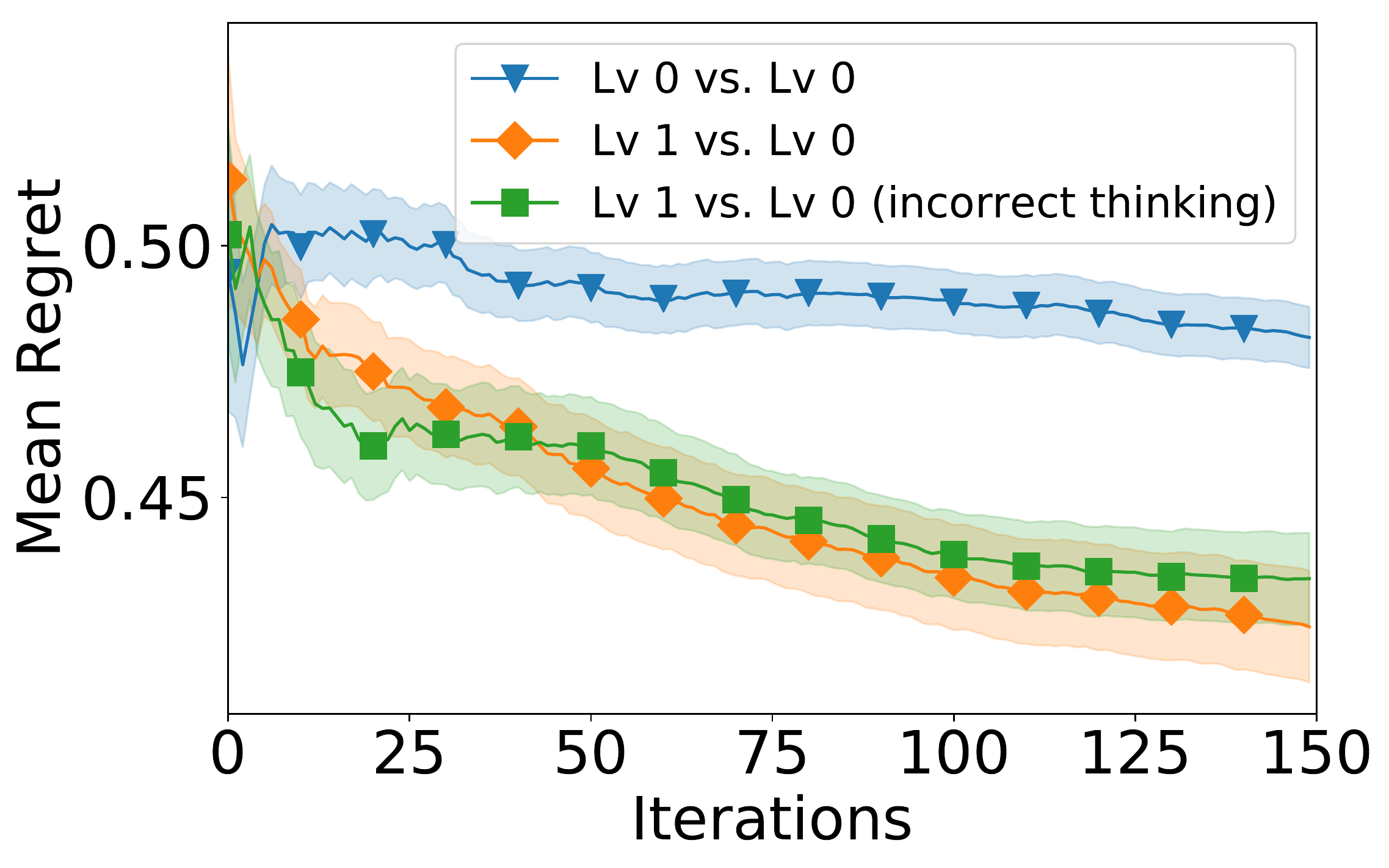}
        \caption{General-sum games.}
    \end{subfigure}
    \begin{subfigure}[b]{0.45\linewidth}
        \includegraphics[width=\linewidth]{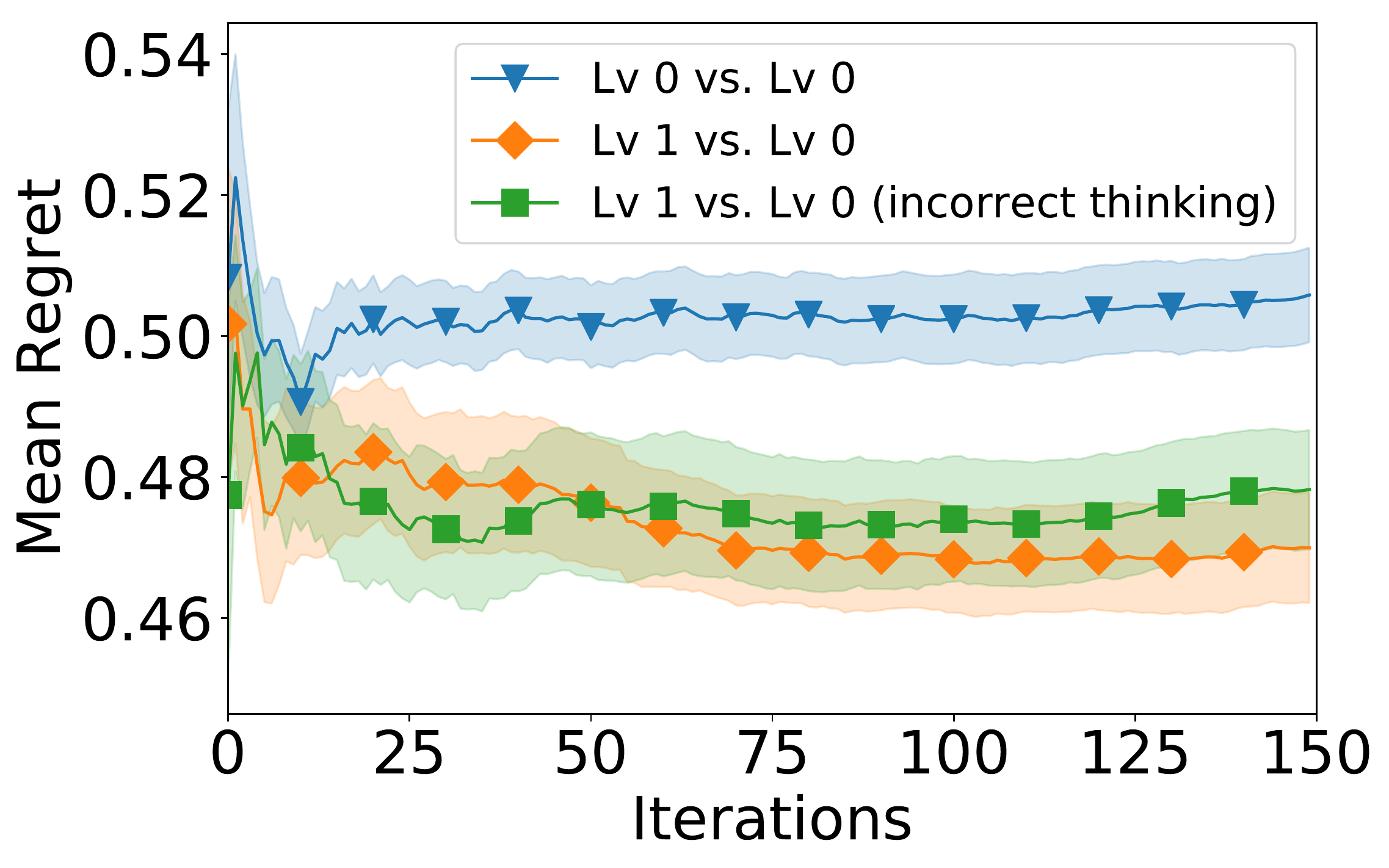}
        \caption{Constant-sum games.}
    \end{subfigure}
    \caption{Agent $1$'s performance of level-$1$ reasoning (agent $2$ reasons at level $0$) when agent $1$'s thinking about agent $2$'s level-$0$ strategy is incorrect. I.e., agent $2$ uses GP-MW as the level-$0$ strategy, while agent $1$ thinks that
    agent $2$ uses the random search level-$0$ strategy.}
    \label{fig:wrong_belief_gp_mw_as_random}
\end{figure}

\begin{figure}
    \centering
    \begin{subfigure}[b]{0.45\linewidth}
        \includegraphics[width=\linewidth]{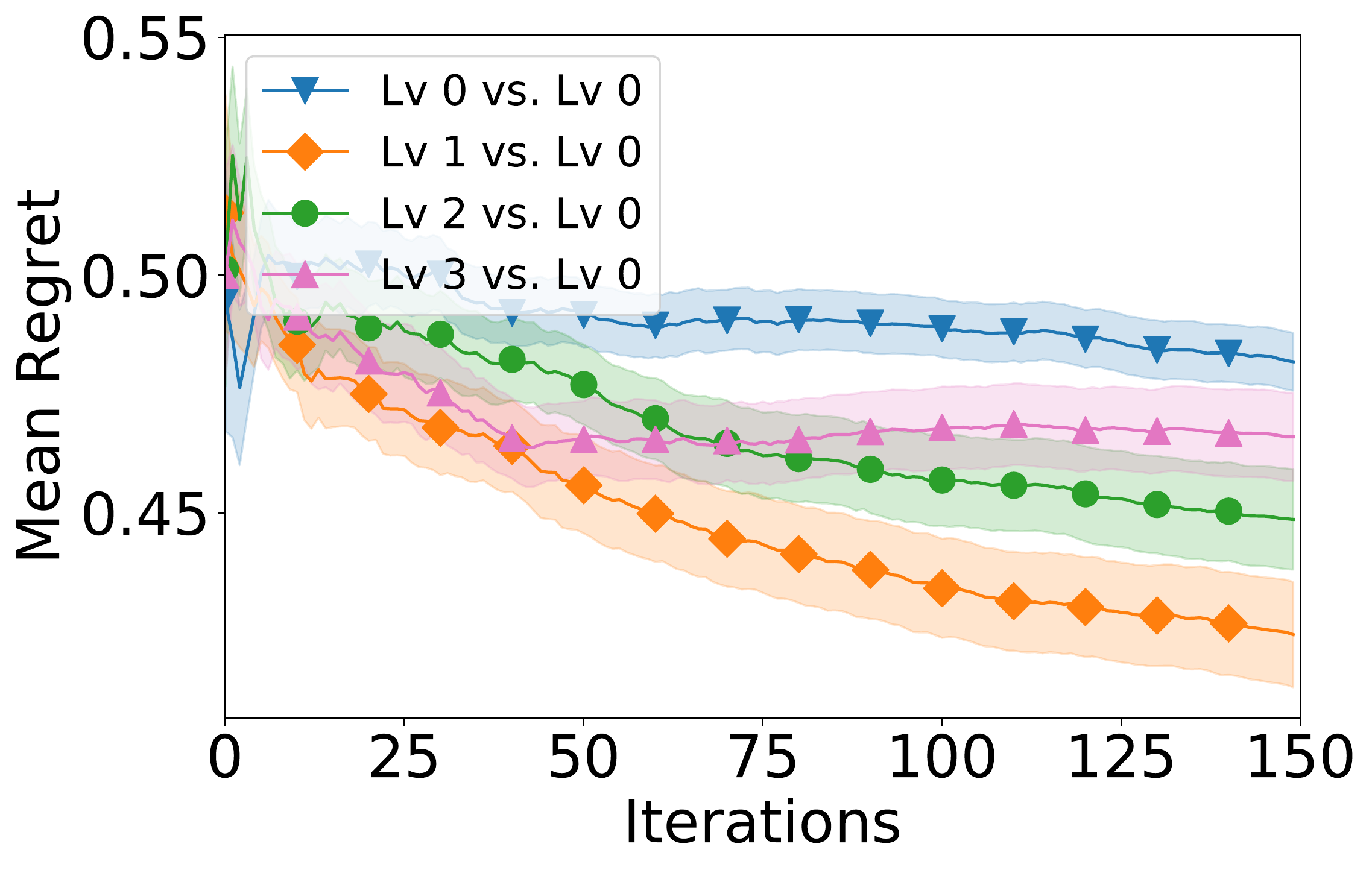}
        \caption{General-sum games.}
    \end{subfigure}
    \begin{subfigure}[b]{0.45\linewidth}
        \includegraphics[width=\linewidth]{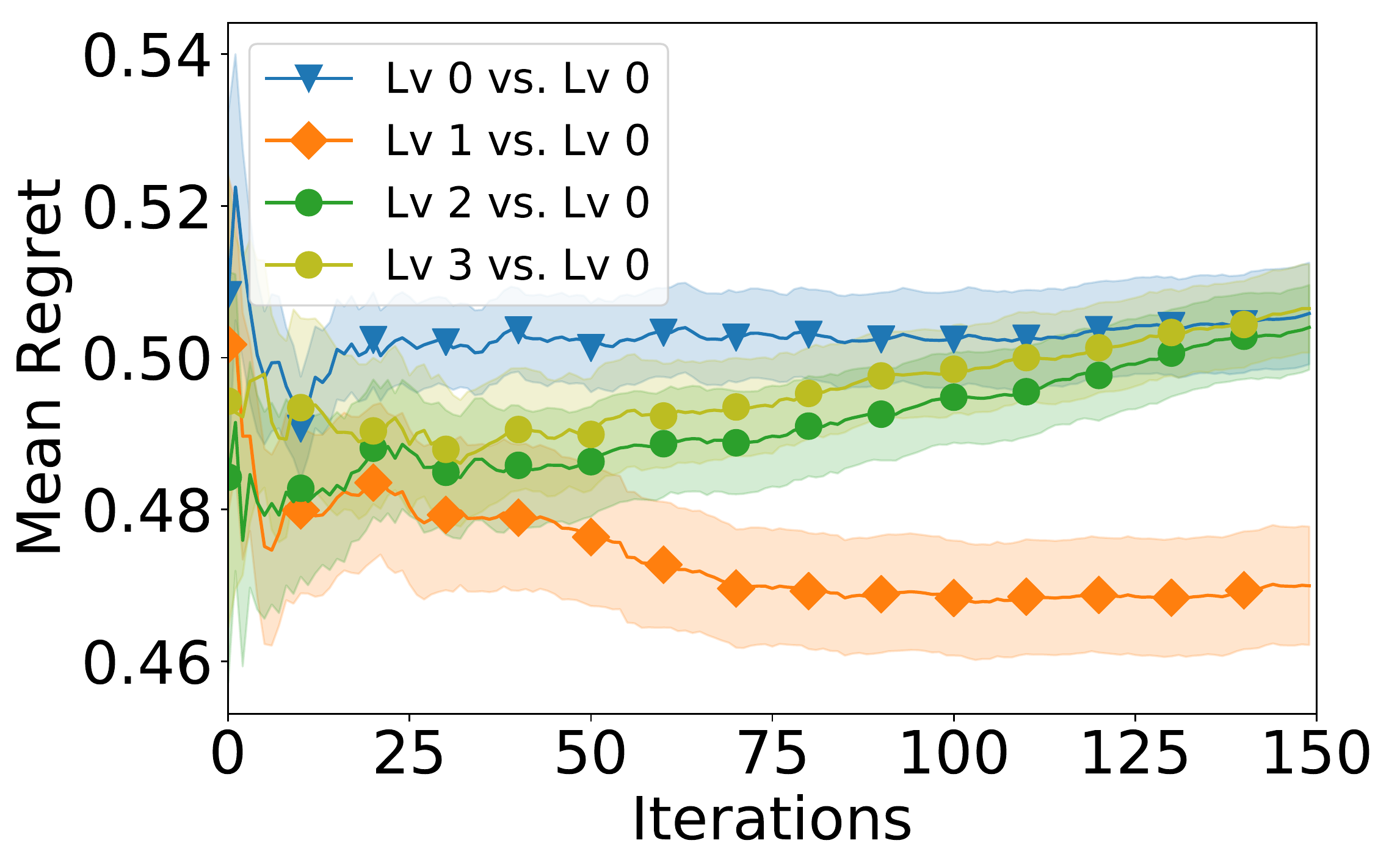}
        \caption{Constant-sum games.}
    \end{subfigure}
    \caption{Agent $1$'s performance when its thinking about agent $2$'s reasoning level is incorrect. That is, agent $2$ reasons at level $0$, while agent $1$ reasons at levels $1$, $2$ and $3$,
    where the last two settings result from agent $1$'s incorrect thinking about agent $2$'s reasoning level.}
    \label{fig:wrong_belief_against_level_0}
\end{figure}

\textbf{(b) More Results on the Impact of Incorrect Thinking about the Other Agent}\\
We further investigate how the performance of an agent is affected by incorrect thinking about the other agent.
Fig.~\ref{fig:wrong_belief_gp_mw_as_random} plots the performance of agent $1$ when agent $1$ and agent $2$ reason at levels $1$ and $0$ respectively, while agent $1$'s thinking about agent $2$'s
level-$0$ strategy is incorrect. The figures demonstrate that in the presence of an incorrect thinking about the other agent's level-$0$ strategy, 
the performance of agent $1$ only suffers from a marginal drop, although the theoretical guarantee offered by Theorem~\ref{theorem_level_1} no longer holds.
Fig.~\ref{fig:wrong_belief_against_level_0} illustrates the impacts of an incorrect thinking about the other agent's reasoning level.
As shown in the figure, when agent $2$'s reasoning level is fixed at level $0$, agent $1$ obtains the best performance when reasoning at level $1$, which agrees with our theoretical analysis
since by reasoning at level $1$, agent $1$'s performance is theoretically guaranteed (Theorem~\ref{theorem_level_1}).
Meanwhile, when agent $1$ reasons at a higher level (e.g., level $2$ or level $3$), the performance becomes worse (compared with reasoning at level $1$) yet is still better than reasoning at level $0$ (the blue curve);
this might be attributed to the fact that when agent $1$ reasons at level $2$ or $3$, even though agent $1$'s GP-UCB value is highly likely to be maximized with respect to the wrong action in every iteration~\eqref{eq:determ_best_response}, 
this could still help agent $1$ to eliminate some potentially ``dominated actions'', i.e., those actions which yield small GP-UCB values regardless of the action of agent $2$.
This ability to discard those dominated actions gives agent $1$ a preference to avoid selecting actions with small GP-UCB values, and
thus might help agent $1$ obtain a better performance compared with reasoning at level $0$.

\begin{figure}
    \centering
    \begin{subfigure}[b]{0.325\linewidth}
        \includegraphics[width=\linewidth]{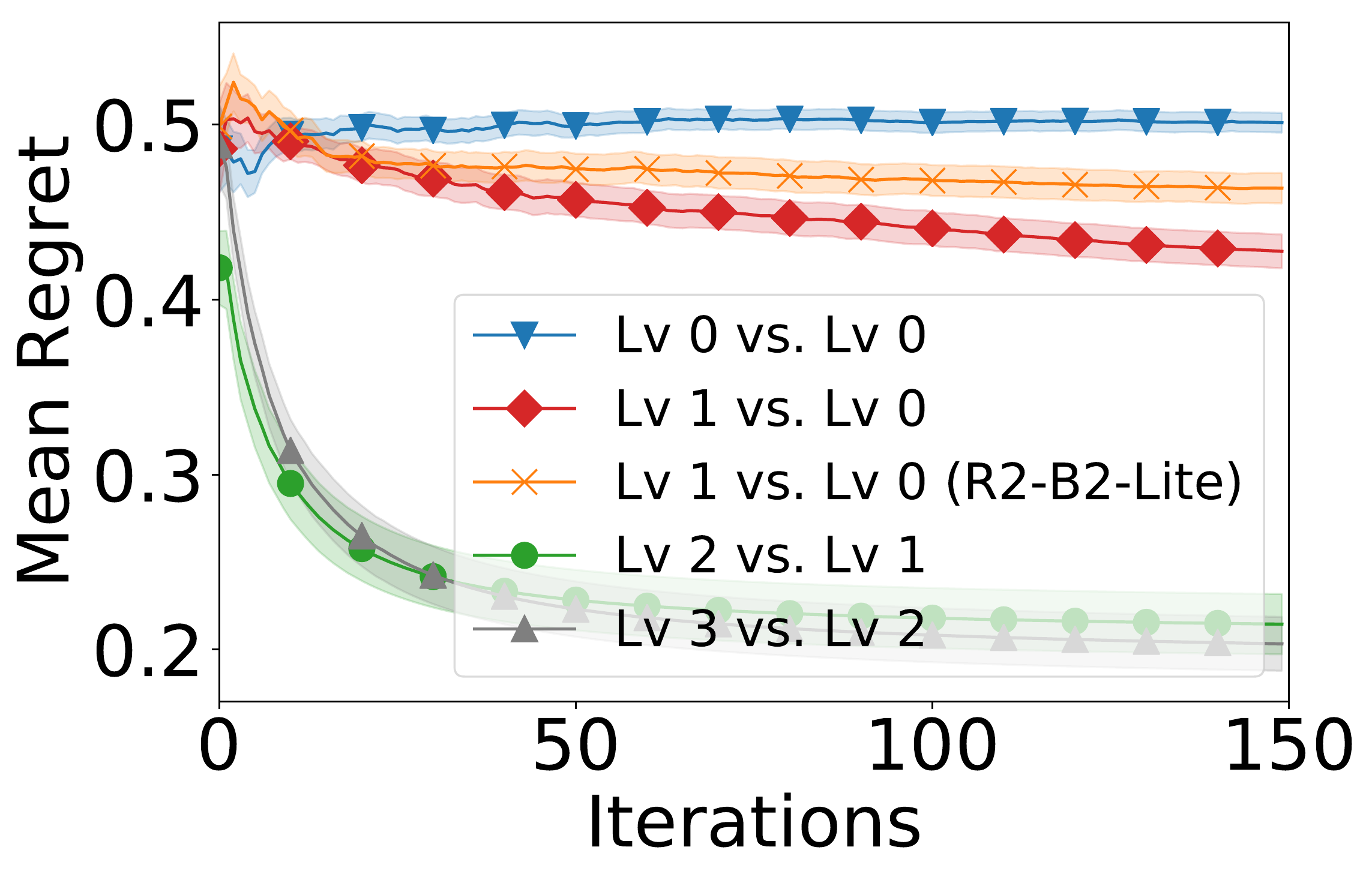}
        \caption{Common-payoff games.}
    \end{subfigure}
    \begin{subfigure}[b]{0.325\linewidth}
        \includegraphics[width=\linewidth]{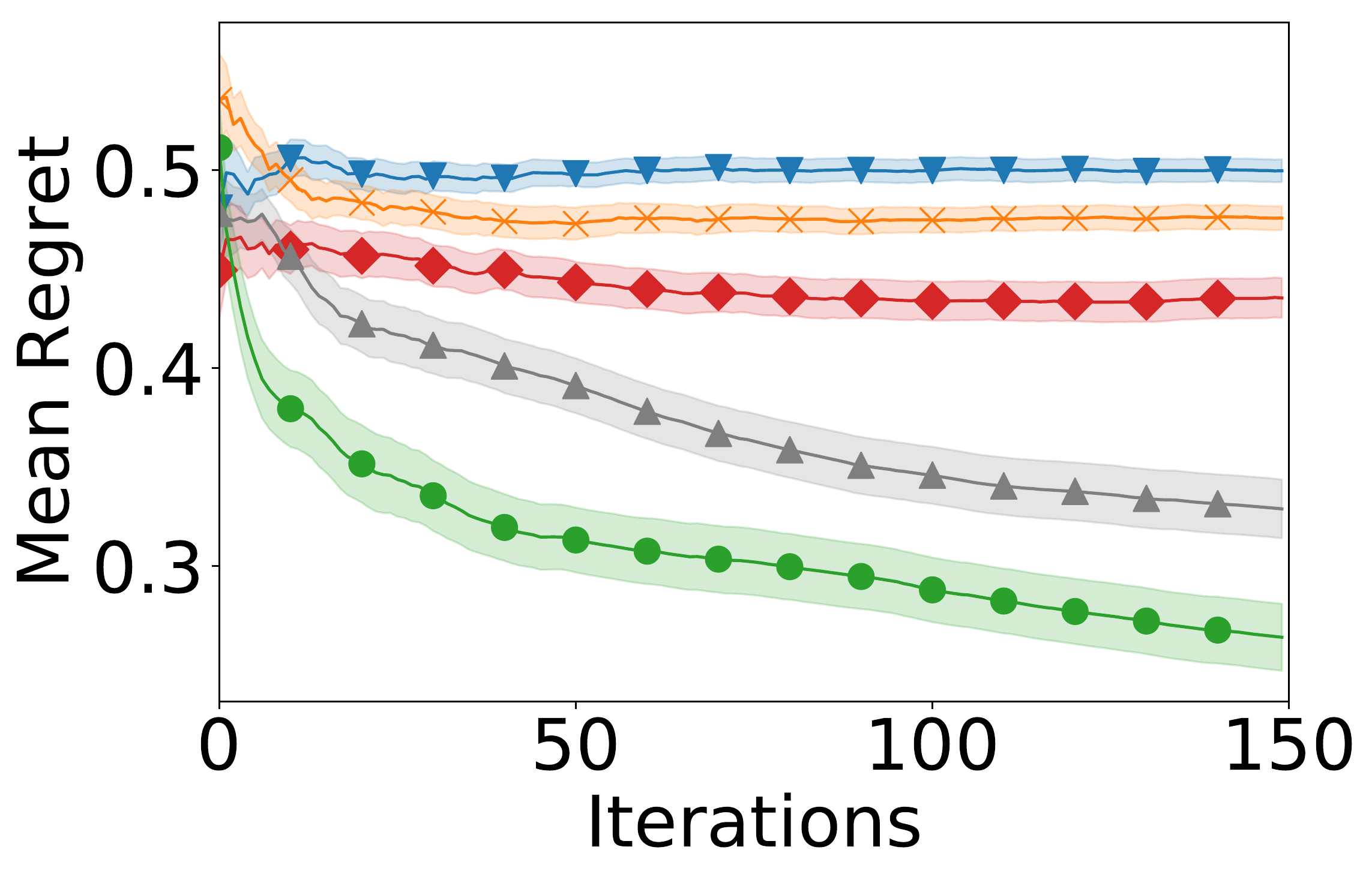}
        \caption{General-sum games.}
    \end{subfigure}
    \begin{subfigure}[b]{0.325\linewidth}
        \includegraphics[width=\linewidth]{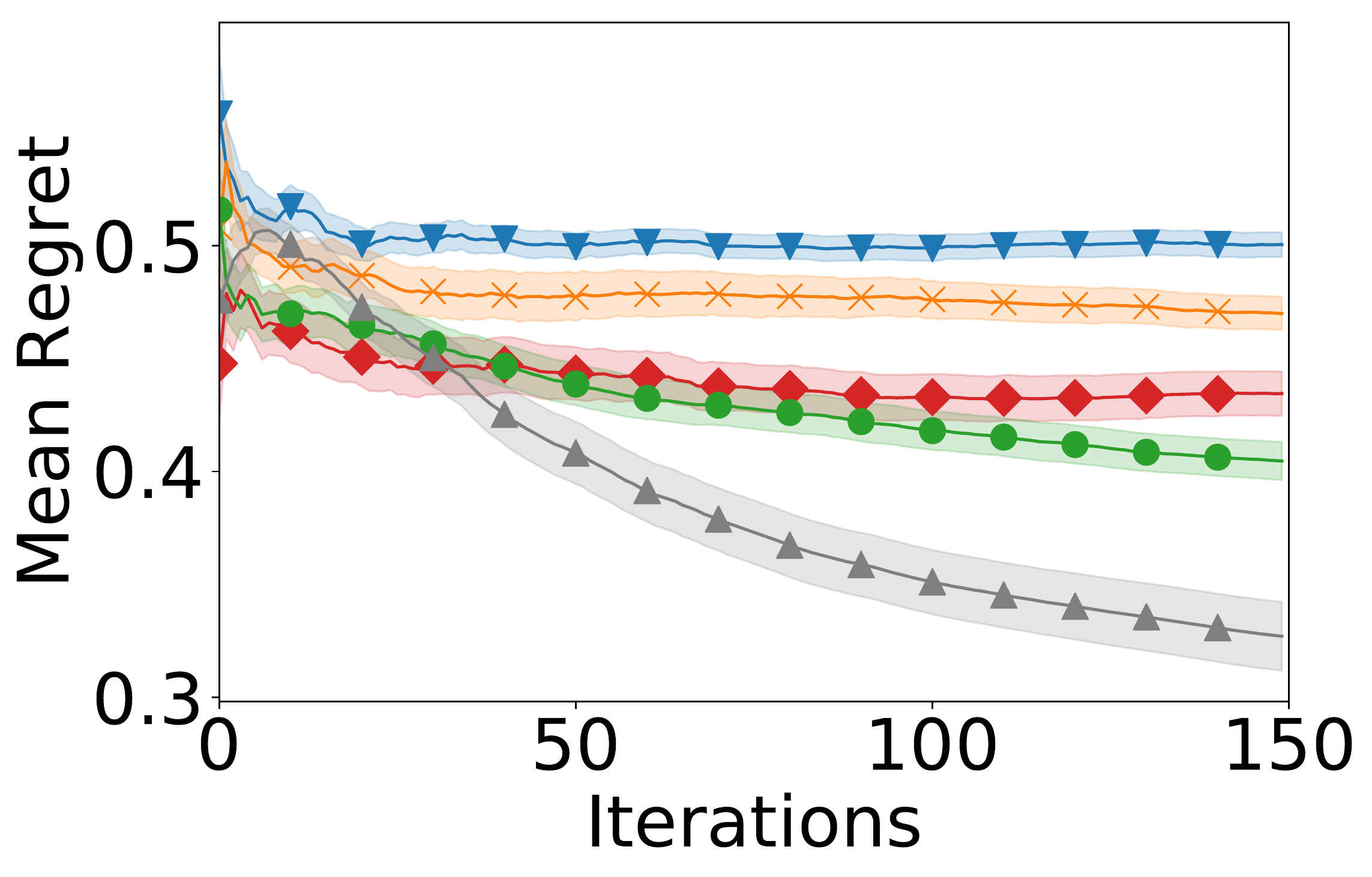}
        \caption{Constant-sum games.}
    \end{subfigure}
    \caption{Mean regret of agent $1$ in different types of synthetic games, with agent $2$ taking the random search level-$0$ strategy.}
    \label{fig:player_1_against_random}
\end{figure}

\begin{figure}
    \centering
    \begin{subfigure}[b]{0.325\linewidth}
        \includegraphics[width=\linewidth]{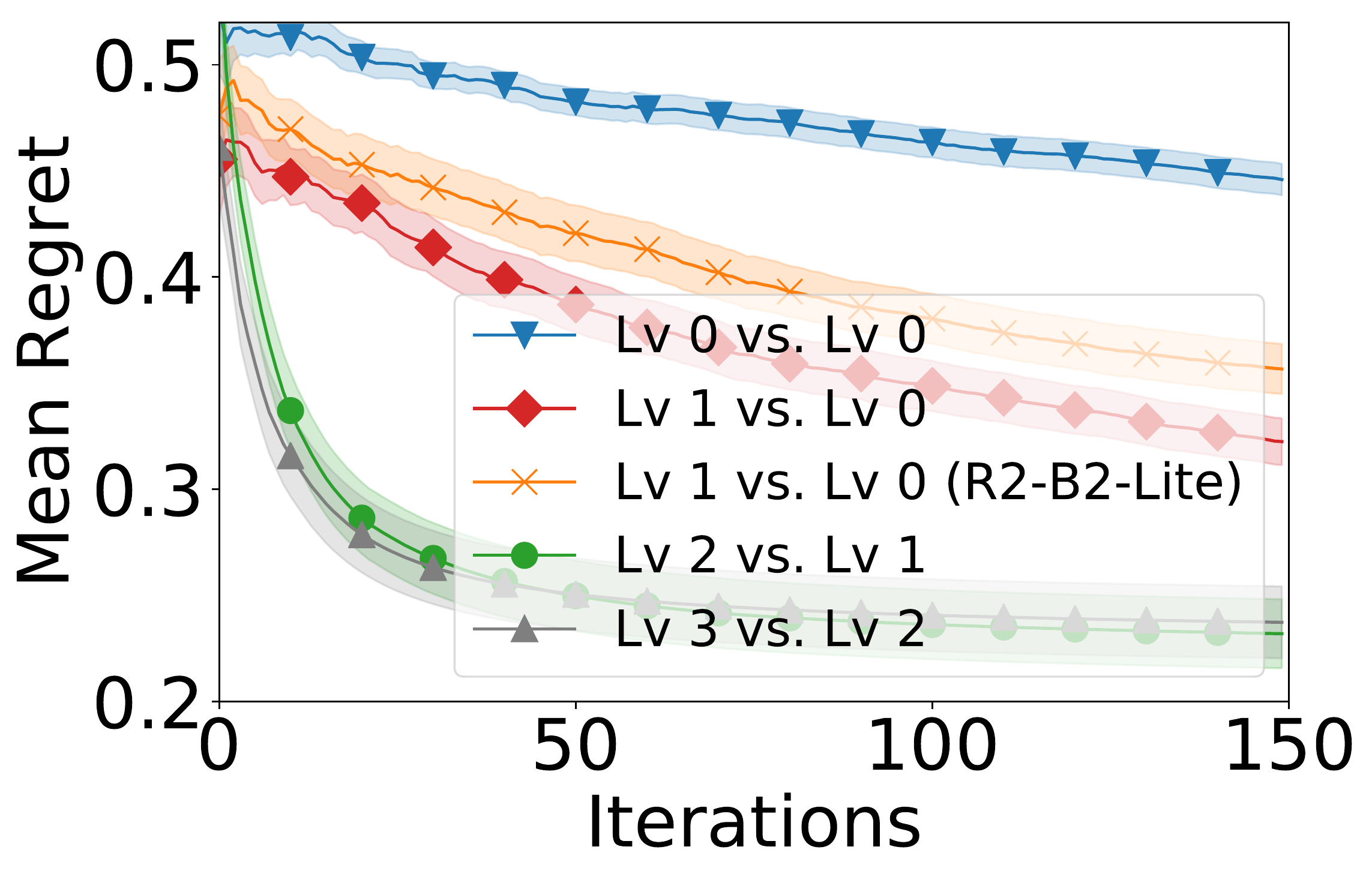}
        \caption{Common-payoff games.}
    \end{subfigure}
    \begin{subfigure}[b]{0.325\linewidth}
        \includegraphics[width=\linewidth]{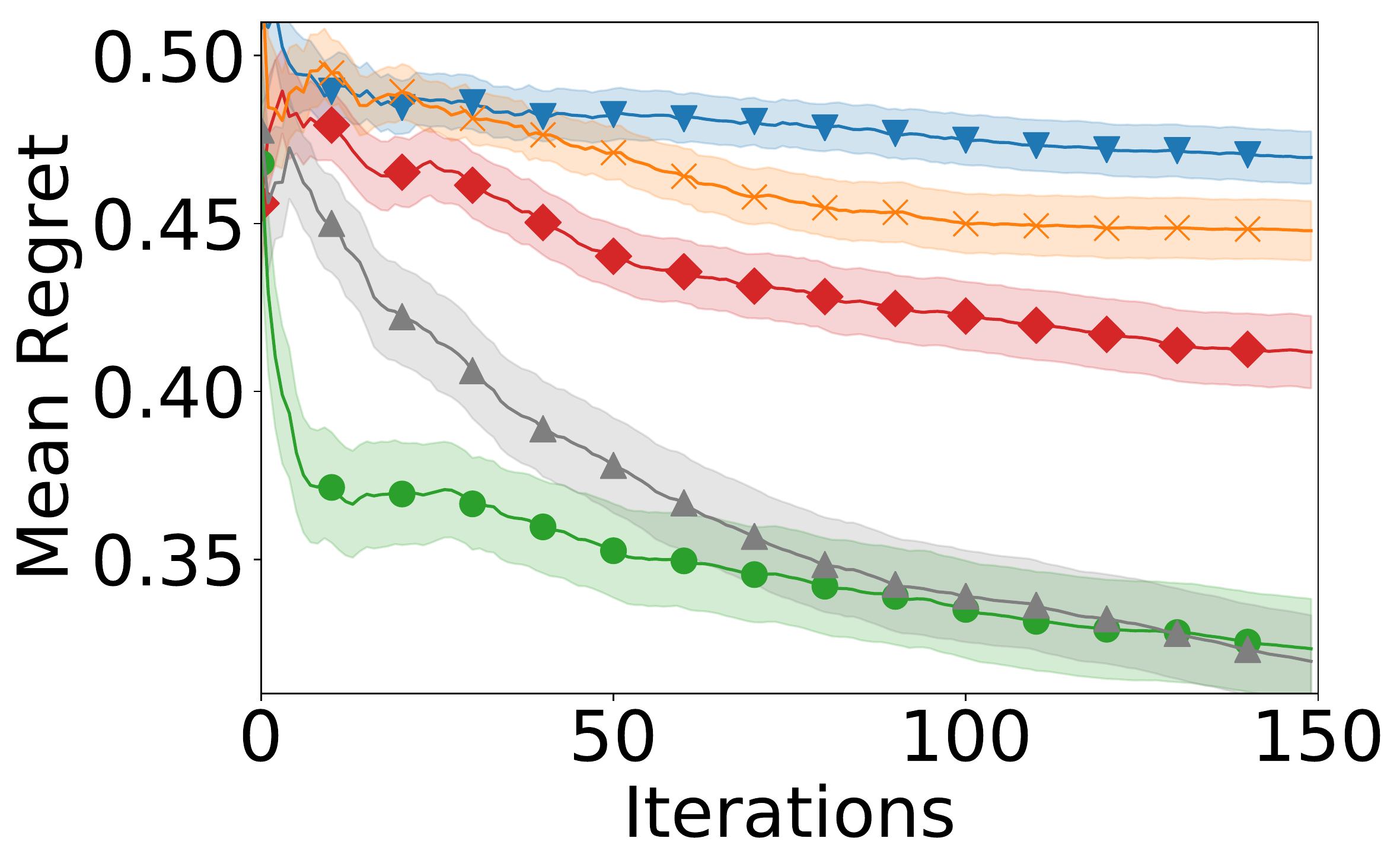}
        \caption{General-sum games.}
    \end{subfigure}
    \begin{subfigure}[b]{0.325\linewidth}
        \includegraphics[width=\linewidth]{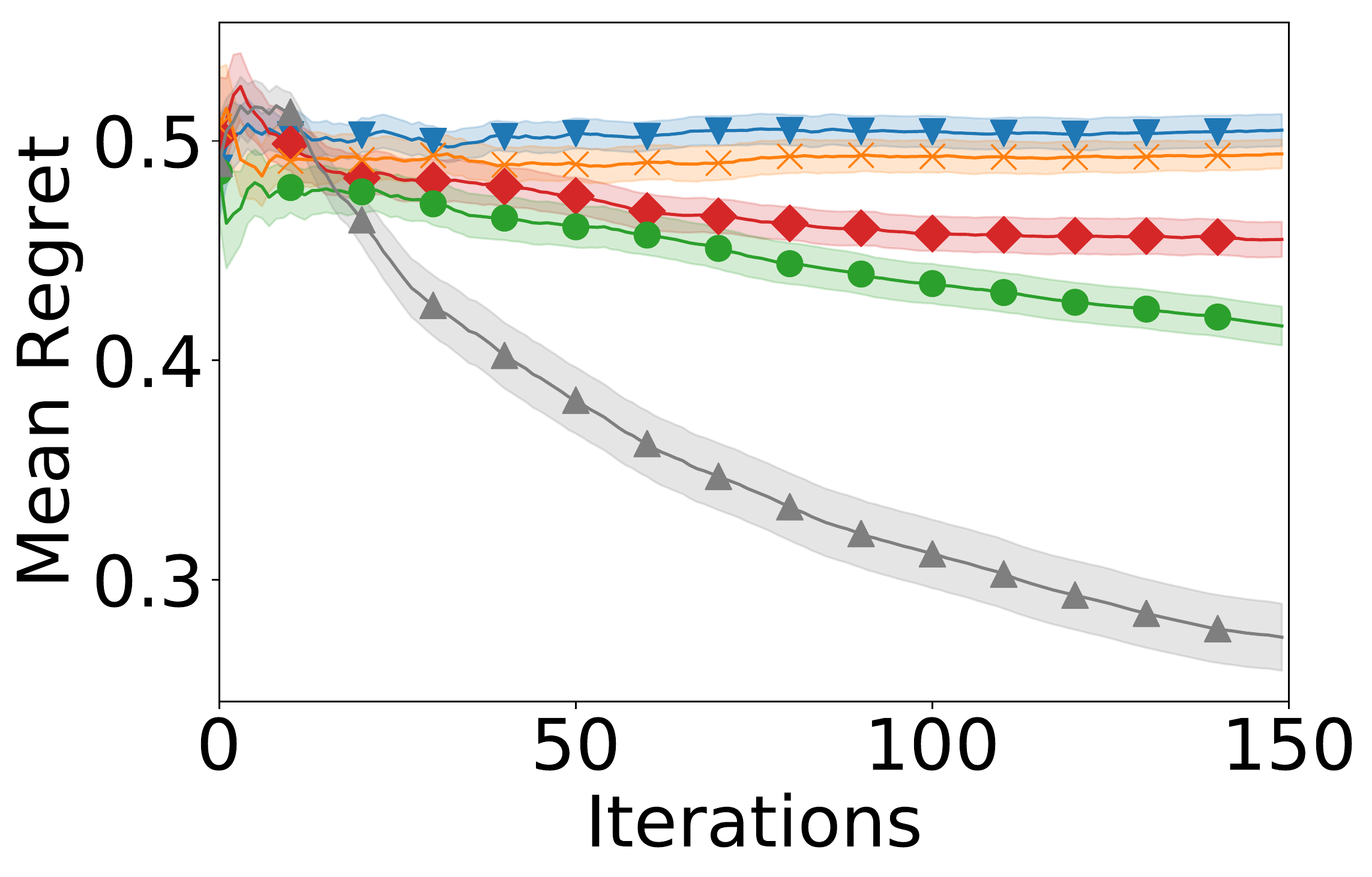}
        \caption{Constant-sum games.}
    \end{subfigure}
    \caption{Mean regret of agent $1$ in different types of synthetic games, with agent $2$ taking the EXP-$3$ level-$0$ strategy.}
    \label{fig:player_1_against_bandit}
\end{figure}

\textbf{(c) Results Using Other Level-$0$ Strategies}\\
In addition to the results presented in the main text which use GP-MW as the level-$0$ strategy (Fig.~\ref{fig:player_1}a to c), 
the entire set of experiments are repeated for the random search and EXP3 level-$0$ strategies, 
whose corresponding results are presented in Figs.~\ref{fig:player_1_against_random} and~\ref{fig:player_1_against_bandit}. 
These results yield the same observations and interpretations as Figs.~\ref{fig:player_1}a to c, and demonstrate the robustness of our R2-B2 algorithm 
with respect to the choice of the level-$0$ strategy.
Another interesting observation regarding different level-$0$ strategies is that in common-payoff and general-sum games, when both agents reason at level $0$, 
running a no-regret level-$0$ strategy (e.g., GP-MW or EXP3), instead of random search, leads to decreasing mean regret.
Specifically, when both agents reason at level $0$, the mean regret in common-payoff and general-sum games 
is decreasing if either GP-MW (Fig.~\ref{fig:player_1}a and b) or EXP3 (Fig.~\ref{fig:player_1_against_bandit}a and b) is used as the level-$0$ strategy
(with the decreasing trend more discernible in common-payoff games),
while the random search level-$0$ strategy results in a non-decreasing mean regret (Fig.~\ref{fig:player_1_against_random}a and b).
This observation demonstrates the benefit of adopting a better/more strategic level-$0$ strategy (instead of a non-strategic level-$0$ strategy such as random search) when reasoning at level $0$.

For the EXP3 level-$0$ strategy, we follow the practice of the work of~\citet{rahimi2008random}.
That is, we firstly draw $d'_1=5$ samples of $[\omega_i]_{i=1,\ldots,d'_1}$ from the spectral density of the GP kernel (i.e., the Squared Exponential kernel with length scale $0.1$),
and $d'_1$ samples of $[b_i]_{i=1,\ldots,d'_1}$ from the uniform distribution over $[0, 2\pi]$;
then, for every input $\mathbf{x}_1\in \mathcal{X}_1$ in the domain, we use $[\sqrt{2/d'_1}\cos(\omega_i \mathbf{x}_1 + b_i)]_{i=1,\ldots,d'_1}$ as the $d'_1$-dimensional feature representing $\mathbf{x}_1$.
Subsequently, the GP surrogate can be replaced with a linear surrogate model with the resulting features as inputs, and thus the EXP3 algorithm for adversarial linear bandit can be applied.

\subsubsection{Synthetic Games with $M>2$ Agents}
\label{app:exp_multi_player_games}
We also use synthetic games with $M>2$ agents to evaluate the effectiveness of our R2-B2 algorithm when more than two agents are involved.
We consider two types of synthetic games involving three agents. 
In the first type of games, the payoff functions of the three agents are independently sampled from a GP. 
The second type of games includes one adversary and two (cooperating) agents, the payoff function for the adversary, $f_1(\mathbf{x}_1, \mathbf{x}_2, \mathbf{x}_3)$, is a function sampled from a GP (and scaled to the range $[0, 1]$), 
whereas the payoff functions for the two agents are identical and defined as $1-f_1(\mathbf{x}_1, \mathbf{x}_2, \mathbf{x}_3)$.
We use GP-MW as the level-$0$ strategy.
\begin{figure}
    \centering
    \begin{subfigure}[b]{0.45\linewidth}
        \includegraphics[width=\linewidth]{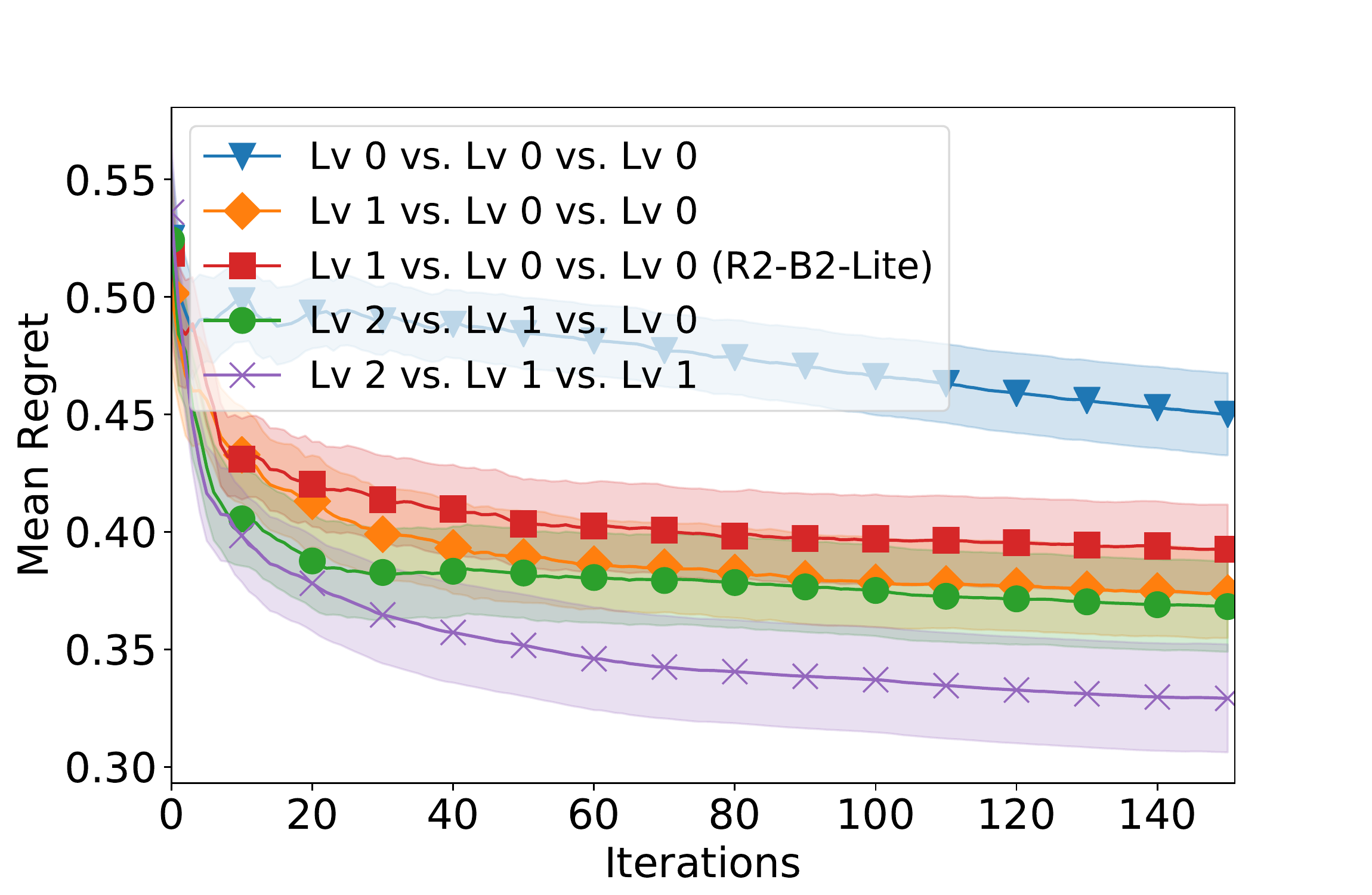}
        \caption{Mean regret of agent $1$ in the three-agent game with independent payoff functions. The reasoning levels are in the form of agent $1$ vs agent $2$ vs agent $3$.}
    \end{subfigure}
	\hspace{3mm}
    \begin{subfigure}[b]{0.45\linewidth}
        \includegraphics[width=\linewidth]{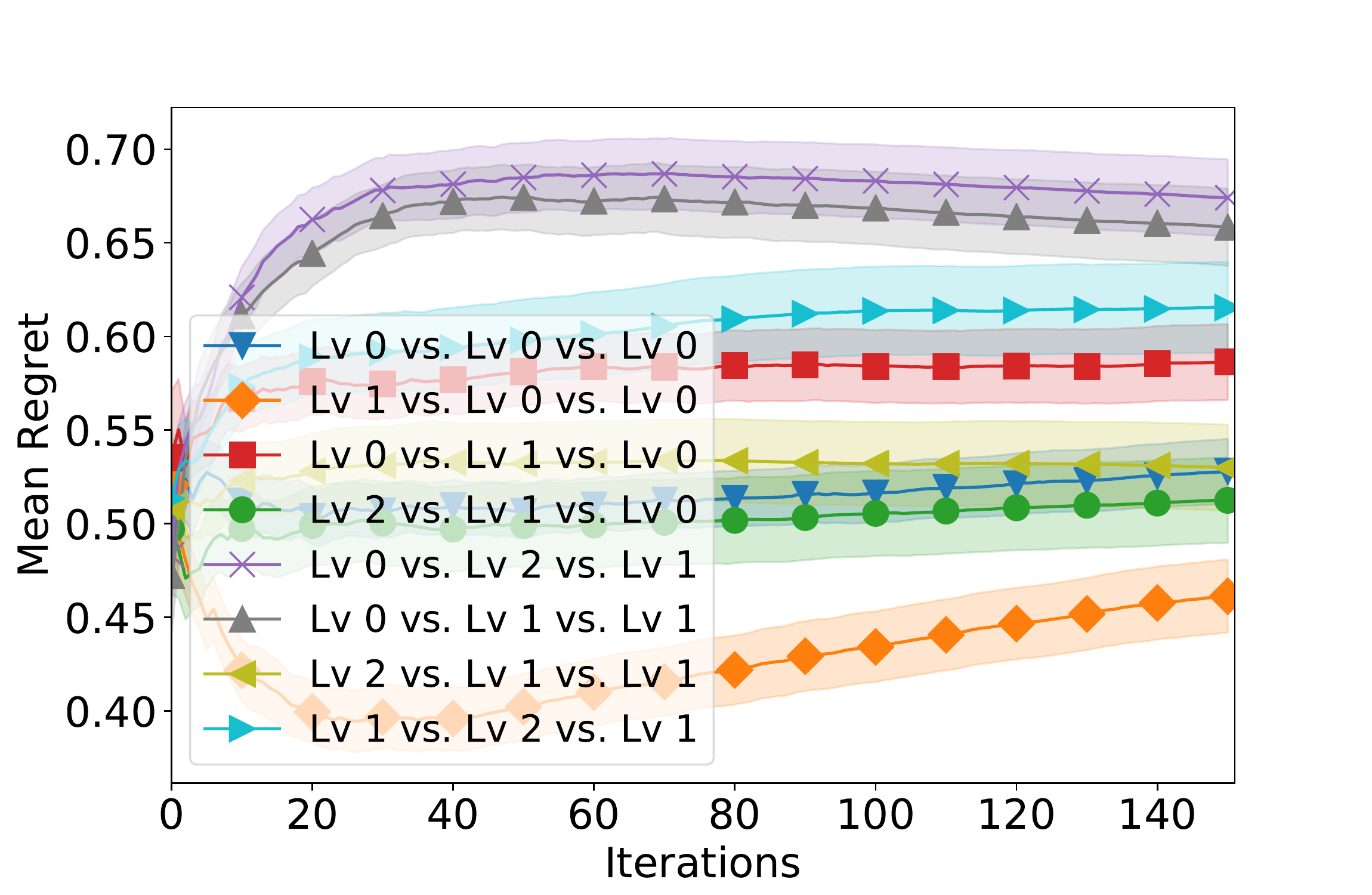}
        \caption{Mean regret of the adversary in the three-agent game with $1$ adversary and $2$ agents. The reasoning levels are in the form of adversary vs agent $1$ vs agent $2$.}
    \end{subfigure}
    \caption{Mean regret in three-agent games.}
    \label{fig:synth_multiplayer}
\end{figure}

Fig.~\ref{fig:synth_multiplayer}a displays the mean regret of agent $1$ in the first type of games, i.e., games with independent payoff functions.
The figure shows that in games with more than two agents, agent $1$ gains benefit by following the R2-B2 algorithm presented in Appendix~\ref{subsec:more_than_two_players}.
Specifically, the orange and red curves demonstrate the advantage of level-$1$ reasoning using R2-B2~\eqref{eq:level_1_defender_more_than_2} and 
R2-B2-Lite~\eqref{eq:borr_light_level_1_more_than_2} respectively, 
and the green and purple curves illustrate the benefit of level-$k>2$ reasoning~\eqref{eq:level_2_defender_more_than_2}.

Fig.~\ref{fig:synth_multiplayer}b shows the mean regret of the adversary in the second type of games involving one adversary and two agents. 
Note that the mean regret of the two agents can be directly read from the figure since it is equal to $1-$ the mean regret of the adversary.
A number of interesting insights can be drawn from Fig.~\ref{fig:synth_multiplayer}.
Comparing the orange and blue curves (similarly the green and red curves, and the yellow and gray curves) shows that the adversary obtains smaller regret by reasoning at a higher level than both agents; 
similarly, comparison of the blue and red curves (as well as the blue vs the purple, gray, and cyan curves) demonstrates that both agents enjoy a smaller regret when at least one of them reasons at a higher level than the adversary; 
comparing the gray and red curves reveals that when both agents reason at a higher level (in contrast to when one of them reasons at a higher level), the agents benefit more in terms of regret; 
comparison of the cyan and purple curves shows that given that the two agents reason at levels $2$ and $1$ respectively, the adversary reduces its deficit in regret by reasoning at level $1$ instead of level $0$.

\subsection{Adversarial ML}
\subsubsection{R2-B2 for Adversarial ML}
\label{app:adv_ml}
\textbf{(a) Detailed Experimental Setting}\\
We focus on the standard black-box setting, i.e., both $\mathcal{A}$ (the attacker) and $\mathcal{D}$ (the defender) can only access the target ML model by querying the model 
and observing the corresponding predictive probabilities for different classes~\cite{tu2019autozoom}.
Query efficiency is of critical importance for a black-box attacker since each query of the target ML model can be costly and 
an excessive number of queries might lead to the risk of being detected. 
Similarly, when defending against an attacker who adopts a query-efficient algorithm, it is also reasonable for the defender to defend in a query-efficient manner.
This justifies the use of BO-based methods for both adversarial attack and defense methods, since BO has been repeatedly demonstrated to be sample-efficient~\cite{shahriari2016taking}
and has been successfully applied to black-box adversarial attacks~\cite{ru2020bayesopt}.
The GP hyperparameters are optimized by maximizing the marginal likelihood after every $10$ iterations.

Both the MNIST and CIFAR-$10$ datasets can be downloaded using the Keras package in Python\footnote{\url{https://keras.io/}}.
All pixel values of all images are normalized into the range $[0, 1]$.
For the MNIST dataset, we use a convolutional neural network (CNN) model\footnote{\url{https://github.com/keras-team/keras/blob/master/examples/mnist_cnn.py}} with $99.25\%$ validation accuracy (trained on $60,000$ samples and validated using $10,000$ samples) 
as the target ML model, and for CIFAR-10, we use a ResNet model\footnote{\url{https://github.com/keras-team/keras/blob/master/examples/cifar10_resnet.py}} with $92.32\%$ validation accuracy (trained using $50,000$ samples and validated on $10,000$ samples, data augmentation is used). 
All test images used in the experiments for attack/defense are randomly selected among those correctly classified images from the validation set.
To improve the query efficiency of black-box adversarial attacks, different dimensionality reduction techniques such as autoencoder have been adopted to reduce the dimensionality of image data~\cite{tu2019autozoom}.
In this work, we let both $\mathcal{A}$ and $\mathcal{D}$ use Variational Autoencoders (VAEs)~\cite{kingma2013auto} for dimensionality reduction in a realistic setting:
In every iteration of the repeated game, $\mathcal{A}$ encodes the test image into a low-dimensional latent vector (i.e., the mean vector of the encoded latent distribution) using a VAE, 
perturbs the vector, and then decodes the perturbed vector to obtained the resulting image with perturbations; 
next, $\mathcal{D}$ receives the perturbed image, uses a VAE to encode the perturbed image to obtain a low-dimensional latent vector (i.e., the mean vector of the encoded latent distribution), 
adds transformations (perturbations) to the latent vector, and finally decodes the vector into the final image to be passed as input to the target ML model.
In the experiments, the same VAE is used by both $\mathcal{A}$ and $\mathcal{D}$, but the use of different VAEs can be easily achieved.
The latent dimension (LD) is $d_1=d_2=2$ for MNIST and $d_1=d_2=8$ for CIFAR-$10$; the action space for both $\mathcal{A}$ and $\mathcal{D}$ (i.e., the space of allowed perturbations
to the latent vectors) is $[-2, 2]^{2}$ for MNIST, and $[-2, 2]^{8}$ for CIFAR-$10$.
For MNIST, the VAE\footnote{\url{https://github.com/keras-team/keras/blob/master/examples/variational_autoencoder.py}} is a multi-layer perceptron (MLP) with ReLU activation, 
in which the input image is flattened into a $28 \times 28$-dimensional vector and both the encoder and decoder consist of a $512$-dimensional hidden layer.
Regarding CIFAR-$10$, the encoder of the VAE uses $3$ convolutional layers followed by a fully connected layer, 
whereas the decoder uses $2$ fully connected layers followed by $3$ de-convolutional layers\footnote{\url{https://github.com/chaitanya100100/VAE-for-Image-Generation}}.

For both $\mathcal{A}$ and $\mathcal{D}$, the image produced by the decoder of their VAE is clipped such that the requirement of bounded perturbations in terms of the infinity norm (as mentioned in Section~\ref{subsec:adv_ml} of the main text) is satisfied.
We consider \emph{untargeted attacks} in this work, i.e., the attacker's (defender's) goal is to cause (prevent) misclassification of the ML model.
However, our framework can also deal with \emph{targeted attacks} (i.e., the attacker aims at causing the target ML model to misclassify a test image into a particular class) through slight modifications to the payoff functions.
The payoff function value for $\mathcal{A}$ ($f_1(\mathbf{x}_1, \mathbf{x}_2)$, referred to as the \emph{attack score}) for a pair of perturbations 
selected by $\mathcal{A}$ ($\mathbf{x}_1$) and $\mathcal{D}$ ($\mathbf{x}_2$) is the maximum predictive probability (corresponding to the probability that test input belongs to a class) 
among all \emph{incorrect classes}, which is bounded in $(0, 1)$. 
For example, in a $10$-class classification model (i.e., for both MNIST and CIFAR-10), if the correct/ground-truth class for a test image is $0$, 
the value of the payoff function for $\mathcal{A}$ is the maximum predictive probability among classes $1$ to $9$. 
The payoff function for $\mathcal{D}$ is $f_2(\mathbf{x}_1, \mathbf{x}_2)=1-f_1(\mathbf{x}_1, \mathbf{x}_2)$ 
since the defender attempts to make sure that the predictive probability of the correct class remains the largest by minimizing the maximum predictive probability among all incorrect classes. 

As reported in the main text (Section~\ref{subsec:adv_ml}), we use GP-MW and random search as the level-$0$ strategies for MNIST, and only use random search for CIFAR-$10$.
The reason is that GP-MW requires a discrete input domain (or a discretized continuous input domain) since it needs to maintain and update a discrete distribution over the input domain.
Therefore, it is difficult to apply GP-MW to a high-dimensional continuous input domain (e.g., the $8$-dimensional domain in the CIFAR-$10$ experiment) 
since an accurate discretization of the high-dimensional domain would lead to an intractably large domain for the discrete distribution,
making it intractable to update and sample from the distribution.
Similarly, the application of the EXP3 algorithm is also limited to low-dimensional input domains for the same reason.

\textbf{(b) Results Using Multiple Images}\\
Note that different images may be associated with different degrees of difficulty to attack and to defend, 
i.e., some images are easier to attack (and thus harder to defend) and others may be easier to defend (and thus harder to attack).
Therefore, for those images that are easier to attack than to defend, it is easier for the attacker to increase the attack score than for the defender to reduce the attack score; 
as a result, the advantage achieved by the defender (i.e., lower attack score) when the defender reasons at one level higher would be less discernible since the defender's task (i.e., to decrease the attack score) is more difficult.
On the other hand, for those images that are easier to defend than to attack (e.g., the MNIST dataset as demonstrated below), the benefit obtained by the attacker (i.e., higher attack score) when it reasons at one level higher would be harder to delineate
since the attacker's task of increasing the attack score is more difficult.
The image from MNIST/CIFAR-$10$ that is used to produce the results reported in the main text (Fig.~\ref{fig:player_1}d to f) is selected to ensure that 
the difficulties of attack and defense are comparable such that the effects of both attack and defense can be clearly illustrated.

\begin{figure}
    \centering
    \includegraphics[width=0.5\linewidth]{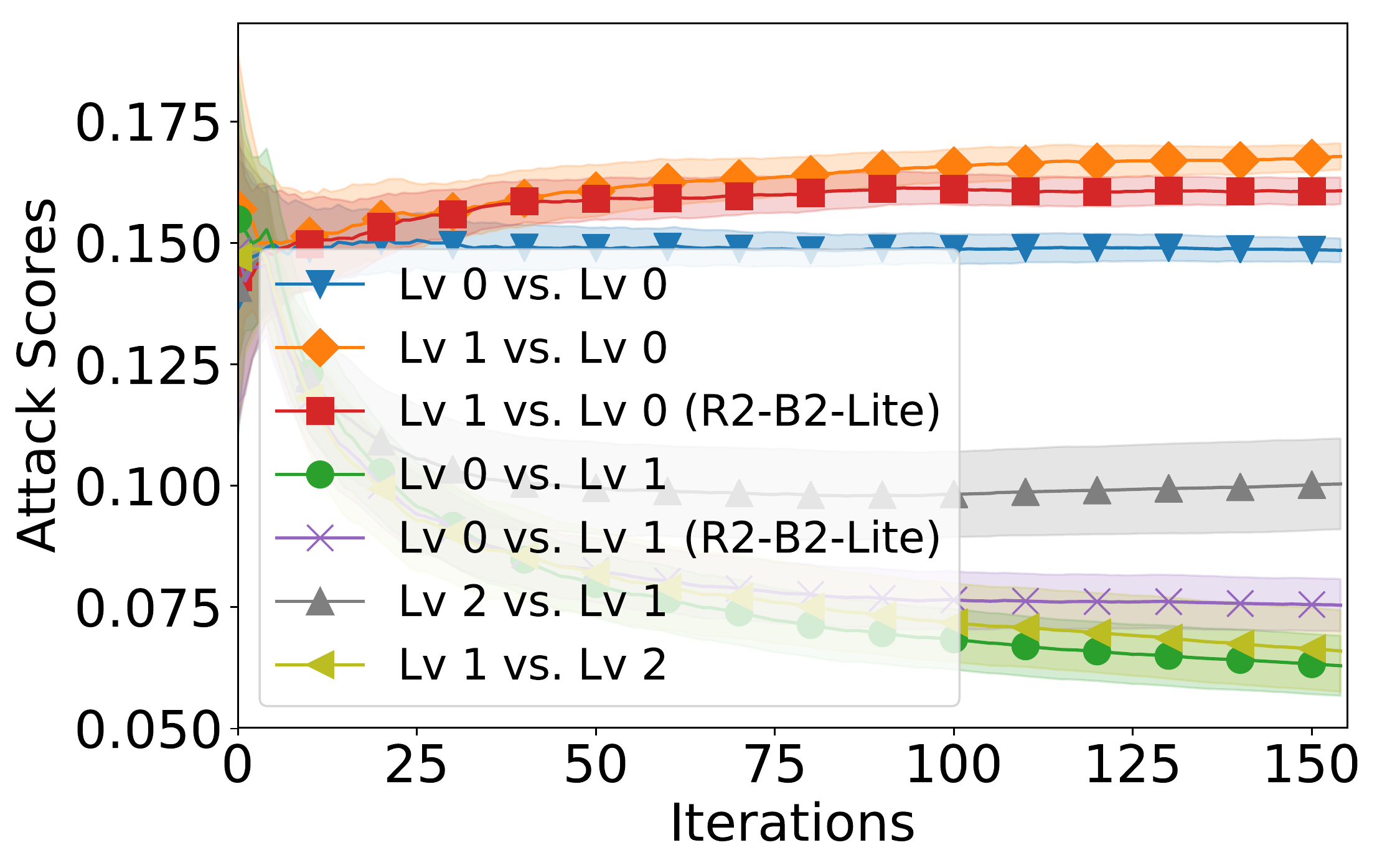}
    \caption{Attack scores averaged over $30$ images from MNIST. Each image is again averaged over $5$ initializations of $5$ randomly selected actions.}
    \label{fig:mnist_multiple_images}
\end{figure}

\begin{figure}
    \centering
    \includegraphics[width=0.5\linewidth]{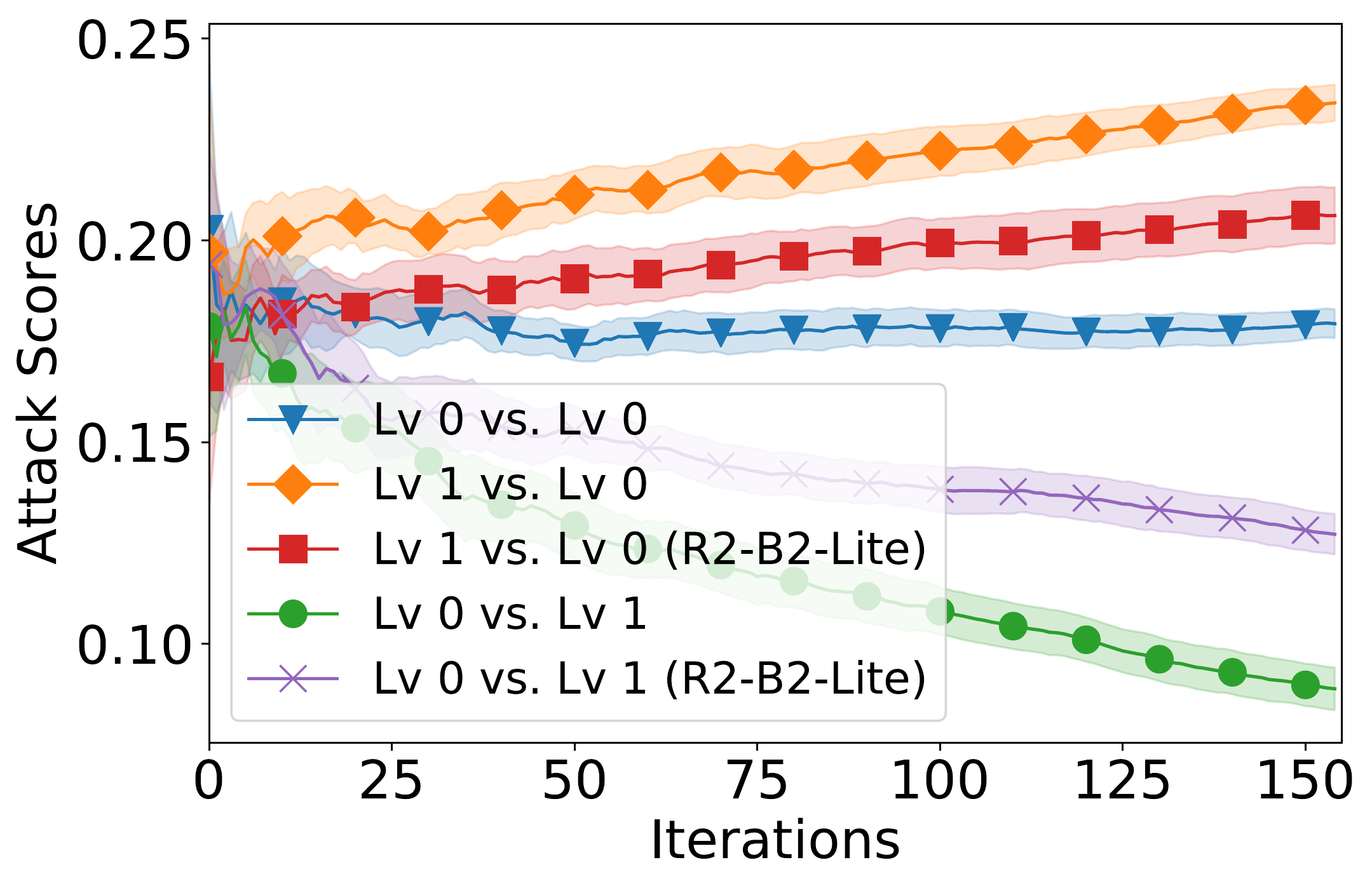}
    \caption{Attack scores averaged over $9$ images from CIFAR-$10$. Each image is again averaged over $5$ initializations of $5$ randomly selected actions.}
    \label{fig:cifar_multiple_images}
\end{figure}

Figs.~\ref{fig:mnist_multiple_images} and~\ref{fig:cifar_multiple_images} show the attack scores on the MNIST and CIFAR-$10$ datasets averaged over multiple randomly selected images 
($30$ images for MNIST and $9$ images for CIFAR-$10$).
These figures yield consistent observations with those presented in the main text, except that for MNIST (Fig.~\ref{fig:mnist_multiple_images}), the attack scores are generally lower 
(compared with the blue curve where both $\mathcal{A}$ and $\mathcal{D}$ reason at level $0$), 
which could be explained by the fact that the images in the MNIST dataset are generally easier to defend than to attack (i.e., it is easier to make the attack score lower than to make it higher, as explained in the previous paragraph)
because of the simplicity of the dataset and the high accuracy of the target ML model (i.e., a validation accuracy of $99.25\%$).
As a result, when $\mathcal{A}$ reasons at level $2$ and $\mathcal{D}$ reasons at level $1$, the attack score is lower than when both agents reason at level $0$ (compare the gray and blue curves in Fig.~\ref{fig:mnist_multiple_images}).
In addition to the above-mentioned factor that the MNIST dataset is in general harder to attack (i.e., harder to make the attack score higher than to make it lower),
this deviation from our theoretical result (Theorem~\ref{theorem_level_k}) might also be attributed to the error in approximating the expectation operator in level-$1$ reasoning.
However, the benefit of reasoning at one level higher can still be observed in this case, since when the reasoning level of $\mathcal{D}$ is fixed at $1$, 
it is still beneficial for $\mathcal{A}$ to reason at level $2$ (i.e., the gray curve) instead of level $0$ (i.e., the green curves).
The corresponding average number of successful attacks in $150$ iterations for different reasoning levels yield the same observations and interpretations as Figs.~\ref{fig:mnist_multiple_images} and~\ref{fig:cifar_multiple_images}:
For MNIST (Fig.~\ref{fig:mnist_multiple_images}), the number of successful attacks are (in the order of the figure legend from top to bottom) $20.4, 23.0, 21.3, 9.7, 11.0, 12.4, 7.9$,
for CIFAR-$10$ (Fig.~\ref{fig:cifar_multiple_images}), they are $32.9, 43.0, 38.8, 12.2, 21.0$.

\begin{figure}
    \centering
    \includegraphics[width=0.5\linewidth]{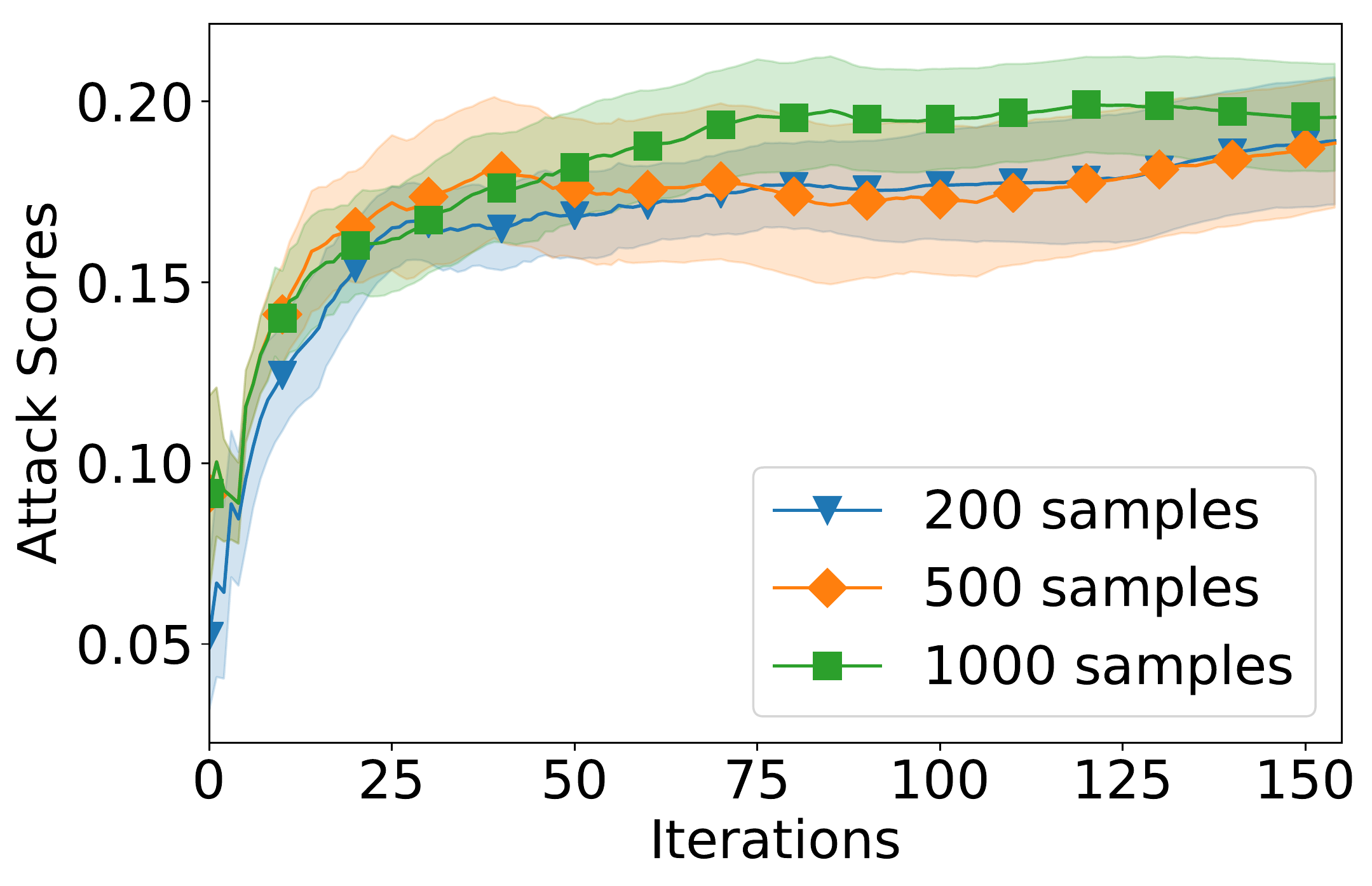}
    \caption{Attack scores for MNIST when $\mathcal{A}$ (the attacker) and $\mathcal{D}$ (the defender) reason at levels $2$ and $1$ respectively,
     with different number of samples used for approximating the expectation for level $1$ reasoning. 
    The corresponding number of successful attacks (for $200$, $500$ and $1000$ samples) are $2.6$, $3.0$ and $3.3$.}
    \label{fig:mnist_approx_samples}
\end{figure}

\textbf{(c) Impact of the Number of Samples Used for Approximating the Expectation in Level-$1$ Reasoning}\\
For the results reported in the main text, the number of samples used to approximate the expectation in level-$1$ reasoning are $500$ for MNIST (Fig.~\ref{fig:player_1}d and e) and $1,000$ for CIFAR-$10$ (Fig.~\ref{fig:player_1}f).
Note that since the input dimension is higher for CIFAR-$10$, a larger number of samples is needed to accurately approximate the level-$0$ mixed strategy (over which the expectation in level-$1$ reasoning is taken).
Here, we further investigate the impact of the number of samples used in the approximation of the expectation operator in level-$1$ reasoning~\eqref{eq:level_1_defender}.
Fig.~\ref{fig:mnist_approx_samples} shows the attack scores for the MNIST dataset when $\mathcal{A}$ and $\mathcal{D}$ reason at levels $2$ and $1$ respectively
when different number of samples are used for the approximation. Random search is used as the level-$0$ mixed strategy.
The figure, as well as the corresponding number of successful attacks, demonstrates that the attack becomes more effective as more samples are used for the approximation.
The benefit offered by using more samples for the approximation results from the fact that with a better accuracy at estimating $\mathcal{D}$'s level-$1$ action~\eqref{eq:level_1_attacker} 
(i.e., the level-$1$ action of $\mathcal{D}$ simulated by $\mathcal{A}$ is more likely to be the same as the actual level-$1$ action selected by $\mathcal{D}$), 
the attacker is able to best-respond to $\mathcal{D}$'s action more accurately~\eqref{eq:level_2_defender}, thus leading to an improved performance.

\subsubsection{Defense against State-of-the-art Adversarial Attack Methods}
\label{app:adv_ml_parsimonious}
\begin{figure}
    \centering
    \includegraphics[width=0.75\linewidth]{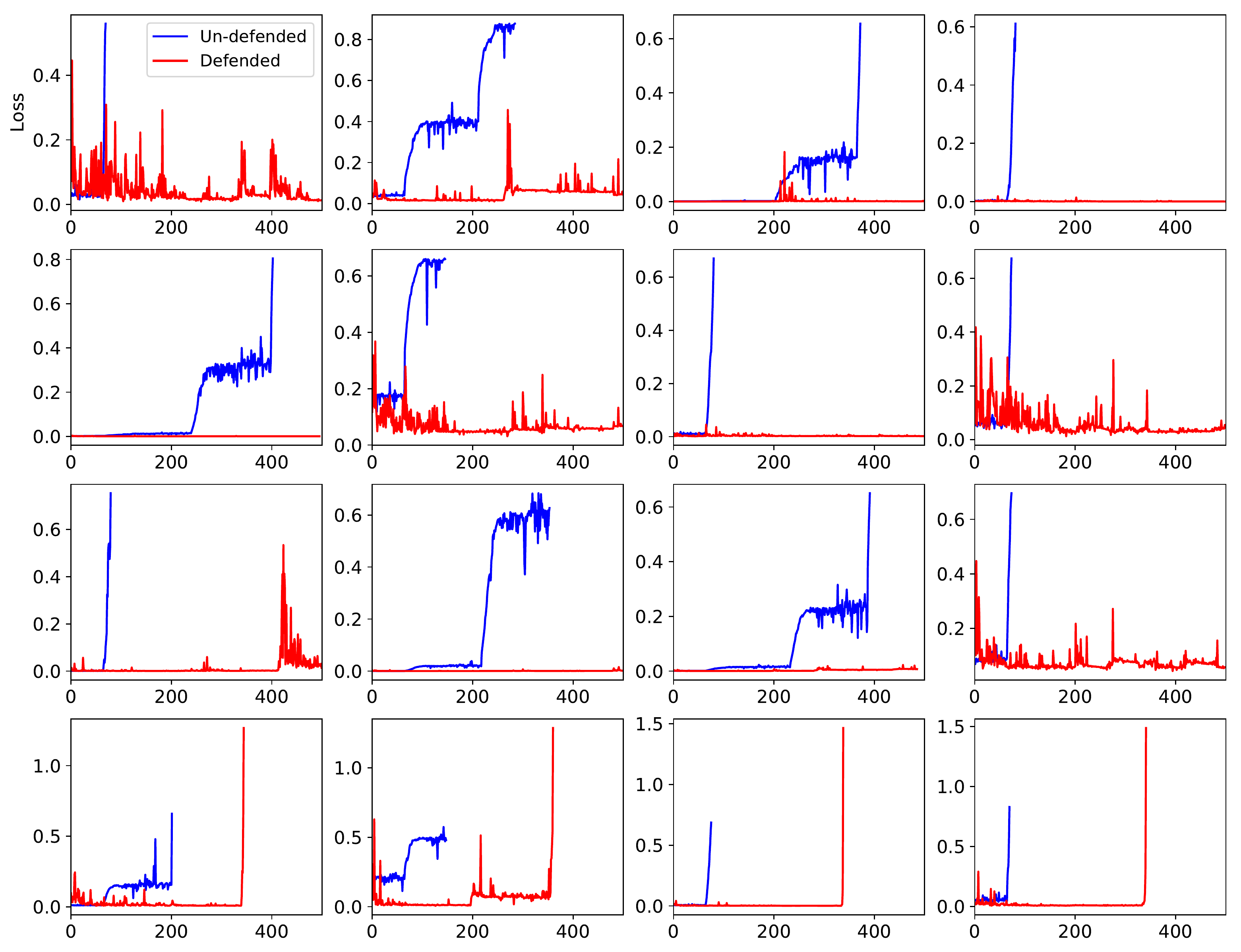}
    \caption{The loss of the Parsimonious algorithm with and without our level-$1$ R2-B2 defender on some selected images.
    For the images on the first three rows, Parsimonious fails to achieve any successful attack;
    for the images on the last row, our level-$1$ R2-B2 defender requires Parsimonious to use a significantly larger number of queries to obtain a successful attack.
    }
    \label{fig:parsimonious_all}
\end{figure}

\textbf{(a) Against the Parsimonious Attacker\footnote{\url{https://github.com/snu-mllab/parsimonious-blackbox-attack}}}\\
Since the Parsimonious algorithm is deterministic (assuming that the random seed is fixed), it corresponds to a level-$0$ pure strategy, which is equivalent to
a mixed strategy with all probability measure concentrated on a single action. 
Therefore, in our setting, when $\mathcal{D}$ (the defender) is selecting its level-$1$ strategy in iteration $t$ using R2-B2, 
it knows exactly the action (perturbations) that $\mathcal{A}$ (the attacker) will select in the current iteration $t$.
To make the setting more practical, we use the (encoded) image perturbed by $\mathcal{A}$ (instead of the encoded perturbations as in the experiments 
in Section~\ref{subsec:adv_ml}) as the action of $\mathcal{A}$, $\mathbf{x}_1$.
Specifically, every time $\mathcal{D}$ receives the perturbed image from $\mathcal{A}$, $\mathcal{D}$ encodes the image using its VAE, 
and use the encoded latent vector (i.e., the mean vector of the encoded latent distribution) as the input from $\mathcal{A}$ in the current iteration
(i.e., $\mathbf{x}_{1,t}$).
As a result, in every iteration, $\mathcal{D}$ naturally gains access to the action of $\mathcal{A}$ in the current iteration $\mathbf{x}_{1,t}$
and can thus reason at level $1$ by best-responding to $\mathbf{x}_{1,t}$.
Therefore, $\mathcal{D}$ has natural access to $\mathcal{A}$'s history of selected actions, which, combined with the fact that the game is constant-sum (which allows $\mathcal{D}$ to 
know $\mathcal{A}$'s payoff by observing $\mathcal{D}$'s own payoff), 
satisfies the requirement of perfect monitoring.
Note that Parsimonious maximizes the loss (instead of the attack score as in the experiments in Section~\ref{subsec:adv_ml}) of a test image as the objective of attack, 
so to be consistent with their algorithm, we use the negative loss as the payoff function of our level-$1$ R2-B2 defender.
Refer to Fig.~\ref{fig:parsimonious_all} for the loss values achieved by Parsimonious with and without our level-$1$ R2-B2 defender for some selected images.
The losses for different images are reported individually since they are highly disparate across different images, thus making their average losses hard to visualize.

\begin{figure}
    \centering
    \includegraphics[width=0.5\linewidth]{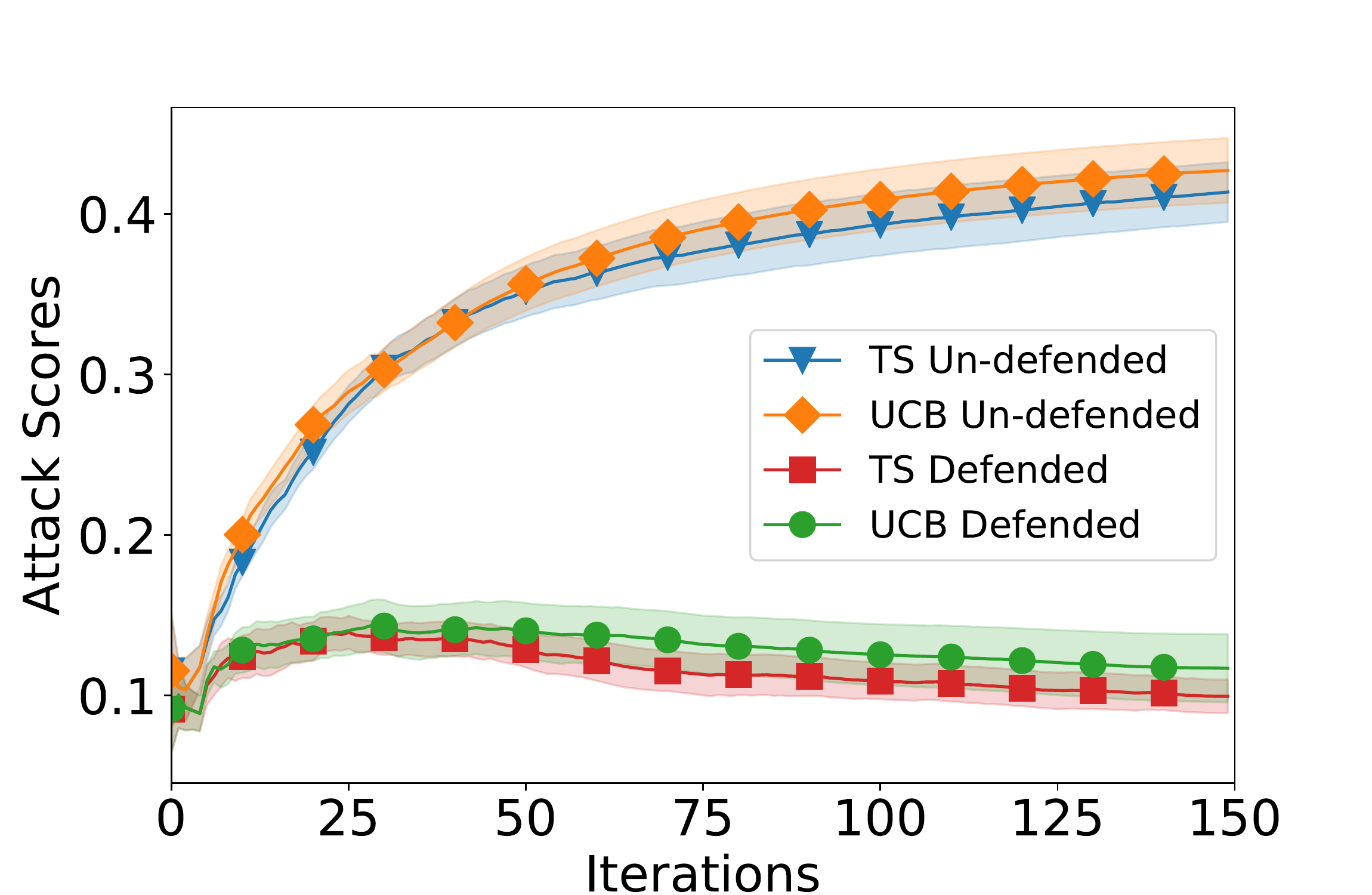}
    \caption{Attack scores achieved by the black-box attacker using BO with the GP-UCB and Thompson sampling acquisition functions, with and without our level-$1$ R2-B2-Lite defender.
    The corresponding number of successful attacks are $70.1$, $67.0$, $0.8$ and $0.7$ respectively (in the order of the figure legend from top to bottom).}
    \label{fig:independent_attacker_bo}
\end{figure}

\textbf{(b) Against the BO Attacker}\\
In addition to evaluating the effectiveness of our level-$1$ R2-B2 defender using the state-of-the-art Parsimonious algorithm (Section~\ref{exp:comparison_parsimonious}), we also investigate 
whether our level-$1$ R2-B2 defender is able to defend against black-box adversarial attacks using BO, which has recently become popular as a sample-efficient black-box method for adversarial attacks~\cite{ru2020bayesopt}.
Specifically, as a gradient-free technique to optimize black-box functions, BO can be naturally used to maximize the attack score (i.e., the output) over the space of adversarial perturbations (i.e., the input).
Note that in contrast to the attacker in Section~\ref{subsec:adv_ml}, the BO attacker here is not aware of the existence of the defender and thus the input to its GP surrogate only consists of 
the (encoded) perturbations of the attacker.
We adopt two commonly used acquisition functions for BO: (a) Thompson sampling (TS) which, as a randomized algorithm, corresponds to a level-$0$ mixed strategy, and (b) GP-UCB,
which represents a level-$0$ pure strategy.
For both types of adversarial attacks, we let our level-$1$ defender run the R2-B2-Lite algorithm.
In particular, when the attacker uses the GP-UCB acquisition function, in each iteration, the defender calculates/simulates the action (perturbations) that would be selected by the attacker in the current iteration,
and best-responds to it; when TS is adopted by the attacker as the acquisition function, the defender draws a sample using the attacker's randomized level-$0$ TS strategy in the current iteration, and best-responds to it.
Fig.~\ref{fig:independent_attacker_bo} shows the results of adversarial attacks using the TS and GP-UCB acquisition functions with and without our level-$1$ R2-B2-Lite defender.
As demonstrated in the figure, our level-$1$ R2-B2-Lite defender is able to effectively defend against 
and almost eliminate the impact of both types of adversarial attacks (i.e., allow the attacker to succeed for less than once over $150$ iterations).

\subsection{Multi-Agent Reinforcement Learning}
\label{app:marl}
The multi-agent particle environment adopted in our experiment can be found at \url{https://github.com/openai/multiagent-particle-envs}. 
The state and action of the two predators (referred to as predator $1$ and predator $2$ for simplicity), are represented by a $14$-dimensional vector and a $5$-dimensional vector respectively,
whereas the state and action of the prey are represented by a $12$-dimensional vector and a $5$-dimensional vector correspondingly.
For simplicity, we perform direct policy search using a linear policy space. That is, the policy of each predator is represented by a $14\times 5$ matrix, which maps a $14$-dimensional state vector to 
a $5$-dimensional action vector, thus producing the action to be taken by the predator according to the current policy when the predator is in a particular state. 
Similarly, the policy of the prey corresponds to a $12\times 5$ matrix, which is able to map a $12$-dimensional state vector to a $5$-dimensional action vector.
To further simplify the setting and reduce the dimensionality of the policy space, we use rank-$1$ approximations of the policy matrices.
That is, the $14\times 5$ policy matrix of each predator is obtained by the outer product of a $14$-dimensional vector and a $5$-dimensional vector, whereas the $12\times 5$ policy matrix of the prey
is attained by the outer product of a $12$-dimensional vector and a $5$-dimensional vector. As a result, the policy of each predator is represented by $14+5=19$ parameters, whereas
the policy of the prey is characterized by $12+5=17$ parameters. Therefore, the dimension of the input to the GP surrogate models is $19+19+17=55$.
For every one of the $55$ input dimensions, the search space is $[-1, 1]$.
In each iteration of the repeated game, after all agents have selected their policy parameters, the agents use their respective policies to interact in the environment for $50$ steps and use
their obtained returns (i.e., cumulative rewards) as the corresponding payoff; every iteration of the repeated game involves $5$ independent runs in the environment (with different initializations) using the selected policy parameters,
and the averaged return over the $5$ independent runs is reported as the corresponding observed payoff.
For ease of visualization, the returns are clipped and scaled into the range $[0, 1]$.
All agents use random search as the level-$0$ strategy due to the high dimension of input action space; refer to Appendix~\ref{app:adv_ml}a for a detailed explanation about this choice.
The GP hyperparameters are optimized via maximizing the marginal likelihood after every $10$ iterations.

\begin{figure}
    \centering
    \includegraphics[width=0.3\linewidth]{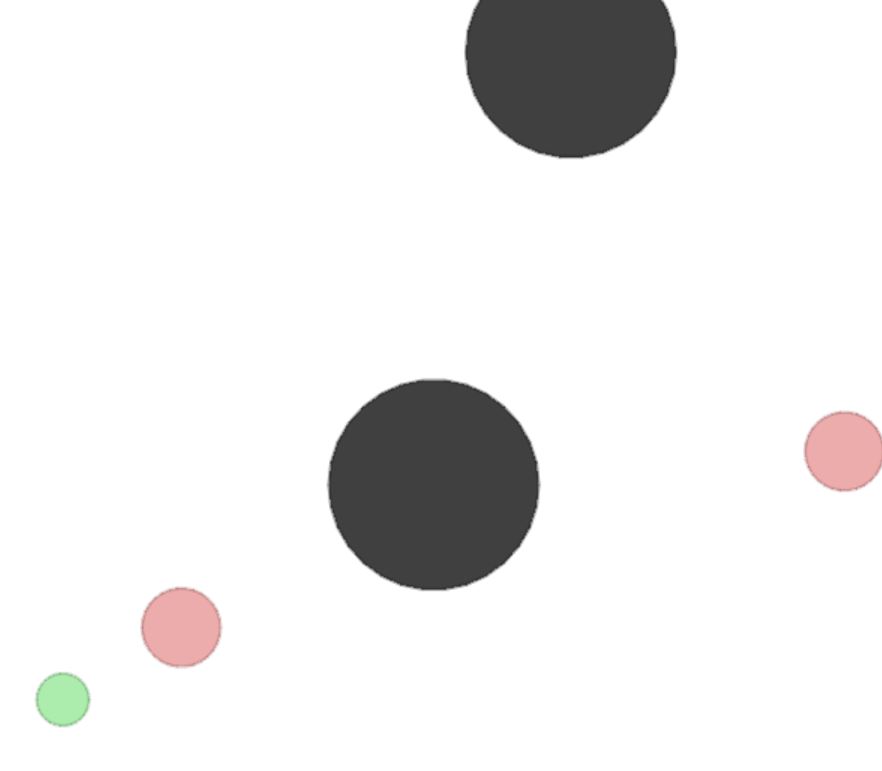}
    \caption{Illustration of the predator-prey game. Red: predators; green: prey; black: obstacles.}
    \label{fig:simple_tag_illu}
\end{figure}

%%%%%%%%%%%%%%%%%%%%%%%%%%%%%%%%%%%%%%%%%%%%%%%%%%%%%%%%%%%%%%%%%%%%%%%%%%%%%%%
%%%%%%%%%%%%%%%%%%%%%%%%%%%%%%%%%%%%%%%%%%%%%%%%%%%%%%%%%%%%%%%%%%%%%%%%%%%%%%%

\end{document}